\def\wacvPaperID{1423}
\def\confName{WACV}
\def\confYear{2023}

\def\paperTitle{Sketch-based Video Object Localization}

\def\authorBlock{
    Sangmin Woo$^1$ \quad
    So-Yeong Jeon$^{1,2}$ \quad
    Jinyoung Park$^{1}$ \quad
    Minji Son$^{3}$ \quad
    Sumin Lee$^{1}$ \quad
    Changick Kim$^{1}$\\
    \textsuperscript{1}KAIST \quad \textsuperscript{2}Korea Agency for Defense Development \quad \textsuperscript{3}LG Electronics \\
    {\tt\small \textsuperscript{1}\{smwoo95, presentover, jinyoungpark, suminlee94, changick\}@kaist.ac.kr \textsuperscript{3}minji13.son@lge.com} \\
    {\tt \url{https://github.com/sangminwoo/SVOL}}
}

\newif\ifreview 
\newif\ifarxiv \newcommand{\arxiv}{\arxivtrue}
\newif\ifcamera 
\newif\ifrebuttal 

\arxiv

\pdfoutput=1
\documentclass[10pt,twocolumn,letterpaper]{article}
\ifreview \usepackage[review,applications]{wacv} \fi
\ifarxiv \usepackage[pagenumbers]{wacv} \fi
\ifrebuttal \usepackage[rebuttal]{wacv} \fi
\ifcamera \usepackage{wacv} \fi

\usepackage{times}
\usepackage{epsfig}
\usepackage{graphicx}
\usepackage{amsmath}
\usepackage{amssymb}
\usepackage{booktabs}
\usepackage{comment}
\usepackage{color}
\usepackage{caption}
\usepackage[numbers,sort]{natbib}
\usepackage{multirow}
\usepackage{algorithm}
\usepackage{algorithmic}
\usepackage{dsfont}
\usepackage{bm}  
\usepackage{xspace}
\usepackage{colortbl}
\usepackage{enumitem}
\usepackage{placeins}
\usepackage{footnote}
\usepackage{tikz}
\usepackage{amsfonts}    
\usepackage{bbding}
\usepackage{nicefrac}       
\usepackage{microtype}      
\usepackage{tabularx}
\usepackage{float}
\usepackage{tabulary}	
\usepackage{wrapfig}
\usepackage{relsize}

\usepackage{times}
\usepackage{microtype}
\usepackage{epsfig}
\usepackage{caption}
\usepackage{float}
\usepackage{placeins}
\usepackage{color, colortbl}
\usepackage{stfloats}
\usepackage{enumitem}
\usepackage{tabularx}
\usepackage{xstring}
\usepackage{multirow}
\usepackage{xspace}
\usepackage{url}
\usepackage{subcaption}
\usepackage{xcolor}
\usepackage[hang,flushmargin]{footmisc}

\ifcamera \usepackage[accsupp]{axessibility} \fi

% ...

\ifarxiv  \fi

\newcommand{\R}[1]{{%
    \textbf{%
        \ifstrequal{#1}{1}{\textcolor{red}{R#1}}{%
        \ifstrequal{#1}{2}{\textcolor{blue}{R#1}}{%
        \ifstrequal{#1}{3}{\textcolor{magenta}{R#1}}{%
        \ifstrequal{#1}{4}{\textcolor{teal}{R#1}}{%
                           \textcolor{cyan}{R#1}%
        }}}}%
    }%
}}

\newcommand{\eqnsm}[2]{\begin{equation}\label{eq:#1}#2\end{equation}}
\newcommand{\eqnalism}[2]{\begin{equation}\begin{aligned}\label{eq:#1}#2\end{aligned}\end{equation}}

\newlength\abovesecmargin
\newlength\belowsecmargin
\newlength\abovesubsecmargin
\newlength\belowsubsecmargin
\newlength\abovesubsubsecmargin
\newlength\belowsubsubsecmargin
\newlength\paramargin
\newlength\abovetabcapmargin
\newlength\belowtabcapmargin
\newlength\abovefigcapmargin
\newlength\belowfigcapmargin

\makeatletter\renewcommand\paragraph{\@startsection{paragraph}{4}{\z@}
  {.5em \@plus1ex \@minus.2ex}{-.5em}{\normalfont\normalsize\bfseries}}\makeatother

\makeatletter
\DeclareRobustCommand\onedot{\futurelet\@let@token\@onedot}
\def\@onedot{\ifx\@let@token.\else.\null\fi\xspace}

\def\eg{\emph{e.g}\onedot} 
\def\ie{\emph{i.e}\onedot} 
 
\def\etc{\emph{etc}\onedot} \def\vs{\emph{vs}\onedot}
\def\wrt{w.r.t\onedot} 
 
\def\aka{a.k.a\onedot} \def\etal{\emph{et al}\onedot}
\makeatother

\newcommand{\argmin}{\mathop{\rm argmin}\limits}

\newlength\savewidth\newcommand\shline{\noalign{\global\savewidth\arrayrulewidth
  \global\arrayrulewidth 1pt}\hline\noalign{\global\arrayrulewidth\savewidth}}
\newcommand{\tablestyle}[2]{\setlength{\tabcolsep}{#1}\renewcommand{\arraystretch}{#2}\centering\footnotesize}

\newcolumntype{x}[1]{>{\centering\arraybackslash}p{#1pt}}
\newcolumntype{y}[1]{>{\raggedright\arraybackslash}p{#1pt}}
\newcolumntype{z}[1]{>{\raggedleft\arraybackslash}p{#1pt}}

\definecolor{Gray}{gray}{0.9}
\definecolor{Green}{rgb}{0.2, 0.7, 0.1}
\definecolor{Orange}{rgb}{0.8, 0.5, 0.2}
\definecolor{Yellow}{RGB}{255, 192, 0}
\definecolor{todo}{rgb}{0.8, 0.4, 0.2}

\definecolor{citecolor}{HTML}{0071bc}
\definecolor{linkcolor}{HTML}{ED1C24}
\definecolor{plus}{HTML}{0071bc}
\definecolor{minus}{RGB}{153,10,10}
\definecolor{quickdraw}{rgb}{.5, .0, .5}
\definecolor{tuberlin}{rgb}{0, 0.3, 0.8}
\definecolor{sketchy}{rgb}{0, .5, 0}

\definecolor{urlcolor}{rgb}{0.2, 0.7, 0.1}
\definecolor{scorered}{HTML}{e4485a}
\definecolor{scoreblue}{HTML}{4a7ee8}
\definecolor{scoregreen}{HTML}{80ba0e}

\definecolor{purple0}{HTML}{e9e9f3}
\definecolor{purple}{HTML}{dcdaed}
\definecolor{purple1}{HTML}{bab6da}

\usepackage{pifont}
\newcommand{\cmark}{\ding{51}}
\newcommand{\xmark}{\ding{55}}

\newcommand{\pacc}[1]{{\bf \fontsize{7}{42}\selectfont \color{plus!80} #1}}

\newcommand*\rot{\rotatebox{90}}

\newcommand{\indic}[1]{\mathds{1}_{\{#1\}}}
\newcommand{\noobject}{\varnothing}
\renewcommand{\Sigma}{\mathfrak{S}}

\newcommand{\ours}{\mbox{SVANet}\xspace}
\newcommand{\sketchy}{\color{sketchy} Sketchy}
\newcommand{\quickdraw}{\color{quickdraw} QuickDraw}
\newcommand{\tuberlin}{\color{tuberlin} TU-Berlin}

\usepackage{xr-hyper}

\makeatletter
\newcommand*{\addFileDependency}[1]{
  \typeout{(#1)}
  \@addtofilelist{#1}
  \IfFileExists{#1}{}{\typeout{No file #1.}}
}

\makeatother

\usepackage[pagebackref,breaklinks,colorlinks]{hyperref}
\usepackage[capitalize]{cleveref}
\crefname{section}{Sec.}{Secs.}
\crefname{table}{Table}{Tables}
\crefname{figure}{Fig.}{Figs.}

\begin{document}
\title{\paperTitle}
\author{\authorBlock}
\maketitle

\begin{abstract}
We introduce Sketch-based Video Object Localization (SVOL), a new task aimed at localizing spatio-temporal object boxes in video queried by the input sketch.
We first outline the challenges in the SVOL task and build the Sketch-Video Attention Network (\ours) with the following design principles: (i) to consider temporal information of video and bridge the domain gap between sketch and video; (ii) to accurately identify and localize multiple objects simultaneously; (iii) to handle various styles of sketches; (iv) to be classification-free.
In particular, \ours is equipped with a \textit{Cross-modal Transformer} that models the interaction between learnable object tokens, query sketch, and video through attention operations, and learns upon a \textit{per-frame set matching} strategy that enables frame-wise prediction while utilizing global video context.
We evaluate \ours on a newly curated SVOL dataset.
By design, \ours successfully learns the mapping between the query sketches and video objects, achieving state-of-the-art results on the SVOL benchmark.
We further confirm the effectiveness of \ours via extensive ablation studies and visualizations.
Lastly, we demonstrate its transfer capability on unseen datasets and novel categories, suggesting its high scalability in real-world applications.
\end{abstract}
\begin{figure}[t!]
    \centering
    \resizebox*{\linewidth}{!}{%
        \includegraphics{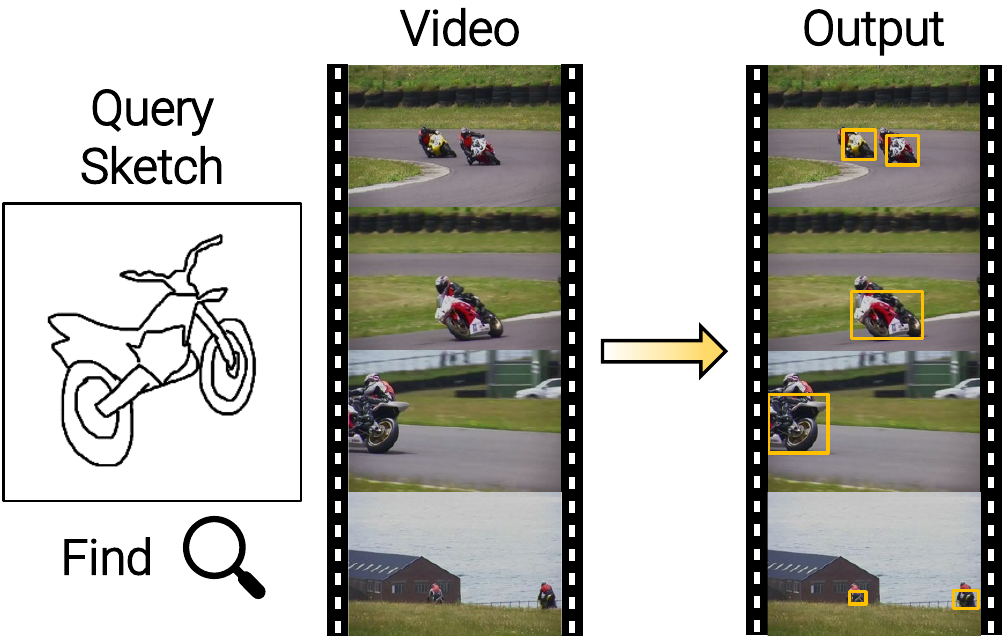}
    }
    \vskip \abovefigcapmargin
    \caption{
        \textbf{Illustration of the SVOL task.} Given a query sketch, the goal is to find all \textit{object boxes} (colored in {\color{Yellow}yellow}) spatio-temporally that match the sketch object in a video.
        Query sample is randomly drawn from~{\sketchy} dataset.
    }
    \label{fig:teaser}
    \vspace{\belowfigcapmargin}
\end{figure}

\vspace{\abovesecmargin}\vspace{-5mm}
\section{Introduction}
\label{sec:intro}
\vspace{\belowsecmargin}
A sketch is worth a thousand words.
It can even convey ideas that are hard to explain in words.
Due to the concise and abstract nature of the sketch, it can be illustrative, making it an excellent tool for a variety of applications~\cite{chen2018sketchygan,isola2017image,kwan2019mobi3dsketch,portenier2018faceshop,yu2016sketch,bhunia2022doodle,wang2021sketch,2022_Sketch_survey}.
Meanwhile, query-based localization is one of the long-sought goals for visual understanding.
The literature has been studied at a variety of query types (\eg, image, language, sketch) and domains (\eg, image, video)~\cite{liu2013image,chen2021transformer,yang2022lavt,woo2022explore,anne2017localizing,hsieh2019one,kong2014you,li2018high,tripathi2020sketch,riba2021localizing,su2021stvgbert}.
While numerous studies have shown remarkable results using image or language as query, both have their own limitations.
Images containing a specific object of interest may be difficult to collect due to privacy or copyright issues~\cite{bhunia2022doodle}, and the utility of language is limited as it varies per country.
As an alternative, using sketch as a query brings several advantages.
It allows for immense expressive flexibility and can transcend language barriers~\cite{literat2013pencil}. 
Moreover, due to the recent spread of touchscreen devices (\eg, smartphones, tablets), sketches have become easier to obtain than ever~\cite{taele2023sketchrec}.
Amid the explosive growth of video data, sketch has emerged as an appealing candidate for user interface in online video platforms thanks to these properties.
Despite its promise, the use of sketch as query for object localization in the video domain has not yet been explored.

In this work, we propose a new task called Sketch-based Video Object Localization (SVOL) that aims to localize objects in videos with the query sketch (see~\cref{fig:teaser}).
We first identify several challenges that the setting of SVOL yields, including but not limited to:
\textbf{(i)} As objects move, they can generate motion blurs or occlude parts of other objects, thus distorting their appearances~\cite{liang2018planar}.
Moreover, objects in the scene may suddenly disappear, or objects that were not in the scene may suddenly appear.
These dynamic changes over time complicate the matching of sketch to its corresponding objects.
\textbf{(ii)} Multiple objects can appear in a video.
Therefore, it is important not only to accurately differentiate between the target objects from multiple objects belonging to different categories, but also to find all objects that match the sketch query simultaneously. 
\textbf{(iii)} As for sketch, a single object can be drawn in various ways~\cite{sain2021stylemeup}.
Unlike natural videos, sketches lack color, texture, and background information, resulting in a high degree of freedom.
This allows sketches to be drawn in a variety of styles (\ie, different abstraction levels).
\textbf{(iv)} There should be no explicit category prediction (\ie, no fixed classes) in the SVOL system.
As with all query-based localization tasks, SVOL requires finding the best matching objects given a query sketch (not the category itself).

Driven by this analysis, we propose \ours that serves as a strong baseline for the SVOL task.
\ours takes extracted video and sketch representations as inputs and predicts box coordinates and objectness scores end-to-end.
Our \ours is built on several design principles:
\textbf{First}, we propose a novel \textit{Cross-modal Transformer (CMT)} that not only closes the domain discrepancy between sketch and video but also models video temporal context.
We equip CMT with four attention operations~\cite{vaswani2017attention} to leverage their strong relational modeling capability.
By design, CMT emphasizes important content by learning the correlation between sketch and video representations, and incorporates temporal context by modeling intra-content relationships.
Also, CMT takes object tokens as inputs and transforms them into predictions of box coordinates and objectness scores by learning their internal interactions and by referring to joint sketch-video representations.
\textbf{Second}, we formulate the SVOL task as a set prediction problem~\cite{carion2020end} and employ a \textit{per-frame set matching} strategy.
We predict all bounding boxes across the video frames, and find the best matching between predicted and ground truth boxes that minimizes the matching cost.
The overall training loss is then defined based on the matching results.
Instead of matching whole video-level results with video-level ground truths, we perform set matching frame-by-frame.
This enables the prediction of multiple objects in parallel while utilizing the global video context.
\textbf{Third}, \ours is designed to be compatible with various sketch styles.
\ours learns to embed the sketch objects of the same category into a similar subspace of a high-dimensional latent space, regardless of differences in sketch styles (\eg, shape, pose, line thickness, \etc).
This style-agnostic property enables \ours to generalize well on unseen sketch datasets with varying degrees of abstraction.
\textbf{Last}, \ours has no explicit classification in the pipeline.
This allows \ours to learn the mapping between sketch and video objects based on implicit similarity (\eg, symbolic meaning, appearance, \etc).
This classification-free property of \ours extends its applicability to any kind of free-form sketches, allowing us to query over novel object classes.

\begin{figure}[t!]
    \centering
    \scriptsize
    \includegraphics[width=\linewidth]{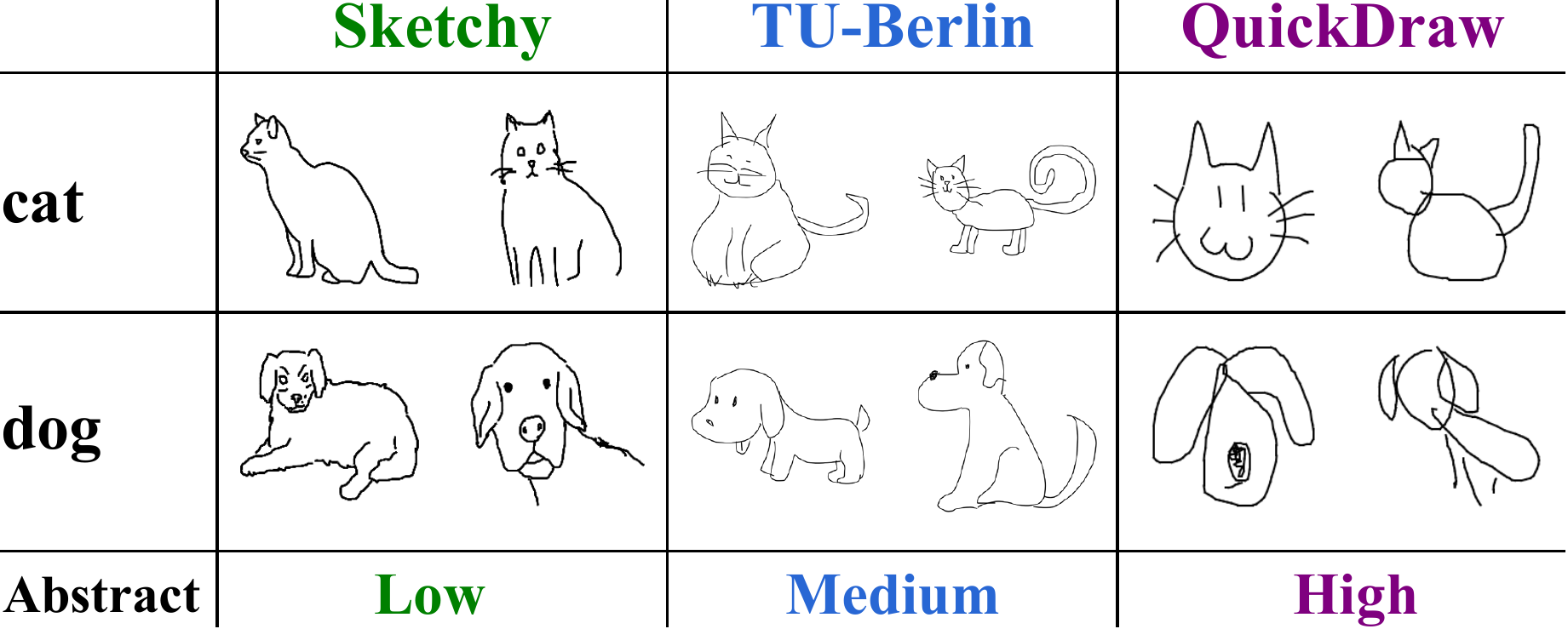}
    \vskip \abovefigcapmargin
    \caption{
        \textbf{Sketch datasets comparison.} {\bf \sketchy}~\cite{sangkloy2016sketchy} is the most realistic since it is drawn after photographic objects, {\bf \quickdraw}~\cite{jongejan2016quick} has the highest level of abstraction due to limited drawing time ($<$ 20 secs), and {\bf \tuberlin}~\cite{eitz2012humans} lies halfway between them.
    }
    \label{fig:dataset}
    \vspace{\belowfigcapmargin}
\end{figure}

To benchmark our approach and show the potential of using sketch as query, we present a new SVOL dataset curated from the video dataset, ImageNet-VID~\cite{russakovsky2015imagenet}, and three sketch datasets with varying degrees of abstraction (see~\cref{fig:dataset}): {\sketchy}~\cite{sangkloy2016sketchy}, {\tuberlin}~\cite{eitz2012humans}, and {\quickdraw}~\cite{jongejan2016quick}.
We show that \ours outperforms the strong image-level baseline (Sketch-DETR)~\cite{riba2021localizing} by a significant margin: $\sim$29.4\%, 17.7\%, and 16.8\% improvement of mIoU using {\sketchy}, {\tuberlin}, and {\quickdraw} sketch datasets, respectively.
This implies that \ours effectively resolves the limitation of image-level baselines with temporal video context.
Moreover, we verify the effectiveness of several design choices of \ours through extensive ablation studies and analyze its behavior with several visualizations.
Finally, we evaluate transfer performances of \ours on unseen datasets (with different abstraction levels or sketch styles) and novel categories that are unseen during training.
The results demonstrate that \ours is robust to style variations and that the learned sketch-video mapping function generalizes well to novel classes of sketches.
These appealing properties are ideal for several query-based applications in practice, such as large-scale video platforms, in that the system can flexibly respond to diverse inputs from users.

Our contributions are summarized as follows:

\noindent\textbf{New task.}
We opened up a new challenging task, Sketch-based Video Object Localization (SVOL).
We also identified several challenges that the SVOL task setting brings.

\noindent\textbf{New dataset.}
We presented a new SVOL dataset curated from the video dataset (ImageNet-VID~\cite{russakovsky2015imagenet}) and three sketch datasets with different styles ({\sketchy}~\cite{sangkloy2016sketchy}, {\tuberlin}~\cite{eitz2012humans}, {\quickdraw}~\cite{jongejan2016quick}) and provide a benchmark with comparison against the frame-level baselines and several variants.

\noindent\textbf{Strong baseline.}
We proposed a novel framework named \ours, equipped with SVOL-tailored designs such as Cross-modal Transformer and per-frame set matching, that serves as a strong baseline on the benchmark: \ours improves mIoU by 29.4\%, 17.7\%, 16.8\% over the strong counterpart, Sketch-DETR~\cite{riba2021localizing}, using {\sketchy}, {\tuberlin}, {\quickdraw} sketch dataset, respectively.

\noindent\textbf{Extensive experiments.}
We thoroughly investigated the effects of model components with comprehensive ablative studies in various aspects. Last but not least, we demonstrated the strong generalizability of \ours on unseen datasets and novel categories, which makes sketch as query highly practical in real-world scenarios.


\vspace{\abovesecmargin}
\section{Related Work}
\label{sec:related}
\vspace{\belowsecmargin}
Our work builds on previous work in several areas, including sketch-based applications, query-based localization, and Transformer architecture.

    \vspace{\abovesubsecmargin}\subsection{Sketch-based Applications}\vspace{\belowsubsecmargin}
        Our work builds on the idea of using sketches as a way to query visual data.
        Sketch is a universal communication tool that is not bound by age, race, language, or national boundaries.
        Recently, sketch-based applications have grown at an unprecedented rate due to the widespread use of touchscreen devices such as smartphones and tablets that enable acquiring sketch data much easier than ever.
        Here are some examples of various applications using sketch~\footnote{For a more detailed list of sketch-based applications, we recommend referring to~\cite{2022_Sketch_survey}.}:
        
        \noindent\textbf{Image retrieval}: a user sketches an object or scene and the system retrieves similar image from a database~\cite{bhunia2022doodle,dey2019doodle,sain2021stylemeup,yu2016sketch}.

        \noindent\textbf{Image synthesis}: a user sketches an image and the system generates a photorealistic version of the image~\cite{chen2018sketchygan,isola2017image,wang2021sketch};

        \noindent\textbf{Image editing}: a user sketches desired changes to an image, and the system makes the changes automatically~\cite{portenier2018faceshop,yang2020deep}.

        \noindent\textbf{Robot interface}: a user sketches a task for a robot to perform, such as picking up an object and placing it in a specific location~\cite{sakamoto2009sketch,boniardi2016autonomous}.
        
        \noindent\textbf{3D modeling}: a user sketches a 3D object and the system generates a 3D model of the object~\cite{lun20173d,wang2015sketch}.
        
        \noindent\textbf{Augmented reality}: a user sketches an object, and the system overlays a 3D model of the object in the real-world environment~\cite{jiang2021handpainter,kwan2019mobi3dsketch}.

        Additionally, sketch is particularly effective at representing detailed features of an object like its shape, pattern, and pose.
        This ability to convey such fine-grained information has made sketches a popular tool in a variety of studies, including image~\cite{yu2021fine}, scene~\cite{liu2022scenesketcher}, and video~\cite{xu2020fine} retrieval.
        For example, in fine-grained image retrieval, sketch are used as queries to retrieve specific objects within images, such as a specific breed of dog or type of car~\cite{song2017deep,sun2022dli}.
        This allows users to search for images with specific visual characteristics, such as the shape of a dog's ears or a car's grille, so that objects of the same category can be differentiated.

        In the SVOL problem, while sketches have the capability to provide fine-grained information, we opt to focus on category-level object localization, \ie, localization is carried out in a shape- and pose-agnostic manner within the same category.
        This is because it is not natural to match a static sketch that has a specific shape and pose with objects in a video, whose shape and pose dynamically change over time.
        Additionally, by focusing on category-level localization, we can take advantage of the abstract nature of sketches.
        We can identify the location of objects in a video by sketching only some key features of the object, such as the headlights and grille of a car.

    \vspace{\abovesubsecmargin}\subsection{Query-based Localization}\vspace{\belowsubsecmargin}
        SVOL is related to the literature on object detection and tracking in videos, with added constraint of using a sketch as the query.
        Query-based object localization is similar to object detection~\cite{lin2017feature,liu2016ssd,redmon2016you,ren2015faster} (or video object detection~\cite{chen2020memory,feichtenhofer2017detect,han2016seq,wu2019sequence,zhou2022transvod,wang2022ptseformer}) in that they both aim to locate the bounding boxes of objects in an image (or a video).
        However, query-based localization is grounded on the given query, rather than pre-defined object classes.
        Query-based localization tasks have been studied using various query types in diverse dimensions.
        
        \vspace{\paramargin}\paragraph{Query.}
        \textbf{Image} queries can be localized based on appearance similarity, allowing themselves to be easily transferred to other objects with just a few image samples.
        This desirable property opens up a new avenue for research on one/few-shot localization~\cite{fan2020few,hsieh2019one,osokin2020os2d,vinyals2016matching}.
        However, image queries are hard to acquire in some privacy or security-related situations, making their usage in some applications difficult.
        \textbf{Language} queries, on the other hand, are highly useful given that we just need to describe the objects of interest in natural language.
        However, its universality is limited since the assumed language (English) may not be familiar to some people (non-native English speakers).
        As such, when the language is re-targeted, the neural network may require extra learning or translation before providing the query.
        \textbf{Sketch} queries differ from image queries in that they lack rich information such as color, texture, and background information; most free-hand sketches are composed solely of monochromatic lines, with no texture and context.
        In addition, since the sketches are drawn by envisioning abstract objects, even the same object may be drawn differently by a different person.
        These characteristics make sketch-based localization challenging.
        Nevertheless, we argue that this line of research is valuable since it offers the highest degree of freedom among the three query types and can transcend the language barrier (\ie, the sketch of `cat' can be understood whether or not you are a native English speaker).
        Our work focuses on the emerging role of sketch in the context of query-based localization.
        
        \vspace{\paramargin}\paragraph{Dimension.}
        \textbf{Temporal} localization aims to identify the temporal span (1D) in which the query object appears in the video.
        \textbf{Spatial} localization attempts to locate all object instances that match the query object within a still image (2D).
        \textbf{Spatio-temporal} localization seeks to locate all object instances that match the query object in every frame of video (3D).
        Our work belongs to the spatio-temporal category.

        \vspace{\paramargin}\paragraph{Task.}
        The challenging and open-ended nature of the query-based localization problem lends itself to a variety of tasks: image-based localization in natural images~\cite{bay2006surf,dalal2005histograms,liu2013image,lowe2004distinctive} and videos~\cite{chen2021transformer,chen2020siamese,li2019siamrpn++,li2018high} (\aka, visual object tracking); language-based localization in natural images~\cite{deng2018visual,deng2021transvg,fukui2016multimodal,hu2016natural,kong2014you,yang2022lavt} (\aka, visual grounding or referring expression comprehension) and videos~\cite{anne2017localizing,fan2020person,gao2017tall,lei2019tvqa+,su2021stvgbert,woo2022explore,zhang2020does,botach2022end} (\aka, video grounding or natural language video localization); and sketch object localization in natural images~\cite{tripathi2020sketch,riba2021localizing}.
        The most relevant tasks to ours are video grounding~\cite{fan2020person,lei2019tvqa+,su2021stvgbert,zhang2020does} and sketch object localization in images~\cite{tripathi2020sketch,riba2021localizing}.

        \vspace{\paramargin}\paragraph{Comparison.}
        In the realm of query-based localization research, various query types and domains have been explored, such as image, language, and video. However, one noticeable gap in the existing literature pertains to the absence of studies focused on sketch queries in the video domain. We seek to address this particular gap in knowledge. To address this research void, we propose a novel task ``Sketch-based Video Object Localization". This task is designed to facilitate the precise localization of spatio-temporal object boxes within video content, with the query input provided in the form of a sketch. This novel approach bridges the gap between sketch-based queries and video object localization, opening up new avenues for exploration and advancement in the field.

    \vspace{\abovesubsecmargin}\subsection{Sketch-based Image Object Localization}\vspace{\belowsubsecmargin}
        There are few methods for image object localization using sketch queries~\cite{tripathi2020sketch,riba2021localizing}, and we are the first to propose a sketch-based video object localization approach.
        We adopt Cross-modal Attention~\cite{tripathi2020sketch} and Sketch-DETR~\cite{riba2021localizing} as the image-level baseline in our SVOL benchmark.
    
        \noindent\textbf{Tripathi \etal, Cross-modal Attention}~\cite{tripathi2020sketch} generates object proposals that match the query sketch in an image.
        This mechanism operates above the off-the-shelf object detection framework, Faster R-CNN~\cite{ren2015faster}, in which the key component is region proposal network (RPN).
        Tripathi \etal modifies the RPN structure to integrate the sketch information in order to create object proposals that are relevant to the query sketch.
        In more detail, feature vectors of different regions in the image feature map are scored using the global sketch representation to determine compatibility.
        The attention feature is then calculated by multiplying these compatibility scores with image feature maps.
        These attention feature maps are concatenated with the original feature maps and projected to a lower-dimensional space, which is then input to RPN to yield relevant object proposals.
        The pooled object proposals are scored using a sketch feature vector to localize the object of interest.
        
        \noindent\textbf{Riba \etal, Sketch-DETR}~\cite{riba2021localizing} is built on the DETR~\cite{carion2020end} architecture.
        Given a natural image and a query sketch, Sketch-DETR~\cite{riba2021localizing} transforms each input with a separate CNN backbone, and generates feature maps for each input modality.
        They are then fused via concatenation.
        Specifically, the sketch is inflated by the resolution of the image feature map then projected using a 1$\times$1 convolution operation.
        The obtained feature map are flattened before being fed into the Transformer encoder-decoder.
        The final bounding boxes and their respective score are predicted through a shared feed-forward network (FFN).

    \vspace{\abovesubsecmargin}\subsection{Vision and Multimodal Transformers}\vspace{\belowsubsecmargin}
        Transformer~\cite{vaswani2017attention} is a universal sequence processor with an attention-based architecture that is originally designed for machine translation.
        The primary components of Transformer are self-attention that captures long-range interactions within a single context and cross-attention that considers token correspondences between two sequences.
        
        \vspace{\paramargin}\paragraph{Vision Transformers.}
        Beyond natural language processing~\cite{brown2020language,devlin2018bert,radford2018improving,radford2019language}, Transformers have rapidly become the {\it de facto} standard in a variety of computer vision applications: image recognition~\cite{dosovitskiy2020image}, object detection~\cite{carion2020end}, panoptic segmentation~\cite{wang2021max}, human object interaction~\cite{kim2021hotr}, action recognition~\cite{woo2022towards,lee2023modality}, and object tracking~\cite{chen2021transformer}.
        Among these, it is worth noting that Detection Transformer (DETR)~\cite{carion2020end} has made a significant breakthrough in the field of object detection by successfully adopting Transformer design and bipartite matching algorithm.
        By design, DETR eliminates the need for heuristics (\eg, non-maximum suppression) in the detection pipeline while leveraging the capability of global relation modeling.
        Inspired by the recent successes of Transformers, particularly DETR, we view the SVOL task as a set prediction problem and build our model on the Transformer architecture.
        Furthermore, since videos are a sequence of frames, the Transformer is well-suited to model temporal information of videos.

        \vspace{\paramargin}\paragraph{Multimodal Transformers.}
        Transformers have shown to be particularly effective in multimodal processing due to their ability to selectively attend to relevant information from multiple modalities (\eg, text, image, and audio).
        This has been demonstrated in various tasks, including image captioning~\cite{radford2021learning}, visual question answering~\cite{chang2022webqa}, natural language video grounding~\cite{woo2022explore}, text-to-image synthesis~\cite{ramesh2021zero}, and text-to-speech~\cite{li2019neural}.
        Recent studies such as CLIP~\cite{radford2021learning} and DALL-E~\cite{ramesh2021zero} highlighted the potential of pre-training Transformer-based models on a vast amount of image-text pairs using a contrastive loss.
        This provides a strong starting point when fine-tuning on downstream tasks, thereby allowing the model to generalize well unseen datasets.
        Furthermore, Transformers have shown to transfer well across tasks, making them a versatile model for various multimodal processing tasks~\cite{lu2019vilbert,li2020oscar}.
        These properties of Transformer make itself a strong candidate for the SVOL task, where the intricate relationship between the query sketch and video objects must be modeled to bridge the gap between natural video and sketches with various styles.
\vspace{\abovesecmargin}
\section{Preliminary: Transformer}
\label{sec:transformer}
\vspace{\belowsecmargin}
We build our Cross-modal Transformer (CMT) on top of Transformer design~\footnote{We leave an original paper as a reference~\cite{vaswani2017attention} for further details of Transformer building blocks.}, which:
(i) densely relates every pair of elements in the input sequence;
(ii) captures long-range context with minimal inductive bias (compared to CNNs or RNNs);
(iii) effectively models interaction between multi-modal (cross-domain) sequences.
These desirable properties of Transformer makes itself well-suited for the CMT design.

The common practice is to use attention mechanism with residual connection, dropout, and layer normalization.
Attention in the general ${\bf QKV}$ form is a popular yet strong mechanism for neural systems.
Given that attention operations are key building blocks of CMT, we first briefly discuss their general form.
    
\vspace{\abovesubsecmargin}\subsection{QKV Attention}\vspace{\belowsubsecmargin}
\label{sec:qkv_attention}
Given input sequences ${\bf q} \in \mathbb{R}^{L_1 \times D}$, ${\bf k} \in \mathbb{R}^{L_2 \times D}$ and ${\bf v} \in \mathbb{R}^{L_2 \times D}$, we project them into separate embedding spaces.
We call the embedded representations as query (${\bf Q}$), key (${\bf K}$), and value (${\bf V}$). 
\eqnsm{Q}{
    {\bf Q} = ({\bf q} + \texttt{pos}_q){{\bf W_q}},
}
\eqnsm{K}{
    {\bf K} = ({\bf k} + \texttt{pos}_k){{\bf W_k}},
}
\eqnsm{V}{
    {\bf V} = {\bf v}{{\bf W_v}},
}
where ${{\bf W_q}}, {{\bf W_k}}, {{\bf W_v}} \in \mathbb{R}^{D \times D_h}$ are learnable weights.
Since the Transformer is inherently permutation-invariant w.r.t input sequence, we add positional encoding $\texttt{pos}_q \in \mathbb{R}^{L_1 \times D}$ and $\texttt{pos}_k \in \mathbb{R}^{L_2 \times D}$ (fixed absolute encoding to represent positions using sine and cosine functions of different frequencies) to embedded sequences.

The attention weights ${\bf A}_{ij}$ are computed by comparing two elements of the sequence (dot products) to their respective query ${\bf Q}_i$ and key ${\bf K}_j$ representations, normalized by $D_h$.
\eqnsm{A}{
    {\bf A} = {\rm softmax}\left(\frac{{\rm\bf QK}^T}{\sqrt{D_h}} \right),
}

Finally, we calculate a weighted sum over all value representation ${\bf V}$.
\eqnsm{Att}{
    {\rm Att}({\bf q}, {\bf k}, {\bf v}) = {\bf A}{\bf V}.
}

We call this operation as \textit{Self-Attention} for the special case where ${\bf q}$, ${\bf k}$, and ${\bf v}$ are all the same. 

\vspace{\abovesubsecmargin}\subsection{Multi-Head Attention}\vspace{\belowsubsecmargin}
Multi-Head Attention (MHA) allows the model to jointly attend to information from different representation subspaces at different positions.
It is a simple extension of Attention in which several Attentions heads are executed in parallel followed by a projection of their concatenated outputs.
To maintain the computed value and the number of parameters constant when changing the number of heads $k$, $D_h$ is typically set to $D/k$.
\eqnalism{MHA}{
    &{\rm MHA({\bf q}, {\bf k}, {\bf v})} = \\
    &[{\rm Att}_1({\bf q}, {\bf k}, {\bf v}); \cdots; {\rm Att}_k({\bf q}, {\bf k}, {\bf v});]{\bf W_{MHA}},
}
where [;] denotes concatenation on the channel axis and ${\bf W_{MHA}} \in \mathbb{R}^{k\cdot D_h \times D}$ is learnable weight.
\begin{figure*}[t!]
    \centering
    \footnotesize
    \includegraphics[width=\linewidth]{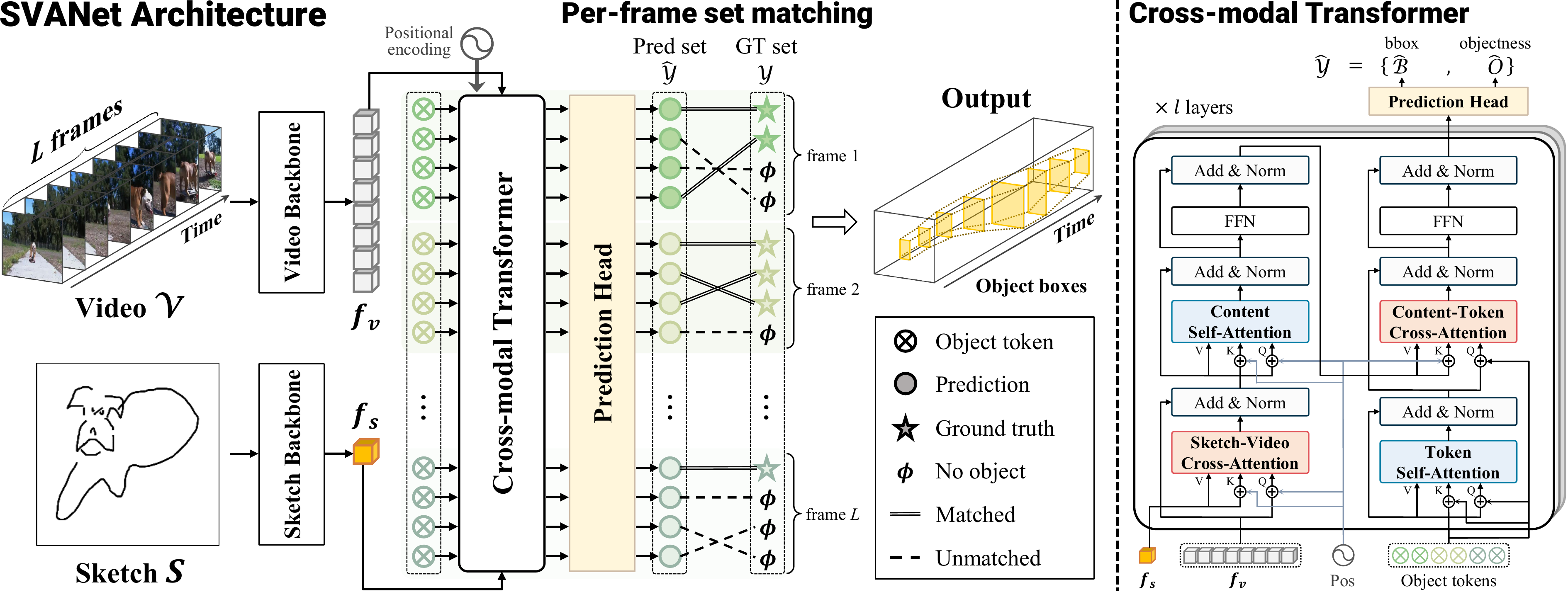}
    \vskip \abovefigcapmargin
    \caption{
        {\bf Overview of \ours.}
        Given a video $\mathcal{V}$ and a query sketch $\mathcal{S}$, \ours processes them in a separate encoding pipeline, yielding a sequence of frame representations $f_v$ and a sketch representation $f_s$.
        The Cross-modal Transformer (CMT) then takes $f_v$, $f_s$, and a set of learnable object tokens as inputs.
        Through the CMT layers, object tokens learn interactions between themselves and attend to sketch-video joint representation to produce accurate predictions.
        \textbf{Per-frame set matching.}
        During training, \ours finds the best matching that minimizes the matching cost (see~\cref{eq:match_cost}) between the prediction set and the ground truth set \textit{for each frame}.
        To assign a unique matching between the two sets, the ground truth set is padded with additional \texttt{No object} ($\phi$) elements.
        The overall loss is defined based on the matching results (see~\cref{eq:set_loss}).
        As a result, \ours outputs the spatio-temporal object boxes.
        \textbf{Cross-modal Transformer.}
        In CMT, sequences of representations are added with positional encoding before every attention operation.
        CMT first highlights important contents by learning the correspondence between sketch $f_s$ and video $f_v$ representations, and models intra-content relationships.
        CMT then transforms the object token set into a set of predictions (box coordinates and objectness scores) by learning token-token interactions and referring to joint sketch-video representations.
        More details are in~\cref{sec:method}.
        Best viewed in color.
    }
    \label{fig:overview}
    \vspace{\belowfigcapmargin}
\end{figure*}

\vspace{\abovesecmargin}
\section{Method}
\label{sec:method}
\vspace{\belowsecmargin}
    We begin by describing the SVOL task and present an end-to-end trainable \ours that predicts a set of objects based on dense pair-wise relation modeling.
    Next, we introduce a per-frame set matching strategy that imposes a unique match between predicted and ground truth boxes at each frame; then, define an overall training loss.
    An overview of \ours is depicted in~\Cref{fig:overview}.
    
    \vspace{\paramargin}\paragraph{SVOL task definition.}
    Given a query sketch $\mathcal{S}$ and a video $\mathcal{V}$, the goal of SVOL is to find all spatio-temporal boxes $\mathcal{Y}$ that \textit{match} the sketch object in the video.
    We consider the video as a sequence of $L$ frames, $\mathcal{V} = [V_i]_{i=1}^{L}$, and aim to find all boxes $\mathcal{Y}=[B_i]_{i=1}^{L}$ over the video frames, where $B_i \in \mathbb{R}^{K_i \times 4}$ is a set of bounding boxes at video frame $V_i$, $B_i = \{b_{i}^{j}\}_{j=1}^{K_i}$.
    The number of boxes $K_i$ at frame $V_i$ can vary throughout the video, since objects can be occluded, disappear or appear in the scene.
    We predict a total of $N$ bounding boxes across $L$ frames, $\hat{\mathcal{B}}=[\hat{B}_i]_{i=1}^{L}$, $M$ boxes per frame, $\hat{B}_i = \{b_{i}^{j}\}_{j=1}^{M}$, where $N = L \times M$.
    The predictions are considered as \textit{correct} if IoU between the predicted box $\hat{b}$ and the ground truth box $b$ is higher than the threshold $\mu$.
    The bounding box is defined as a 4D vector normalized \wrt the frame resolution: $b \in [0, 1]^4$.  
    We also predict the likelihoods that the predicted boxes contain the target object, referred to as objectness scores $\hat{\mathcal{O}}$, where each element $\hat{o} \in [0, 1]$.
    In short, the predictions are a set of bounding boxes and their corresponding objectness scores: $\hat{\mathcal{Y}}=\{\hat{\mathcal{B}}, \hat{\mathcal{O}}\}$.
    As we view SVOL as a set prediction problem, we find the best matching between the ground truth set $\mathcal{Y}$ and the prediction set $\hat{\mathcal{Y}}$.
    
    As we set the SVOL problem as category-level localization, the system is trained to perform the consistent bounding box localization for sketches belonging to the same category, regardless of variations in shape or pose.
    This allows the system to operate robustly, even in the presence of different levels of abstraction or diverse styles in the sketches.
    However, it is worth noting that there is \textit{no} explicit category prediction inside the system, instead it relies on implicit similarity (\eg, symbolic meaning, appearance, \etc) to learn sketch-video object matching.

    \vspace{\abovesubsecmargin}\subsection{\ours Architecture}
    \ours is designed to address the challenge of bridging the gap between two distinct modalities, sketch and natural video, in order to perform object localization.
    The system incorporates attention operations that consider a wide range of contexts and inter-dependencies between elements within the input sequences. 
    This leads our system to acquire the capability to learn powerful representations of the input sequences and delivers accurate video object localization using sketches as queries.
            
        \vspace{\paramargin}\paragraph{Video \& sketch backbones.}
        A video, represented by a sequence of frames, $\mathcal{V} \in \mathbb{R}^{L \times C_0 \times H_0 \times W_0}$, where $L=32, C_0=3, H_0=W_0=224$, is initially processed using the ResNet-50 architecture~\cite{he2016deep}, generating high-dimensional feature maps $f_v \in \mathbb{R}^{L \times C \times H \times W}$, where $C=512, H=W=7$.
        Likewise, a sketch $\mathcal{S}$ is processed using ResNet-18~\cite{he2016deep}, followed by a spatial pooling operation that compresses it into 1D representation $f_s \in \mathbb{R}^{C}$.
        Finally, the outputs $f_v$ and $f_s$ are passed through the Cross-modal Transformer.
        To address the sparse nature of sketch information, we use a lighter CNN backbone (ResNet-18) compared to the video (ResNet-50).
        
        \vspace{\paramargin}\paragraph{Cross-modal Transformer \& prediction head.}
        In addition to $f_v$ and $f_s$, the Cross-modal Transformer (CMT)\footnote{More details about original Transformer are in~\cref{sec:transformer}.} takes a set of $N$ learnable embeddings initialized with random weights, which we refer to as \textit{object tokens}, and transforms them into a set of $N$ predictions.
        
        CMT consists of $l$ layers, and each layer contains four attention operations:
        \textit{(i) Sketch-Video Cross-Attention} (\textbf{SVCA}) assigns higher attention weights to the important elements of the input sequence (video patches), that are relevant for accurate bounding box localization based on the input sketch query.
        This is achieved by modeling the inter-modality relationship between video $f_v$ and sketch $f_s$ representations. 
        SVCA bridges the gap between sketches and videos by effectively integrating the information from both modalities.
        \textit{(ii) Content Self-Attention} (\textbf{CSA}) is responsible for modeling the temporal relationship between the elements in the input sequence (\ie, output of SVCA).
        By considering the pair-wise relationship of these elements, CSA enables a more comprehensive understanding of the broader video context.
        \textit{(iii) Token Self-Attention} (\textbf{TSA}) receives object tokens as input and models interactions between them, enabling them to globally reason about all objects.
        \textit{(iv) Content-Token Cross-Attention} (\textbf{CTCA}) transforms object tokens to meaningful outputs by relating them with contextual representation of content (\ie, output of CSA).
        Since the attention operations are permutation-invariant (\ie, produce the same output regardless of the order of elements in the input sequence), we supplement the input sequence with temporal order information by adding absolute positional encoding prior to every attention operation (except for TSA; we instead add object tokens to the input sequence of each TSA operation).
    
        All attention operations in CMT are in the form of MHA (\cref{eq:MHA}) with 8 heads (\ie, $k=8$).
        Let a video representation $f_v \in \mathbb{R}^{L \times C \times H \times W}$ as ${\bf v}^{(0)}$ and a sketch representation $f_s \in \mathbb{R}^{1 \times C}$ as ${\bf s}$.
        Given ${\bf v}^{(0)}$ and ${\bf s}$ the $i$-th CMT layer calculates:
        \eqnsm{SVCA}{
            {\bf x}^{(i)} = {\rm LN}({\bf SVCA}^{(i)}({\bf v}^{(i)}, {\bf s}, {\bf s}) + {\bf v}^{(i)}),
        }
        \eqnsm{CSA}{
            {\bf y}^{(i)} = {\rm LN}({\bf CSA}^{(i)}({\bf x}^{(i)}, {\bf x}^{(i)}, {\bf x}^{(i)}) + {\bf x}^{(i)}),
        }
        \eqnsm{FFN1}{
            {\bf v}^{(i+1)} = {\rm LN}({\rm FFN}_{1}^{(i)}({\bf y}^{(i)}) + {\bf y}^{(i)}),
        }
        \eqnsm{TSA}{
            {\bf p}^{(i)} = {\rm LN}({\bf TSA}^{(i)}({\bf r}^{(i)}, {\bf r}^{(i)}, {\bf r}^{(i)}) + {\bf r}^{(i)}),
        }
        \eqnsm{CTCA}{
            {\bf q}^{(i)} = {\rm LN}({\bf CTCA}^{(i)}({\bf p}^{(i)}, {\bf v}^{(i+1)}, {\bf v}^{(i+1)}) + {\bf p}^{(i)}),
        }
        \eqnsm{FFN2}{
            {\bf r}^{(i+1)} = {\rm LN}({\rm FFN}_{2}^{(i)}({\bf q}^{(i)}) + {\bf q}^{(i)}),
        }
        where LN is layer normalization and FFN is 2-layer feed-forward network.
        Here, $r^{(0)} = O_{N \times C}$ ($N \times C$-sized zero matrix), thus TSA operation~\cref{eq:TSA} can be omitted in the first CMT layer.

        TSA (\cref{eq:TSA}) and CTCA operations (\cref{eq:CTCA}) slightly differ with the standard QKV attention in that they consider the object tokens \texttt{tkn} as learnable positional encoding for the query ({\bf q}) inputs, \ie, ${\bf q}$ is added with $\texttt{tkn}$ instead of $\texttt{pos}_q$ in~\cref{eq:Q}.
        \eqnsm{Q2}{
            {\bf Q} = ({\bf q} + \texttt{tkn}){{\bf W_q}}.
        }
        In addition, since TSA is Self-Attention operation ({\bf q} = {\bf k} = {\bf v} = ${\bf r}^{(i)}$), \texttt{tkn} is also used as positional encoding for the key ({\bf k}) input in TSA.
        \eqnsm{K2}{
            {\bf K} = ({\bf k} + \texttt{tkn}){{\bf W_k}}.
        }
        The subsequent processes are the same as standard QKV attention.

        We go through $l$ CMT layers, and the final CMT output $r^{(l)}$ is fed into two separate linear layers (\ie, prediction heads) to obtain a set of bounding box coordinates $\hat{\mathcal{B}}$ and objectness scores $\hat{\mathcal{O}}$, respectively.

    \vspace{\abovesubsecmargin}\subsection{SVOL as a Set Prediction}\vspace{\belowsubsecmargin}
        In this work, we formulate SVOL as a set prediction problem.
        In practice, we adopt a Hungarian algorithm~\cite{kuhn1955hungarian} to find an optimal matching between predictions and ground truths in a way that minimizes the matching cost.
        The overall loss function is defined based on the matching results.
        
        \vspace{\paramargin}\paragraph{Per-frame set matching.}
        \ours transforms $N$ object tokens to $N$ predictions (bounding boxes and objectness scores).
        Here, we make each of the $N/L$ (hereafter $M$) tokens to be responsible for predicting the results of each frame $V_i$ by performing per-frame set matching.
        This allows \ours to predict results per frame while being able to access global context information across the video.
        We formally describe the process in the following.
        
        A set of ground truth bounding boxes $\mathcal{Y}$ can be seen as a sequence of $L$ subsets, where $i$-th subset has $K_i$ elements: $[\{b_i^j\}_{j=1}^{K_i}]_{i=1}^{L}$.
        Likewise, we evenly divide a prediction set $\hat{\mathcal{Y}} = \{\hat{\mathcal{B}}, \hat{\mathcal{O}}\}$ of size $N$ into $L$ subsets having $M$ elements each: $[\{\hat{y}_i^j\}_{j=1}^{M}]_{i=1}^{L}$, where $\hat{y}_i^j = (\hat{b}_i^j, \hat{o}_i^j)$.
        Hereafter, we denote the $i$-th subset of ground truths as $Y_i$ and that of predictions as $\hat{Y}_i$ for conciseness.
        The size of $\hat{Y}_i$ is assumed to be larger than the size of $Y_i$: $M > K_i$.
        Since the Hungarian algorithm pairs the elements of two sets one by one, we pad $Y_i$ with \texttt{No object} ($\noobject$) to match the size of $M$.
        For every single $i$ (from $i=1$ to $i=L$), we seek for the best one-to-one matching between $Y_i$ and $\hat{Y}_i$ using a Hungarian algorithm.
        Formally, in the $i$-th prediction subset, let $\hat{y}_{i}^{\sigma_i(j)}$ be the $j$-th element under a permutation of $M$ elements $\sigma_i \in \Sigma_i(M)$.
        We now define the pair-wise matching cost $\mathcal{C}$ as:
        \eqnsm{match_cost}{
            \mathcal{C}(b_i^j, \hat{y}_{i}^{\sigma_i(j)}) = -\indic{b_i^j\neq\noobject}\hat o_i^{\sigma_i(j)} + \indic{b_i^j\neq\noobject} \mathcal{L}_{\rm box}(b_i^j, \hat b_i^{\sigma_i(j)}) \,.
        }
        Note that the \texttt{No object} paddings in the ground truth are not considered when calculating the matching cost.
        For every $i$, we aim to find the optimal assignment $\sigma_{i}^{*} \in \Sigma_i(M)$ that pairs the predictions and ground truths at the lowest cost:
        \eqnsm{optimal_match}{
            \sigma_{i}^{*} = \argmin_{\sigma_i\in\Sigma_i(M)} \sum_{j=1}^{M} \mathcal{C}(b_i^j, \hat y_i^{\sigma_i(j)}) \,.
        }
        \vspace{\paramargin}\paragraph{Overall loss.}
        Based on the matching results, our set prediction loss $\mathcal{L}_{set}(\mathcal{Y}, \hat{\mathcal{Y}})$ is defined as:
        \eqnsm{set_loss}{
            \sum_{i=1}^L \sum_{j=1}^M \left[-\lambda_{\rm obj} \log \hat{o}_i^{\sigma_{i}^{*}(j)} + \indic{b_i^j\neq\noobject} \mathcal{L}_{\rm box}(b_i^j, \hat b_i^{\sigma_{i}^{*}(j)})\right]\,,
        }
        where $\lambda_{\rm obj} \in \mathbb{R}$ is a loss coefficient for objectness scores.
        Here, the log-probability of the \texttt{No object} paddings ($\noobject$) is scaled down by a factor of 10 to strike a balance between object and no-object.
        
        The box loss $\mathcal{L}_{\rm box}$ is defined as a linear combination of $\ell_1$ loss and the generalized IoU (gIoU) loss~\cite{rezatofighi2019generalized}:
        \eqnsm{box_loss}{
            \mathcal{L}_{box}(b_i^j, \hat b_i^{\sigma_i(j)}) = \lambda_{\ell_1}\mathcal{L}_{\ell_1}(b_i^j, \hat b_i^{\sigma_i(j)}) + \lambda_{\rm iou}\mathcal{L}_{iou}(b_i^j, \hat b_i^{\sigma_i(j)})\,,
        }
        where $\lambda_{\ell_1} \in \mathbb{R}$ and $\lambda_{\rm iou} \in \mathbb{R}$ are balancing hyperparameters.
        While both losses have the same goal, object localization, the $\ell_1$ loss will have different scales for small and large boxes even if their relative errors are similar, whereas the gIoU loss is scale-invariant.
        
        We calculate the $\ell_1$ loss as:
        \eqnsm{l1_loss}{
            \mathcal{L}_{\ell_1}(b_i^j, \hat b_i^{\sigma(j)}) = ||b_i^j - \hat b_i^{\sigma(i)}||_1\,.
        }
        The gIoU loss is calculated as (we denote the area with set operations for the sake of argument):
        \eqnalism{iou_loss}{
            &\mathcal{L}_{iou}(b_i^j, \hat b_i^{\sigma(j)}) = \\
            &1 - \bigg(\frac{ |b_i^j| \cap |\hat b_i^{\sigma(j)}|}{|b_i^j| \cup |\hat b_i^{\sigma(j)}|} - \frac{|B(b_i^j,\hat b_i^{\sigma(j)})|  \setminus |b_i^j| \cup |\hat b_i^{\sigma(j)}|}{|B(b_i^j,\hat b_i^{\sigma(j)})|}\bigg)\,,
        }
        where $|.|$ represents the bounding box area, and the symbols $\cup$, $\cap$, and $\setminus$ calculate the area of union, intersection, and subtraction of the two bounding box areas, respectively.
        $B(b_{i}, \hat b_{\sigma(i)})$ denotes the smallest box enclosing $b_{i}$ and $\hat b_{\sigma(i)}$.
        The areas are computed by taking the minimum or maximum value of the linear functions of the box coordinates.
\vspace{\abovesecmargin}
\section{Experiments}
\label{sec:experiments}
\vspace{\belowsecmargin}
The SVOL dataset\footnote{Details about datasets, curation process, and statistics are in~\cref{sec:svol_dataset}.} is curated upon the ImageNet-VID dataset~\cite{russakovsky2015imagenet} and three different sketch datasets with varying levels of abstraction: {\sketchy}~\cite{eitz2012humans} (least abstract), {\tuberlin}~\cite{jongejan2016quick}, and {\quickdraw}~\cite{sangkloy2016sketchy} (most abstract) (see~\cref{fig:dataset}).

\vspace{\abovesubsecmargin}\subsection{Implementation Details}\vspace{\belowsubsecmargin}
We uniformly sampled 32 frames from a video ($L=32$), scaled them to 224$\times$224 dimensions, and used them as an input $\mathcal{V} \in \mathbb{R}^{32 \times 3 \times 224 \times 224}$ (3 for RGB channels).
Likewise, a sketch is rescaled to $224 \times 224$ size, and used as an input $\mathcal{S} \in \mathbb{R}^{224 \times 224}$.
The number of CMT layers is set to two (\ie, $l=2$), and we used 10 object tokens per frame ($M=10$), a total of 320 object tokens ($N=320$).
We adopt ResNet-50 and ResNet-18~\cite{he2016deep} pre-trained on ImageNet~\cite{russakovsky2015imagenet} as our video and sketch backbone, respectively.

Due to excessive number of video-sketch pairs, we used an iteration-based batch sampler and randomly sampled a subset from all possible pairings for training.
\ours is trained using AdamW optimizer~\cite{loshchilov2017decoupled} with an initial learning rate of ${10}^{-4}$ and weight decay of ${10}^{-4}$ for a batch size of 16.
The overall loss weights $\lambda_{\rm L1}:\lambda_{\rm giou}:\lambda_{\rm cls}$ were set to $5:1:2$ throughout training.
We set different learning schedules (number of iterations and learning rate decay steps) for each sketch dataset as below since their sizes are different.
\begin{table}[h!]
    \vspace{-3mm}
    \label{tab:implementation_details}
	\begin{center}
	\tablestyle{1pt}{1.05}
	\resizebox{\linewidth}{!}{
    	\begin{tabular}{x{60}|x{50}x{50}x{50}}
    		{Settings}&\textbf{\sketchy}&\textbf{\tuberlin}&\textbf{\quickdraw}\\\shline
            {\# pairs (train)}&{1,545,801}&{215,040}&{2,958,400}\\ 
    		{Iterations}&{50,000}&{20,000}&{100,000}\\ 
    	    {LR decay step$^{*}$}&{30,000}&{6,000}&{30,000}\\
            \multicolumn{4}{c}{*LR is linearly decayed by a factor of 10 at every LR decay step.}
    	\end{tabular}
    }
	\end{center}
    \vspace{-8mm}
\end{table}

\vspace{\abovesubsecmargin}\subsection{Experimental Setups}

\vspace{\paramargin}\paragraph{Evaluation metrics.}
We adopt two evaluation metrics for SVOL: 
1) \({\bf R^{k}_{\mu}}\) denotes the percentage of samples that have at least one correct result in top-\(k\) retrieved results, \ie, Recall@k, where the correct results indicate that IoU with ground truth is larger than the threshold $\mu$. (we specifically use \(k=1, 5\) and \(\mu=0.5, 0.7\));
2) {\bf mIoU} averages the IoU between predicted boxes and ground truth boxes over all testing samples to compare the overall performance.

\begin{figure}[t!]
    \centering
    \resizebox*{\linewidth}{!}{
        \includegraphics{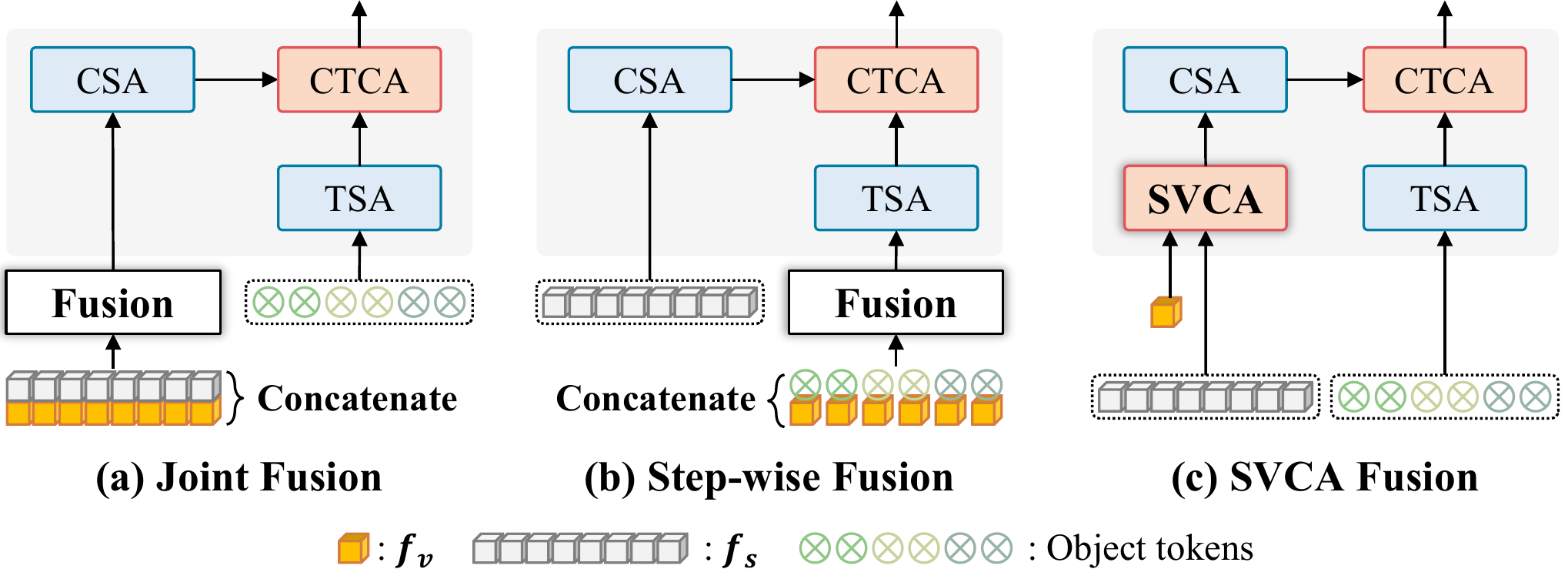}
    }
    \vskip \abovefigcapmargin
    \caption{
        \textbf{Three instantiations of sketch-video fusion} contextualize video and sketch information in different ways:
        \textbf{(a) joint fusion}: copy $f_s$ by the size of $f_v$, concatenate, and fuse them via MLP.
        \textbf{(b) step-wise fusion}: copy $f_s$ by the number of object tokens, concatenate, and fuse them via MLP. The sketch and video representations are later fused through CTCA.
        \textbf{(c) SVCA fusion (ours)}: fuse $f_s$ and $f_v$ with SVCA.
    }
    \label{fig:fusion_instantiations}
    \vspace{\belowfigcapmargin}
\end{figure}

\vspace{\paramargin}\paragraph{Baselines.}
We set image-level sketch object localization approaches~\cite{tripathi2020sketch,riba2021localizing} as the SVOL baselines.
We find significant room for improvement as they were designed to be conditioned on a single frame rather than an entire video sequence. 
In addition, we present several instantiations of sketch-video fusion on \ours, as shown in~\cref{fig:fusion_instantiations}, and compare them with our final model.

\begin{table*}[t!]
    \begin{center}
    \resizebox{\linewidth}{!}{
    \tablestyle{1pt}{1.2} 
    \begin{tabular}{l|x{21}x{21}x{21}x{21}x{21}|x{21}x{21}x{21}x{21}x{21}|x{21}x{21}x{21}x{21}x{21}|x{21}x{21}x{21}x{21}x{21}}
        {}&\multicolumn{5}{c|}{\bf \sketchy}&\multicolumn{5}{c|}{\bf \tuberlin}&\multicolumn{5}{c|}{\bf \quickdraw}&\multicolumn{5}{c}{\bf ALL ({\color{sketchy}S} $\cup$ {\color{tuberlin}T} $\cup$ {\color{quickdraw}Q})}\\
        {\bf Method} (backbone)&$\rm{R}^{1}_{0.5}$&$\rm{R}^{1}_{0.7}$&$\rm{R}^{5}_{0.5}$&$\rm{R}^{5}_{0.7}$&mIoU&$\rm{R}^{1}_{0.5}$&$\rm{R}^{1}_{0.7}$&$\rm{R}^{5}_{0.5}$&$\rm{R}^{5}_{0.7}$&mIoU&$\rm{R}^{1}_{0.5}$&$\rm{R}^{1}_{0.7}$&$\rm{R}^{5}_{0.5}$&$\rm{R}^{5}_{0.7}$&mIoU&$\rm{R}^{1}_{0.5}$&$\rm{R}^{1}_{0.7}$&$\rm{R}^{5}_{0.5}$&$\rm{R}^{5}_{0.7}$&mIoU\\\hline
        CMA$^\dagger$~\cite{tripathi2020sketch} (R50)&{23.18}&{14.89}&{39.76}&{21.29}&{19.76}&{20.25}&{14.55}&{36.87}&{20.80}&{18.64}&{22.69}&{15.53}&{39.86}&{20.80}&{21.52}&{18.89}&{11.41}&{33.53}&{17.23}&{16.14}\\
        Sketch-DETR$^\dagger$~\cite{riba2021localizing} (R50)&{28.78}&{18.56}&{46.65}&{26.50}&{26.09}&{30.75}&{18.97}&{47.76}&{27.54}&{26.24}&{31.10}&{19.47}&{49.39}&{31.05}&{28.59}&{28.23}&{16.74}&{44.21}&{25.54}&{24.30}\\\hline
        \ours/joint (R50) &{33.86}&{22.56}&{52.84}&{30.57}&{31.46}&{31.14}&{19.48}&{50.17}&{28.21}&{29.38}&{33.95}&{20.12}&{54.77}&{34.11}&{31.89}&{29.53}&{16.50}&{47.90}&{27.18}&{28.59}\\
        \ours/step-wise (R50) &{33.31}&{22.81}&{53.00}&{31.07}&{30.29}&{30.23}&{18.19}&{50.66}&{27.20}&{30.41}&{32.05}&{21.17}&{56.34}&{35.39}&{31.98}&{29.64}&{17.61}&{48.21}&{26.29}&{27.99}\\\hline
        \ours (S3D~\cite{xie2018rethinking})&{32.89}&{21.11}&{48.08}&{27.07}&{30.83}&{29.43}&{18.25}&{45.72}&{24.34}&{28.00}&{30.86}&{19.24}&{46.80}&{25.48}&{29.37}&{27.88}&{17.26}&{43.33}&{22.86}&{27.87}\\
        \rowcolor{Gray}
        \ours (R50~\cite{he2016deep})&{\bf 35.60}&{\bf 23.19}&{\bf 54.06}&{\bf 32.95}&{\bf 33.76}&{\bf 32.10}&{\bf 19.60}&{\bf 51.61}&{\bf 30.94}&{\bf 30.89}&{\bf 34.47}&{\bf 22.30}&{\bf 58.13}&{\bf 37.88}&{\bf 33.40}&{\bf 31.80}&{\bf 18.52}&{\bf 51.44}&{\bf 29.90}&{\bf 30.64}\\
    \end{tabular}
    }
    \end{center}
    \vspace{\abovetabcapmargin}
    \vspace{-2mm}
    \caption{
        \textbf{Comparison of \ours with baselines.}
        \ours significantly outperforms baselines on three sketch datasets and on combined dataset (\textbf{ALL}), where we use only overlapping categories between three datasets. 
        $\dagger$ indicates the re-implementation based on our settings.
    }
    \label{tab:benchmark}
    \vspace{-2mm}
\end{table*}

\vspace{\abovesubsecmargin}\subsection{Comparative Study}\vspace{\belowsubsecmargin} 

    We benchmark the model performance on the SVOL task using three different sketch datasets.
    The results are shown in~\Cref{tab:benchmark}.
    Since image-level baselines (CMA, Sketch-DETR) make predictions at each frame, they neglect the global video context.
    In contrast, \ours not only considers spatial context but also effectively models temporal information of video, thereby outperforming them by a significant margin in all metrics across three sketch datasets.
    Especially, \ours improves mIoU by 29.4\%, 17.7\%, and 16.8\% over Sketch-DETR on {\sketchy}, {\tuberlin}, and {\quickdraw}.
    Also, our final \ours yields the best results among the several model variants (joint, step-wise), implying the effectiveness of our attention-based fusion.
    Moreover, we use all three sketch datasets as a single set of query sketches (denoted as ALL in~\Cref{tab:benchmark}) to see how the model performs when the same category contains sketches of different styles.
    Overall, model performances are diminished as a result of a greater diversity of sketch samples.
    However, \ours shows only 0.25\%p mIoU drop compared to the results on {\tuberlin}, which means that \ours is quite robust to sketch style variations. 
    Lastly, we compare two backbones for video encoding: 3D CNN (S3D~\cite{xie2018rethinking}) \vs 2D CNN (ResNet50~\cite{he2016deep}).
    We expected the more sophisticated 3D CNN to work better, but 2D CNN outperformed 3D CNN.
    This shows that CMT can sufficiently complement the temporal modeling capability of 3D CNN.

\begin{table*}[t!]
    \centering
    \resizebox{\linewidth}{!}{
    \subfloat[The models are trained with {\sketchy} dataset and evaluated on {\quickdraw} or {\tuberlin} dataset. To solely see the effect of sketch style differences, we use the same video samples in both training and evaluation, and overlapping categories between the two sketch datasets.
    \label{tab:dataset-transfer}]{
        \hspace{-8pt}
        \tablestyle{1pt}{1.2} 
        \begin{tabular}{l|x{20}x{20}x{20}x{20}x{20}|x{20}x{20}x{20}x{20}x{20}} 
            \multirow{2}{*}{\bf Method}&\multicolumn{5}{c|}{\bf {\sketchy}\(\rightarrow\){\tuberlin}}&\multicolumn{5}{c}{\bf {\sketchy}\(\rightarrow\){\quickdraw}}\\
            {}&\(\rm {R}^{1}_{0.5}\)&\(\rm {R}^{1}_{0.7}\)&\(\rm {R}^{5}_{0.5}\)&\(\rm {R}^{5}_{0.7}\)&mIoU&\(\rm {R}^{1}_{0.5}\)&\(\rm {R}^{1}_{0.7}\)&\(\rm {R}^{5}_{0.5}\)&\(\rm {R}^{5}_{0.7}\)&mIoU\\\hline
            CMA\(^\dagger\)~\cite{tripathi2020sketch}&{38.03}&{26.20}&{49.70}&{39.07}&{30.97}&{40.06}&{29.89}&{49.83}&{37.69}&{32.45}\\
            Sketch-DETR\(^\dagger\)~\cite{riba2021localizing}&{43.49}&{32.25}&{51.71}&{44.39}&{36.76}&{46.02}&{37.87}&{59.50}&{45.34}&{40.21}\\\hline
            \rowcolor{Gray}
            \ours (Ours)&{\bf 54.74}&{\bf 46.67}&{\bf 69.90}&{\bf 56.57}&{\bf 49.01}&{\bf 55.03}&{\bf 47.56}&{\bf 72.04}&{\bf 58.87}&{\bf 49.74}\\
    \end{tabular}}
    \hspace{3mm}
    \subfloat[The models are trained on 14 categories of the {\sketchy} dataset and evaluated on the remaining 5 categories: aircraft, bear, cat, cow, and dog.
    \label{tab:category-transfer}]{
        \hspace{-8pt}
        \tablestyle{1pt}{1.2} 
        \begin{tabular}{l|x{22}x{22}x{22}x{22}x{22}} 
            \multirow{2}{*}{\bf Method}&\multicolumn{5}{c}{\bf Seen \(\rightarrow\)Unseen categories}\\
            {}&\(\rm {R}^{1}_{0.5}\)&\(\rm {R}^{1}_{0.7}\)&\(\rm {R}^{5}_{0.5}\)&\(\rm {R}^{5}_{0.7}\)&mIoU\\\hline
            CMA\(^\dagger\)~\cite{tripathi2020sketch}&{18.16}&{6.89}&{25.52}&{9.78}&{13.37}\\
            Sketch-DETR\(^\dagger\)~\cite{riba2021localizing}&{25.13}&{13.20}&{34.87}&{18.45}&{22.51}\\\hline
            \rowcolor{Gray}
            \ours (Ours)&{\bf 30.13}&{\bf 18.58}&{\bf 41.18}&{\bf 25.51}&{\bf 29.89}\\
    \end{tabular}}
    }
    \vspace{\abovetabcapmargin}
    \vspace{-3mm}
    \caption{
        \textbf{Transfer evaluation} on~\protect\subref{tab:dataset-transfer}~unseen datasets and~\protect\subref{tab:category-transfer}~unseen categories.
    }
    \label{tab:transfer}
    \vspace{\belowtabcapmargin}
\end{table*}
   
\vspace{\abovesubsecmargin}\subsection{Transfer Evaluation}\vspace{\belowsubsecmargin}
    The prediction space of our \ours is not limited to a fixed set of categories.
    By design, it is possible to match even an unseen sketch to the most similar object by comparing feature-level similarity.
    For SVOL system to be more practical in real-world applications, they should be able to operate well even with sketches of various shapes and styles.
    In addition, there should be no constraint that operate only for limited categories, such as object detectors.
    To this end, we devise two transfer tasks to evaluate the generalization capability of the SVOL systems in two aspects: \textbf{(i)} dataset-level transfer and \textbf{(ii)} category-level transfer.

    Formally, we define the transfer evaluation setup as follows.
    Let $\mathcal{V}$, $\{\mathcal{S}_A, \mathcal{S}_B\}$, and $\{\mathbb{C}_A, \mathbb{C}_B\}$ be a video dataset, sketch datasets, and sets of categories in which $\mathcal{S}_A$ and $\mathcal{S}_B$ overlap with $\mathcal{V}$, respectively.
    For dataset-level transfer task, we train the SVOL model on $\mathcal{V}$ and $\mathcal{S}_A$, and evaluate on $\mathcal{V}$ and $\mathcal{S}_B$, only for categories $\mathbb{C}_A \cap \mathbb{C}_B$.
    For category-level transfer task, we first split a sketch dataset $\mathbb{S}_A$ into two subsets: $\mathcal{S}_A^{1}$ and $\mathcal{S}_A^{2}$, where they are mutually exclusive \wrt categories, \ie, $\mathbb{C}_A^{1} \cap \mathbb{C}_A^{2} = \noobject$.
    Then, we train the SVOL model on $\mathcal{V}$ and $\mathcal{S}_A^{1}$, and evaluate on $\mathcal{V}$ and $\mathcal{S}_A^{2}$.
    We note that there can be more variations to evaluate the transferability of the SVOL system.

    \vspace{\paramargin}\paragraph{Transfer to unseen dataset.}
    We study the transferability of the SVOL models across the sketch datasets with style differences (\eg, line thickness, abstraction degree, \etc.).
    The models are trained with {\sketchy} dataset and evaluated on {\quickdraw} or {\tuberlin} dataset.
    To solely examine the transferability on unseen datasets, we use the same video samples in both training and evaluation, and overlapping categories between the two sketch datasets.
    The results are shown in~\Cref{tab:dataset-transfer}.
    The overall transfer performances across the datasets is much higher than the performance of models that are solely trained on dataset itself (\Cref{tab:benchmark}), since transfer settings use the same video set as in training.
    We observe that \ours significantly outperforms baselines in dataset-level transfer, indicating that it effectively learns class-discriminative features independent of sketch style differences.
    Meanwhile, we expected transfer to {\tuberlin} to show better results than transfer to {\quickdraw} as {\tuberlin} appears to be closer to {\sketchy} than {\quickdraw} in terms of visual similarity.
    Contrary to our expectation, transfer to {\quickdraw} shows better results than transfer to {\tuberlin}.
    We understand this is because the system constructs categorical embedding space by matching the query sketch and video objects based on the key features (\eg, cat's whiskers, rabbit's ears, \etc), rather than merely comparing their overall visual appearance.
    The results suggest that implicit similarity such as symbolic representation of sketches are more important for accurate object localization than explicit similarity such as line thickness or proportion.
    
    \vspace{\paramargin}\paragraph{Transfer to unseen categories.}
    In~\Cref{tab:category-transfer}, we evaluate transferability of SVOL models at the category-level.
    We use 14 categories of the {\sketchy} dataset for training, and the remaining 5 categories (aircraft, bear, cat, cow, dog) for evaluation.
    Compared to the results in~\Cref{tab:benchmark}, we observe that \ours degrades 3.87\%p in mIoU since it has never learned which video objects to match the query sketch with.
    Despite this, \ours outperforms the baselines when evaluated on unseen categories, implying that \ours has learned more generalizable representations that can reason about the implicit similarities between sketches and video objects.
    This enables \ours to closely embed sketches of the same category in the feature space.

\begin{table*}[t!]
    \centering
    \resizebox{\linewidth}{!}{
    \subfloat[\textbf{CMT attention operations.} default: SVCA + CTCA.
    \label{tab:ablation_cmt}]{
        \begin{minipage}{0.35\linewidth}{\begin{center}
        \hspace{-8pt}
        \tablestyle{1pt}{1.2} 
        \begin{tabular}{x{15}x{15}x{15}|x{21}x{21}x{21}x{21}x{21}}
            def.&CSA&TSA&\(\rm {R}^{1}_{0.5}\)&\(\rm {R}^{1}_{0.7}\)&\(\rm {R}^{5}_{0.5}\)&\(\rm {R}^{5}_{0.7}\)&mIoU\\\shline
            \cmark&      &      &{32.98}&{19.76}&{51.74}&{31.19}&{30.72}\\ 
            \cmark&\cmark&      &{34.39}&{21.85}&{53.69}&{32.28}&{32.52}\\ 
            \cmark&      &\cmark&{33.83}&{20.89}&{52.29}&{31.73}&{31.69}\\ 
            \rowcolor{Gray}
            \cmark&\cmark&\cmark&{\bf 35.60}&{\bf 23.19}&{\bf 54.06}&{\bf 32.95}&{\bf 33.76}\\
    \end{tabular}
    \end{center}}\end{minipage}}
    \hspace{2mm}
    \subfloat[\textbf{CMT depth.}
    \label{tab:ablation_layers} ]{
        \begin{minipage}{0.29\linewidth}{\begin{center}
        \tablestyle{1pt}{1.2} 
        \hspace{-8pt}
        \begin{tabular}{x{20}|x{21}x{21}x{21}x{21}x{21}}
            {layers}&\(\rm {R}^{1}_{0.5}\)&\(\rm {R}^{1}_{0.7}\)&\(\rm {R}^{5}_{0.5}\)&\(\rm {R}^{5}_{0.7}\)&mIoU\\\shline
            1&{33.53}&{19.95}&{47.05}&{24.17}&{32.30}\\
            \rowcolor{Gray}
            2&{\bf 35.60}&{\bf 23.19}&{54.06}&{32.95}&{\bf 33.76}\\
            3&{35.14}&{20.18}&{56.29}&{32.99}&{32.77}\\
          4&{35.20}&{22.77}&{\bf 58.34}&{\bf 37.39}&{33.28}\\
    \end{tabular}
    \end{center}}\end{minipage}}
    \hspace{2mm}
    \subfloat[\textbf{Per-frame set matching (pfsm).} 
    \label{tab:ablation_matching} ]{
        \begin{minipage}{0.29\linewidth}{\begin{center}
        \tablestyle{1pt}{1.2} 
        \hspace{-8pt}
        \begin{tabular}{x{20}|x{21}x{21}x{21}x{21}x{21}}
            {pfsm}&\(\rm {R}^{1}_{0.5}\)&\(\rm {R}^{1}_{0.7}\)&\(\rm {R}^{5}_{0.5}\)&\(\rm {R}^{5}_{0.7}\)&mIoU\\\shline
            \xmark&{18.36}&{6.78}&{33.54}&{13.64}&{22.34}\\
            \rowcolor{Gray}
            \cmark&{\bf 35.60}&{\bf 23.19}&{\bf 54.06}&{\bf 32.95}&{\bf 33.76}\\
            $\triangle$&{~\pacc{$+$17.24}}&{~\pacc{$+$16.41}}&{~\pacc{$+$20.52}}&{~\pacc{$+$19.31}}&{~\pacc{$+$11.42}}\\
            \multicolumn{6}{c}{$\triangle$: performance gain.}\\
    \end{tabular}
    \end{center}}\end{minipage}}
    }
    \resizebox{\linewidth}{!}{
    \subfloat[\textbf{Object tokens per frame.}
    \label{tab:ablation_tokens}]{
        \begin{minipage}{0.29\linewidth}{\begin{center}
        \hspace{-8pt}
        \tablestyle{1pt}{1.2} 
        \begin{tabular}{x{20}|x{21}x{21}x{21}x{21}x{21}}
            \multicolumn{1}{c|}{tokens}&\(\rm {R}^{1}_{0.5}\)&\(\rm {R}^{1}_{0.7}\)&\(\rm {R}^{5}_{0.5}\)&\(\rm {R}^{5}_{0.7}\)&mIoU\\\shline
            5&{34.53}&{20.50}&{47.61}&{25.21}&{32.34}\\ 
            \rowcolor{Gray}
            10&{\bf 35.60}&{\bf 23.19}&{54.06}&{\bf 32.95}&{\bf 33.76}\\
            15&{33.89}&{22.98}&{\bf 54.98}&{32.64}&{32.69}\\
    \end{tabular}
    \end{center}}\end{minipage}}
    \hspace{2mm}
    \subfloat[\textbf{Loss.} obj loss is set as default as it is essential. 
    \label{tab:ablation_loss}]{
        \begin{minipage}{0.35\linewidth}{\begin{center}
        \hspace{-8pt}
        \tablestyle{1pt}{1.2} 
        \begin{tabular}{x{15}x{15}x{15}|x{21}x{21}x{21}x{21}x{21}}
            obj&\(\ell_1\)&\multicolumn{1}{c|}{gIoU}&\(\rm {R}^{1}_{0.5}\)&\(\rm {R}^{1}_{0.7}\)&\(\rm {R}^{5}_{0.5}\)&\(\rm {R}^{5}_{0.7}\)&mIoU\\\shline
            \cmark&\cmark&      &{33.64}&{20.75}&{46.40}&{24.37}&{32.44}\\ 
            \cmark&      &\cmark&{32.13}&{18.16}&{42.32}&{23.21}&{31.72}\\ 
            \rowcolor{Gray}
            \cmark&\cmark&\cmark&{\bf 35.60}&{\bf 23.19}&{\bf 54.06}&{\bf 32.95}&{\bf 33.76}\\
    \end{tabular}
    \end{center}}\end{minipage}}
    \hspace{2mm}
    \subfloat[\textbf{Input video density.}
    \label{tab:ablation_frames}]{
        \begin{minipage}{0.29\linewidth}{\begin{center}
        \hspace{-8pt}
        \tablestyle{1pt}{1.2} 
        \begin{tabular}{x{20}|x{21}x{21}x{21}x{21}x{21}}
            \multicolumn{1}{c|}{frames}&\(\rm {R}^{1}_{0.5}\)&\(\rm {R}^{1}_{0.7}\)&\(\rm {R}^{5}_{0.5}\)&\(\rm {R}^{5}_{0.7}\)&mIoU\\\shline
            16&{32.66}&{18.79}&{\bf 54.24}&{31.85}&{31.25}\\
            \rowcolor{Gray}
            32&{\bf 35.60}&{\bf 23.19}&{54.06}&{\bf 32.95}&{\bf 33.76}\\
            64&{34.91}&{22.83}&{53.72}&{31.23}&{32.48}\\ 
    \end{tabular}
    \end{center}}\end{minipage}}
    }
    \vspace{\abovetabcapmargin}
    \caption{
        \textbf{Ablative experiments.}
        Our settings are marked in \colorbox{Gray}{gray}.
        All experiments are conducted on the {\sketchy} dataset.
    }
    \label{tab:ablations}
    \vspace{\belowtabcapmargin}
\end{table*}

\vspace{\abovesubsecmargin}\subsection{Ablative Study}

    \vspace{\paramargin}\paragraph{CMT attention operations.}
    We study the effect of four attention operations of CMT in~\Cref{tab:ablation_cmt}.
    Here, SVCA and CTCA are set as default since they are indispensable for making predictions in our design.
    Each is responsible for modeling interaction between sketch and video, and transforming object tokens into predictions conditioned on the sketch-video joint representations.
    CSA models the global context of the input sequence and TSA models relationships between object tokens.
    The default setting work fairly well (mIoU=30.72\%), yet \ours shows better performance with the addition of CSA (+2.33\%p) or TSA (+0.97\%p).
    In particular, CSA plays a crucial role in object localization in video since it is in charge of temporal modeling, thus leading to a substantial performance increase.
    We confirm that all CMT components operate collaboratively on the SVOL task, as they achieve the best performance when used together.
    
    \vspace{\paramargin}\paragraph{CMT depth.}
    We examine the effect of varying the CMT depth (\ie, number of layers) in~\Cref{tab:ablation_layers}.
    A single layer of CMT does not provide sufficient contextualization, resulting in poor ${\rm R}^5$ performance.
    The overall performance seems balanced between two to four layers.
    For ${\rm R}^5$ metric, the deeper the layer, the better the performance, and the best performance is achieved with four CMT layers.
    However, for more strict ${\rm R}^1$ and mIoU metrics, two layers perform the best.
    Therefore, we made two layers as our default setting.

\begin{figure}[t!]
    \centering
    \includegraphics[width=\linewidth]{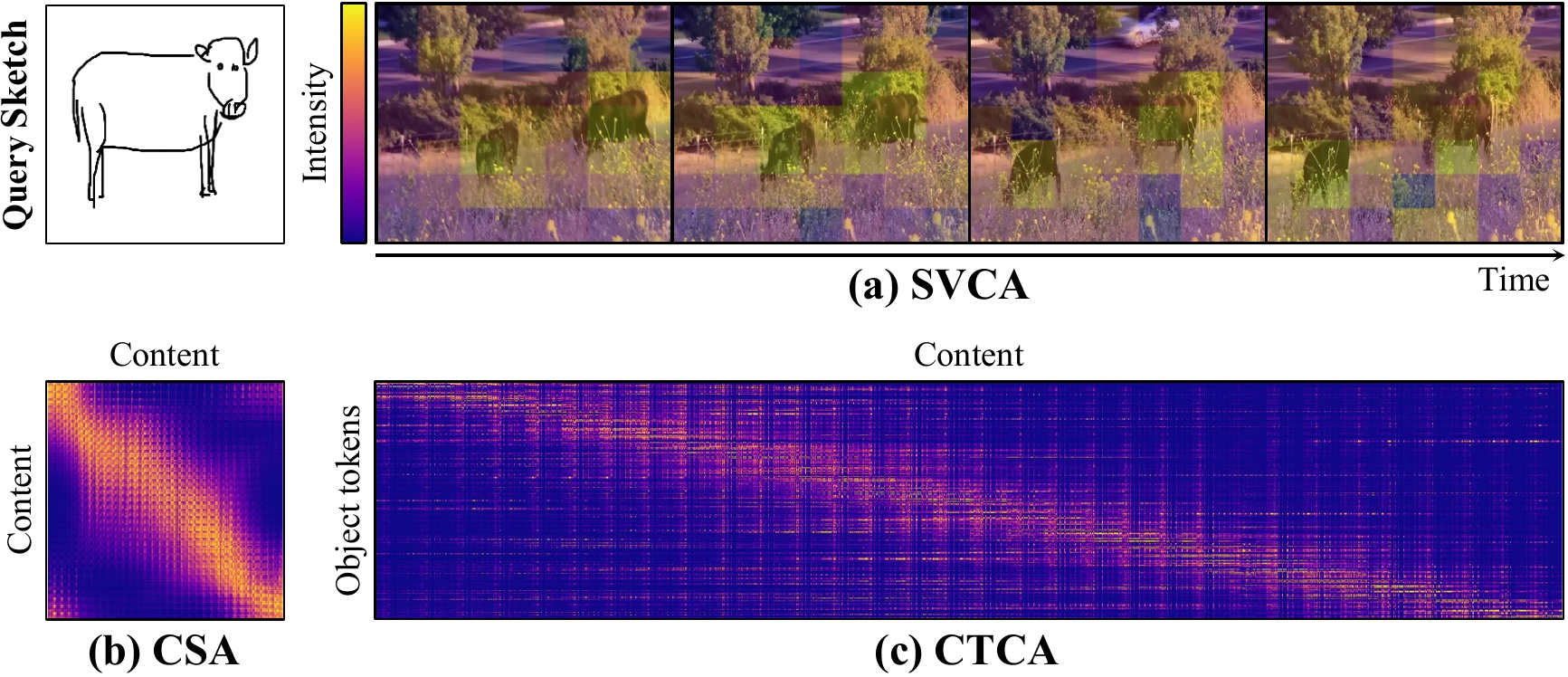}
    \vskip \abovefigcapmargin
    \caption{
        \textbf{Attention visualization.}
        The brighter (yellowish) the color, the higher the attention intensity.
        Given a query sketch,
        (a) SVCA mainly attends to regions where the target objects are located;
        (b) CSA gives a higher attention weight to temporally neighboring contents;
        (b) CTCA learns which content object tokens should mainly focus on temporally.
    }
    \label{fig:attention}
    \vspace{\belowfigcapmargin}
\end{figure}

    \vspace{\paramargin}\paragraph{Per-frame set matching.}
    A straightforward way for training \ours is to match all predictions with all ground truths as a whole.
    Although simple, it requires learning all $N$ object tokens simultaneously, regardless of frame order.
    On the contrary, our per-frame set matching strategy divides $N$ object tokens into $L$ subsets of $M$ object tokens, then matches only a subset to ground truths of its corresponding frame.
    Although set matching is performed frame-by-frame, \ours can still make predictions in parallel.
    We compare our strategy to the straightforward approach in~\Cref{tab:ablation_matching}. 
    Overall, using per-frame set matching resulted in a significant performance improvement.
    We see this is because our strategy not only eases optimization by reducing the set matching complexity, but also brings a strong positional inductive bias for object tokens (see empirical evidence in~\cref{fig:attention}{\color{linkcolor}c}).

    \vspace{\paramargin}\paragraph{Number of object tokens.}
    In order to see the effect of the number of object tokens used in the CMT layers, we varied their number in~\Cref{tab:ablation_tokens}.
    Too few tokens (=5) limit sufficient interactions between foregrounds and backgrounds (\ie, \texttt{No Object}), resulting in poor performance, especially for ${\rm R}^{5}$ metric.
    On the other hand, too many tokens (=15) diminish performance by producing unnecessary backgrounds.
    Having 10 object tokens per frame provides a good balance between foreground and background, resulting in a good performance. 
    As we utilize a per-frame set matching strategy, we set the number of object tokens per frame to 10 for the entire 32 frames, thus using a total of 320 object tokens.
    
    \vspace{\paramargin}\paragraph{Loss components.}
    In~\Cref{tab:ablation_loss}, we toggle the loss components on and off to understand their impact on training.
    The objectness loss is used in all cases since it is essential to determine whether a prediction contains the target object.
    When either $\ell_1$ or gIoU~\cite{rezatofighi2019generalized} loss is disabled, performances drop drastically, especially in ${\rm R}^{5}$ metric.
    This implies that both losses are not only important for accurate box localization, but also for performing overall predictions well.
    As we obtain the best results when using all three losses, we confirm that scale-sensitive $\ell_1$ and scale-invariant gIoU losses operate complementarily with each other.
    
    \vspace{\paramargin}\paragraph{Sampling density of video frames.}
    In~\Cref{tab:ablation_frames}, we study the effect of frame sampling density on input video.
    We uniformly sample a fixed number of frames across the video and use them as an input to \ours.
    By default, we use 32 frames. 
    Compared to the baseline, 16 frames show a particularly sharp performance drop on the strictest metric ${\rm R}^{1}_{0.7}$, and 64 frames show overall sub-optimal performance.
    This is because sparse sampling enables faster processing with less memory, but can easily miss important details since it provides less information.
    In contrast, dense sampling provides more information, but if the motion of objects is not large, it can be redundant and rather hinder optimization.
    Here, we study only simple uniform sampling, but different means of sampling may achieve different results.
    

\begin{figure}[t!]
    \centering
    \includegraphics[width=\linewidth]{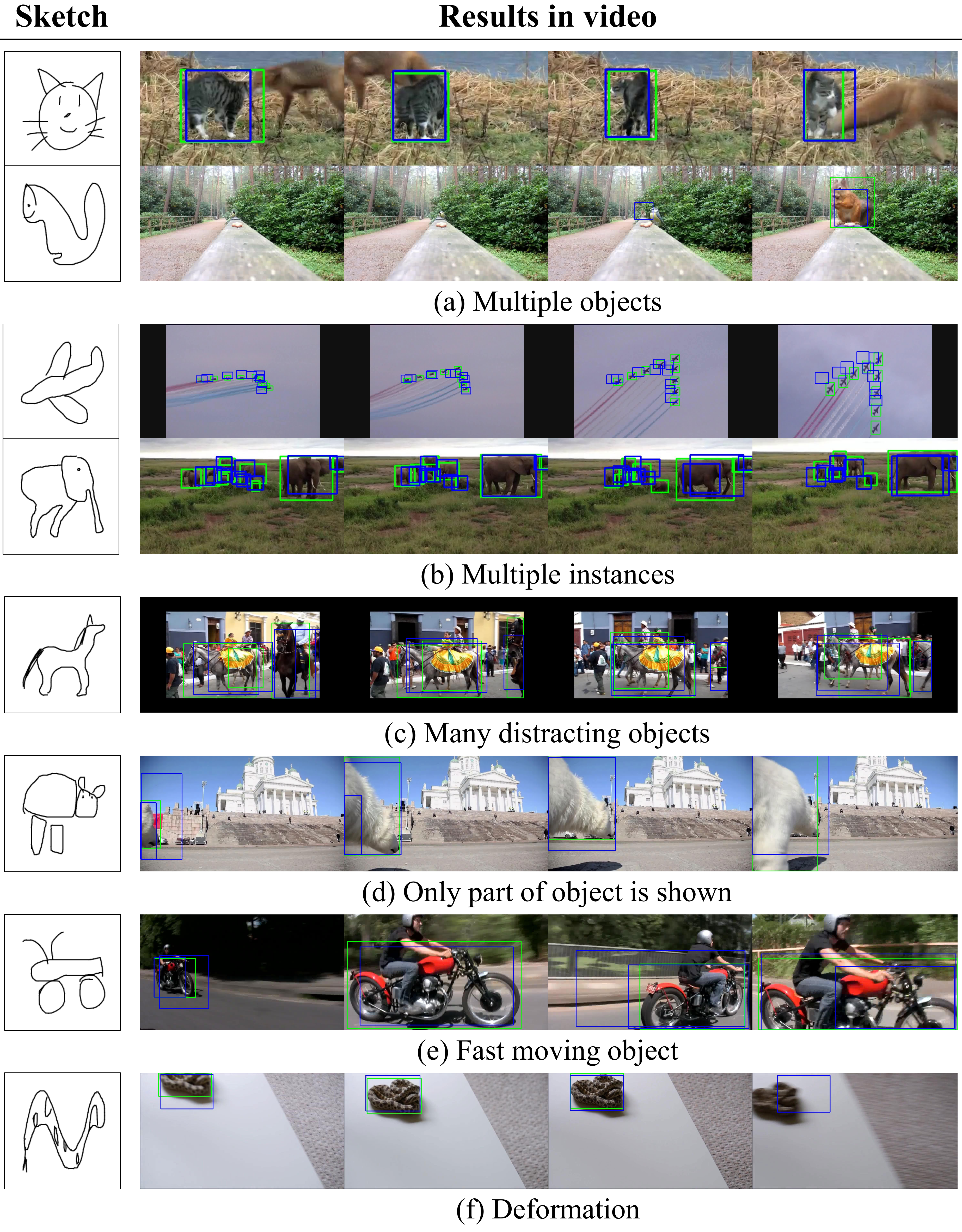}
    \vskip \abovefigcapmargin
    \caption{
        \textbf{Qualitative results} of \ours on {\quickdraw} dataset.
        {\color{green} Green} and {\color{blue} blue} boxes represent {\color{green} ground truths} and {\color{blue} predictions}, respectively.
        \ours performs well in various challenging scenarios, including:
        \textbf{(a)} when there are confusable objects;
        \textbf{(b)} multiple object instances appear in a video;
        \textbf{(c)} there are many distracting objects;
        \textbf{(d)} only part of the object is visible;
        \textbf{(e)} the target object moves quickly;
        \textbf{(f)} the appearance of the target object is not similar to query sketch.
    }
    \label{fig:qualitative}
    \vspace{\belowfigcapmargin}
    \vspace{2mm}
\end{figure}

\vspace{\abovesubsecmargin}\subsection{Qualitative Analysis}
    \vspace{\paramargin}\paragraph{SVOL results.}
    We present qualitative results of \ours in~\cref{fig:qualitative} to illustrate how it works in practice.\footnote{We present additional qualitative results in~\cref{sec:additional_qualitative}.}
    Our system successfully recognizes the objects that correspond to the query sketch and accurately localizes their bounding boxes in a variety of challenging conditions.
    \ours works well even when:
    (a) there are two confusable objects;
    (b) multiple object instances appear in a video;
    (c) there are lots of distracting objects;
    (d) only part of the object is appearing;
    (e) the target object moves quickly;
    (f) the appearance of the query sketch is not similar to that of the target object.

    \vspace{\paramargin}\paragraph{CMT attention visualization.}  
    In order to understand the behavior of CMT, we visualize its attention maps in~\cref{fig:attention}.
    Our observations are as follows:
    (a) SVCA learns \textit{where to look}, as such, the highlighted area on the attention map aligns well with the actual locations of the sketch object.
    (b) CSA learns deeper correlation between temporally adjacent sequences when modeling temporal context.
    (c) CTCA learns \textit{when to look}, thereby giving temporal inductive bias for object tokens in conjunction with per-frame set matching.
\vspace{\abovesecmargin}
\section{Conclusion}
\label{sec:conclusion}
\vspace{\belowsecmargin}
We have introduced a new challenging task termed Sketch-b Video Object Localization (SVOL), which aims to localize objects in video that match a given query sketch.
To tackle this task, we propose a strong baseline model called \ours, which addresses the temporal context of video and the domain gap between sketches and videos.
Our model, \ours, utilizes two key designs to solve the SVOL task as a set prediction problem: a Cross-modal Transformer and per-frame set matching.
To verify our approach, we conduct extensive experiments on a newly curated SVOL dataset and have found that \ours significantly outperforms image-level approaches by 29.4\%, 17.7\%, and 16.8\%.
We further analyze the behavior of \ours through comprehensive ablations and visualizations.
Last but not least, we found that \ours generalizes well to unseen datasets and novel categories, implying its scalability in real-world scenarios.
We hope our work will inspire further research in the field of sketch-based object localization.\footnote{Limitations, future work, and broader impacts are in~\cref{sec:discussion}.}

{\small
\bibliographystyle{ieee_fullname}
\bibliography{_main}

\begin{thebibliography}{10}\itemsep=-1pt

\bibitem{anne2017localizing}
Lisa Anne~Hendricks, Oliver Wang, Eli Shechtman, Josef Sivic, Trevor Darrell, and Bryan Russell.
\newblock {Localizing Moments in Video With Natural Language}.
\newblock In {\em ICCV}, pages 5803--5812, 2017.

\bibitem{bay2006surf}
Herbert Bay, Tinne Tuytelaars, and Luc~Van Gool.
\newblock {Surf: Speeded Up Robust Features}.
\newblock In {\em ECCV}, pages 404--417, 2006.

\bibitem{bhunia2022doodle}
Ayan~Kumar Bhunia, Viswanatha~Reddy Gajjala, Subhadeep Koley, Rohit Kundu, Aneeshan Sain, Tao Xiang, and Yi-Zhe Song.
\newblock {Doodle It Yourself: Class Incremental Learning by Drawing a Few Sketches}.
\newblock In {\em CVPR}, pages 2293--2302, 2022.

\bibitem{bhunia2020sketch}
Ayan~Kumar Bhunia, Yongxin Yang, Timothy~M Hospedales, Tao Xiang, and Yi-Zhe Song.
\newblock {Sketch Less for More: On-The-Fly Fine-Grained Sketch-Based Image Retrieval}.
\newblock In {\em CVPR}, pages 9779--9788, 2020.

\bibitem{boniardi2016autonomous}
Federico Boniardi, Abhinav Valada, Wolfram Burgard, and Gian~Diego Tipaldi.
\newblock {Autonomous Indoor Robot Navigation Using a Sketch Interface for Drawing Maps and Routes}.
\newblock In {\em ICRA}, pages 2896--2901. IEEE, 2016.

\bibitem{botach2022end}
Adam Botach, Evgenii Zheltonozhskii, and Chaim Baskin.
\newblock End-to-end referring video object segmentation with multimodal transformers.
\newblock In {\em CVPR}, pages 4985--4995, 2022.

\bibitem{brown2020language}
Tom Brown, Benjamin Mann, Nick Ryder, Melanie Subbiah, Jared~D Kaplan, Prafulla Dhariwal, Arvind Neelakantan, Pranav Shyam, Girish Sastry, Amanda Askell, et~al.
\newblock {Language Models Are Few-Shot Learners}.
\newblock In {\em NeurIPS}, pages 1877--1901, 2020.

\bibitem{carion2020end}
Nicolas Carion, Francisco Massa, Gabriel Synnaeve, Nicolas Usunier, Alexander Kirillov, and Sergey Zagoruyko.
\newblock {End-to-End Object Detection With Transformers}.
\newblock In {\em ECCV}, pages 213--229, 2020.

\bibitem{chang2022webqa}
Yingshan Chang, Mridu Narang, Hisami Suzuki, Guihong Cao, Jianfeng Gao, and Yonatan Bisk.
\newblock {WebQA: Multihop and Multimodal QA}.
\newblock In {\em CVPR}, pages 16495--16504, 2022.

\bibitem{chen2018sketchygan}
Wengling Chen and James Hays.
\newblock {Sketchygan: Towards Diverse and Realistic Sketch to Image Synthesis}.
\newblock In {\em CVPR}, pages 9416--9425, 2018.

\bibitem{chen2021transformer}
Xin Chen, Bin Yan, Jiawen Zhu, Dong Wang, Xiaoyun Yang, and Huchuan Lu.
\newblock {Transformer Tracking}.
\newblock In {\em CVPR}, pages 8126--8135, 2021.

\bibitem{chen2020memory}
Yihong Chen, Yue Cao, Han Hu, and Liwei Wang.
\newblock {End-to-End Video Object Detection With Spatial-Temporal Transformers}.
\newblock In {\em CVPR}, pages 10337--10346, 2020.

\bibitem{chen2020siamese}
Zedu Chen, Bineng Zhong, Guorong Li, Shengping Zhang, and Rongrong Ji.
\newblock {Siamese Box Adaptive Network for Visual Tracking}.
\newblock In {\em CVPR}, pages 6668--6677, 2020.

\bibitem{dalal2005histograms}
Navneet Dalal and Bill Triggs.
\newblock {Histograms of Oriented Gradients for Human Detection}.
\newblock In {\em CVPR}, pages 886--893, 2005.

\bibitem{deng2018visual}
Chaorui Deng, Qi Wu, Qingyao Wu, Fuyuan Hu, Fan Lyu, and Mingkui Tan.
\newblock {Visual Grounding via Accumulated Attention}.
\newblock In {\em CVPR}, pages 7746--7755, 2018.

\bibitem{deng2021transvg}
Jiajun Deng, Zhengyuan Yang, Tianlang Chen, Wengang Zhou, and Houqiang Li.
\newblock {TransVG: End-to-End Visual Grounding With Transformers}.
\newblock In {\em ICCV}, pages 1769--1779, 2021.

\bibitem{devlin2018bert}
Jacob Devlin, Ming-Wei Chang, Kenton Lee, and Kristina Toutanova.
\newblock {BERT: Pre-Training of Deep Bidirectional Transformers for Language Understanding}.
\newblock {\em arXiv preprint arXiv:1810.04805}, 2018.

\bibitem{dey2019doodle}
Sounak Dey, Pau Riba, Anjan Dutta, Josep Llados, and Yi-Zhe Song.
\newblock {Doodle To Search: Practical Zero-Shot Sketch-Based Image Retrieval}.
\newblock In {\em CVPR}, pages 2179--2188, 2019.

\bibitem{dosovitskiy2020image}
Alexey Dosovitskiy, Lucas Beyer, Alexander Kolesnikov, Dirk Weissenborn, Xiaohua Zhai, Thomas Unterthiner, Mostafa Dehghani, Matthias Minderer, Georg Heigold, Sylvain Gelly, et~al.
\newblock {An Image Is Worth 16x16 Words: Transformers for Image Recognition at Scale}.
\newblock {\em arXiv preprint arXiv:2010.11929}, 2020.

\bibitem{eitz2012humans}
Mathias Eitz, James Hays, and Marc Alexa.
\newblock {How Do Humans Sketch Objects?}
\newblock {\em ACM TOG}, pages 1--10, 2012.

\bibitem{fan2020person}
Hehe Fan and Yi Yang.
\newblock {Person Tube Retrieval via Language Description}.
\newblock In {\em AAAI}, pages 10754--10761, 2020.

\bibitem{fan2020few}
Qi Fan, Wei Zhuo, Chi-Keung Tang, and Yu-Wing Tai.
\newblock {Few-Shot Object Detection With Attention-RPN and Multi-Relation Detector}.
\newblock In {\em CVPR}, pages 4013--4022, 2020.

\bibitem{feichtenhofer2017detect}
Christoph Feichtenhofer, Axel Pinz, and Andrew Zisserman.
\newblock {Detect To Track and Track To Detect}.
\newblock In {\em ICCV}, pages 3038--3046, 2017.

\bibitem{fukui2016multimodal}
Akira Fukui, Dong~Huk Park, Daylen Yang, Anna Rohrbach, Trevor Darrell, and Marcus Rohrbach.
\newblock {Multimodal Compact Bilinear Pooling for Visual Question Answering and Visual Grounding}.
\newblock {\em arXiv preprint arXiv:1606.01847}, 2016.

\bibitem{ganin2015unsupervised}
Yaroslav Ganin and Victor Lempitsky.
\newblock {Unsupervised Domain Adaptation by Backpropagation}.
\newblock In {\em ICML}, pages 1180--1189, 2015.

\bibitem{ganin2016domain}
Yaroslav Ganin, Evgeniya Ustinova, Hana Ajakan, Pascal Germain, Hugo Larochelle, Fran{\c{c}}ois Laviolette, Mario Marchand, and Victor Lempitsky.
\newblock {Domain-Adversarial Training of Neural Networks}.
\newblock {\em JMLR}, pages 2096--2030, 2016.

\bibitem{gao2017tall}
Jiyang Gao, Chen Sun, Zhenheng Yang, and Ram Nevatia.
\newblock {TALL: Temporal Activity Localization via Language Query}.
\newblock In {\em ICCV}, pages 5267--5275, 2017.

\bibitem{han2016seq}
Wei Han, Pooya Khorrami, Tom~Le Paine, Prajit Ramachandran, Mohammad Babaeizadeh, Honghui Shi, Jianan Li, Shuicheng Yan, and Thomas~S Huang.
\newblock {Seq-NMS for Video Object Detection}.
\newblock {\em arXiv preprint arXiv:1602.08465}, 2016.

\bibitem{he2016deep}
Kaiming He, Xiangyu Zhang, Shaoqing Ren, and Jian Sun.
\newblock {Deep Residual Learning for Image Recognition}.
\newblock In {\em CVPR}, pages 770--778, 2016.

\bibitem{hsieh2019one}
Ting-I Hsieh, Yi-Chen Lo, Hwann-Tzong Chen, and Tyng-Luh Liu.
\newblock {One-Shot Object Detection With Co-Attention and Co-Excitation}.
\newblock In {\em NeurIPS}, 2019.

\bibitem{hu2016natural}
Ronghang Hu, Huazhe Xu, Marcus Rohrbach, Jiashi Feng, Kate Saenko, and Trevor Darrell.
\newblock {Natural Language Object Retrieval}.
\newblock In {\em CVPR}, pages 4555--4564, 2016.

\bibitem{isola2017image}
Phillip Isola, Jun-Yan Zhu, Tinghui Zhou, and Alexei~A Efros.
\newblock {Image-to-Image Translation With Conditional Adversarial Networks}.
\newblock In {\em CVPR}, pages 1125--1134, 2017.

\bibitem{jiang2021handpainter}
Ying Jiang, Congyi Zhang, Hongbo Fu, Alberto Cannav{\`o}, Fabrizio Lamberti, Henry~YK Lau, and Wenping Wang.
\newblock {HandPainter-3D Sketching in VR With Hand-Based Physical Proxy}.
\newblock In {\em CHI}, pages 1--13, 2021.

\bibitem{jongejan2016quick}
Jonas Jongejan, Henry Rowley, Takashi Kawashima, Jongmin Kim, and Nick Fox-Gieg.
\newblock {The Quick, Draw!-AI Experiment}.
\newblock {\em Mount View, CA, accessed Feb}, page~4, 2016.

\bibitem{kim2021hotr}
Bumsoo Kim, Junhyun Lee, Jaewoo Kang, Eun-Sol Kim, and Hyunwoo~J Kim.
\newblock {Hotr: End-to-End Human-Object Interaction Detection With Transformers}.
\newblock In {\em CVPR}, pages 74--83, 2021.

\bibitem{kong2014you}
Chen Kong, Dahua Lin, Mohit Bansal, Raquel Urtasun, and Sanja Fidler.
\newblock {What Are You Talking About? Text-to-Image Coreference}.
\newblock In {\em CVPR}, pages 3558--3565, 2014.

\bibitem{kuhn1955hungarian}
Harold~W Kuhn.
\newblock {The Hungarian Method for the Assignment Problem}.
\newblock {\em Naval research logistics quarterly}, pages 83--97, 1955.

\bibitem{kwan2019mobi3dsketch}
Kin~Chung Kwan and Hongbo Fu.
\newblock {Mobi3dsketch: 3D Sketching in Mobile AR}.
\newblock In {\em CHI}, pages 1--11, 2019.

\bibitem{lee2023modality}
Sumin Lee, Sangmin Woo, Yeonju Park, Muhammad~Adi Nugroho, and Changick Kim.
\newblock {Modality Mixer for Multi-modal Action Recognition}.
\newblock In {\em WACV}, pages 3298--3307, 2023.

\bibitem{lei2019tvqa+}
Jie Lei, Licheng Yu, Tamara~L Berg, and Mohit Bansal.
\newblock {TVQA+: Spatio-Temporal Grounding for Video Question Answering}.
\newblock {\em arXiv preprint arXiv:1904.11574}, 2019.

\bibitem{li2019siamrpn++}
Bo Li, Wei Wu, Qiang Wang, Fangyi Zhang, Junliang Xing, and Junjie Yan.
\newblock {Siamrpn++: Evolution of Siamese Visual Tracking With Very Deep Networks}.
\newblock In {\em CVPR}, pages 4282--4291, 2019.

\bibitem{li2018high}
Bo Li, Junjie Yan, Wei Wu, Zheng Zhu, and Xiaolin Hu.
\newblock {High Performance Visual Tracking With Siamese Region Proposal Network}.
\newblock In {\em CVPR}, pages 8971--8980, 2018.

\bibitem{li2019neural}
Naihan Li, Shujie Liu, Yanqing Liu, Sheng Zhao, and Ming Liu.
\newblock {Neural Speech Synthesis With Transformer Network}.
\newblock In {\em AAAI}, volume~33, pages 6706--6713, 2019.

\bibitem{li2020oscar}
Xiujun Li, Xi Yin, Chunyuan Li, Pengchuan Zhang, Xiaowei Hu, Lei Zhang, Lijuan Wang, Houdong Hu, Li Dong, Furu Wei, et~al.
\newblock {OSCAR: Object-Semantics Aligned Pre-training for Vision-Language Tasks}.
\newblock In {\em ECCV}, pages 121--137. Springer, 2020.

\bibitem{liang2018planar}
Pengpeng Liang, Yifan Wu, Hu Lu, Liming Wang, Chunyuan Liao, and Haibin Ling.
\newblock Planar object tracking in the wild: A benchmark.
\newblock In {\em 2018 IEEE International Conference on Robotics and Automation (ICRA)}, pages 651--658. IEEE, 2018.

\bibitem{lin2017feature}
Tsung-Yi Lin, Piotr Doll{\'a}r, Ross Girshick, Kaiming He, Bharath Hariharan, and Serge Belongie.
\newblock {Feature pyramid networks for object detection}.
\newblock In {\em CVPR}, pages 2117--2125, 2017.

\bibitem{literat2013pencil}
Ioana Literat.
\newblock “a pencil for your thoughts”: Participatory drawing as a visual research method with children and youth.
\newblock {\em International Journal of Qualitative Methods}, 12(1):84--98, 2013.

\bibitem{liu2022scenesketcher}
Fang Liu, Xiaoming Deng, Changqing Zou, Yu-Kun Lai, Keqi Chen, Ran Zuo, Cuixia Ma, Yong-Jin Liu, and Hongan Wang.
\newblock {Scenesketcher-V2: Fine-Grained Scene-Level Sketch-Based Image Retrieval Using Adaptive GCNs}.
\newblock {\em TIP}, 31:3737--3751, 2022.

\bibitem{liu2013image}
Jialu Liu.
\newblock {Image Retrieval Based on Bag-of-Words Model}.
\newblock {\em arXiv preprint arXiv:1304.5168}, 2013.

\bibitem{liu2016ssd}
Wei Liu, Dragomir Anguelov, Dumitru Erhan, Christian Szegedy, Scott Reed, Cheng-Yang Fu, and Alexander~C Berg.
\newblock {SSD: Single Shot Multibox Detector}.
\newblock In {\em ECCV}, pages 21--37, 2016.

\bibitem{loshchilov2017decoupled}
Ilya Loshchilov and Frank Hutter.
\newblock {Decoupled Weight Decay Regularization}.
\newblock {\em arXiv preprint arXiv:1711.05101}, 2017.

\bibitem{lowe2004distinctive}
David~G Lowe.
\newblock {Distinctive Image Features From Scale-Invariant Keypoints}.
\newblock {\em IJCV}, pages 91--110, 2004.

\bibitem{lu2019vilbert}
Jiasen Lu, Dhruv Batra, Devi Parikh, and Stefan Lee.
\newblock {ViLBERT: Pretraining Task-Agnostic Visiolinguistic Representations for Vision-and-Language Tasks}.
\newblock {\em NeurIPS}, 32, 2019.

\bibitem{lun20173d}
Zhaoliang Lun, Matheus Gadelha, Evangelos Kalogerakis, Subhransu Maji, and Rui Wang.
\newblock {3D Shape Reconstruction From Sketches via Multi-View Convolutional Networks}.
\newblock In {\em 3DV}, pages 67--77. IEEE, 2017.

\bibitem{osokin2020os2d}
Anton Osokin, Denis Sumin, and Vasily Lomakin.
\newblock {OS2D: One-Stage One-Shot Object Detection by Matching Anchor Features}.
\newblock In {\em ECCV}, pages 635--652, 2020.

\bibitem{peng2019moment}
Xingchao Peng, Qinxun Bai, Xide Xia, Zijun Huang, Kate Saenko, and Bo Wang.
\newblock {Moment Matching for Multi-Source Domain Adaptation}.
\newblock In {\em ICCV}, pages 1406--1415, 2019.

\bibitem{portenier2018faceshop}
Tiziano Portenier, Qiyang Hu, Attila Szabo, Siavash~Arjomand Bigdeli, Paolo Favaro, and Matthias Zwicker.
\newblock {Faceshop: Deep Sketch-Based Face Image Editing}.
\newblock {\em arXiv preprint arXiv:1804.08972}, 2018.

\bibitem{radford2021learning}
Alec Radford, Jong~Wook Kim, Chris Hallacy, Aditya Ramesh, Gabriel Goh, Sandhini Agarwal, Girish Sastry, Amanda Askell, Pamela Mishkin, Jack Clark, et~al.
\newblock {Learning Transferable Visual Models From Natural Language Supervision}.
\newblock In {\em ICML}, pages 8748--8763, 2021.

\bibitem{radford2018improving}
Alec Radford, Karthik Narasimhan, Tim Salimans, and Ilya Sutskever.
\newblock {Improving Language Understanding by Generative Pre-Training}.
\newblock 2018.

\bibitem{radford2019language}
Alec Radford, Jeffrey Wu, Rewon Child, David Luan, Dario Amodei, Ilya Sutskever, et~al.
\newblock {Language Models Are Unsupervised Multitask Learners}.
\newblock {\em OpenAI blog}, page~9, 2019.

\bibitem{ramesh2021zero}
Aditya Ramesh, Mikhail Pavlov, Gabriel Goh, Scott Gray, Chelsea Voss, Alec Radford, Mark Chen, and Ilya Sutskever.
\newblock {Zero-Shot Text-To-Image Generation}.
\newblock In {\em ICML}, pages 8821--8831. PMLR, 2021.

\bibitem{redmon2016you}
Joseph Redmon, Santosh Divvala, Ross Girshick, and Ali Farhadi.
\newblock {You Only Look Once: Unified, Real-Time Object Detection}.
\newblock In {\em CVPR}, pages 779--788, 2016.

\bibitem{ren2015faster}
Shaoqing Ren, Kaiming He, Ross Girshick, and Jian Sun.
\newblock {Faster R-CNN: Towards Real-Time Object Detection With Region Proposal Networks}.
\newblock In {\em NeurIPS}, 2015.

\bibitem{rezatofighi2019generalized}
Hamid Rezatofighi, Nathan Tsoi, JunYoung Gwak, Amir Sadeghian, Ian Reid, and Silvio Savarese.
\newblock {Generalized Intersection Over Union: A Metric and a Loss for Bounding Box Regression}.
\newblock In {\em CVPR}, pages 658--666, 2019.

\bibitem{riba2021localizing}
Pau Riba, Sounak Dey, Ali~Furkan Biten, and Josep Llados.
\newblock {Localizing Infinity-Shaped Fishes: Sketch-Guided Object Localization in the Wild}.
\newblock {\em arXiv preprint arXiv:2109.11874}, 2021.

\bibitem{russakovsky2015imagenet}
Olga Russakovsky, Jia Deng, Hao Su, Jonathan Krause, Sanjeev Satheesh, Sean Ma, Zhiheng Huang, Andrej Karpathy, Aditya Khosla, Michael Bernstein, et~al.
\newblock {Imagenet Large Scale Visual Recognition Challenge}.
\newblock {\em IJCV}, pages 211--252, 2015.

\bibitem{sain2021stylemeup}
Aneeshan Sain, Ayan~Kumar Bhunia, Yongxin Yang, Tao Xiang, and Yi-Zhe Song.
\newblock {Stylemeup: Towards Style-Agnostic Sketch-Based Image Retrieval}.
\newblock In {\em CVPR}, pages 8504--8513, 2021.

\bibitem{sakamoto2009sketch}
Daisuke Sakamoto, Koichiro Honda, Masahiko Inami, and Takeo Igarashi.
\newblock {Sketch and Run: A Stroke-Based Interface for Home Robots}.
\newblock In {\em CHI}, pages 197--200, 2009.

\bibitem{sangkloy2016sketchy}
Patsorn Sangkloy, Nathan Burnell, Cusuh Ham, and James Hays.
\newblock {The Sketchy Database: Learning To Retrieve Badly Drawn Bunnies}.
\newblock {\em ACM TOG}, pages 1--12, 2016.

\bibitem{song2017deep}
Jifei Song, Qian Yu, Yi-Zhe Song, Tao Xiang, and Timothy~M Hospedales.
\newblock {Deep Spatial-Semantic Attention for Fine-Grained Sketch-Based Image Retrieval}.
\newblock In {\em ICCV}, pages 5551--5560, 2017.

\bibitem{su2021stvgbert}
Rui Su, Qian Yu, and Dong Xu.
\newblock {Stvgbert: A Visual-Linguistic Transformer Based Framework for Spatio-Temporal Video Grounding}.
\newblock In {\em ICCV}, pages 1533--1542, 2021.

\bibitem{sun2022dli}
Haifeng Sun, Jiaqing Xu, Jingyu Wang, Qi Qi, Ce Ge, and Jianxin Liao.
\newblock {DLI-Net: Dual Local Interaction Network for Fine-Grained Sketch-Based Image Retrieval}.
\newblock {\em TCSVT}, 32(10):7177--7189, 2022.

\bibitem{taele2023sketchrec}
Paul Taele, Rachel Blagojevic, Tracy Hammond, Samantha Ray, Josh Cherian, and Jung~In Koh.
\newblock Sketchrec 2023: 3rd workshop on sketch recognition.
\newblock In {\em Companion Proceedings of the 28th International Conference on Intelligent User Interfaces}, pages 1--1, 2023.

\bibitem{tripathi2020sketch}
Aditay Tripathi, Rajath~R Dani, Anand Mishra, and Anirban Chakraborty.
\newblock {Sketch-Guided Object Localization in Natural Images}.
\newblock In {\em ECCV}, pages 532--547, 2020.

\bibitem{van2008visualizing}
Laurens Van~der Maaten and Geoffrey Hinton.
\newblock {Visualizing Data Using T-SNE}.
\newblock {\em JMLR}, 2008.

\bibitem{vaswani2017attention}
Ashish Vaswani, Noam Shazeer, Niki Parmar, Jakob Uszkoreit, Llion Jones, Aidan~N Gomez, {\L}ukasz Kaiser, and Illia Polosukhin.
\newblock {Attention is All You Need}.
\newblock In {\em NeurIPS}, pages 5998--6008, 2017.

\bibitem{vinyals2016matching}
Oriol Vinyals, Charles Blundell, Timothy Lillicrap, Daan Wierstra, et~al.
\newblock {Matching Networks for One Shot Learning}.
\newblock In {\em NeurIPS}, 2016.

\bibitem{wang2015sketch}
Fang Wang, Le Kang, and Yi Li.
\newblock {Sketch-Based 3D Shape Retrieval Using Convolutional Neural Networks}.
\newblock In {\em CVPR}, pages 1875--1883, 2015.

\bibitem{wang2022ptseformer}
Han Wang, Jun Tang, Xiaodong Liu, Shanyan Guan, Rong Xie, and Li Song.
\newblock Ptseformer: Progressive temporal-spatial enhanced transformer towards video object detection.
\newblock In {\em ECCV}, pages 732--747. Springer, 2022.

\bibitem{wang2021max}
Huiyu Wang, Yukun Zhu, Hartwig Adam, Alan Yuille, and Liang-Chieh Chen.
\newblock {Max-DeepLab: End-to-End Panoptic Segmentation With Mask Transformers}.
\newblock In {\em CVPR}, pages 5463--5474, 2021.

\bibitem{wang2021sketch}
Sheng-Yu Wang, David Bau, and Jun-Yan Zhu.
\newblock {Sketch Your Own Gan}.
\newblock In {\em ICCV}, pages 14050--14060, 2021.

\bibitem{woo2022towards}
Sangmin Woo, Sumin Lee, Yeonju Park, Muhammad~Adi Nugroho, and Changick Kim.
\newblock {Towards Good Practices for Missing Modality Robust Action Recognition}.
\newblock {\em arXiv preprint arXiv:2211.13916}, 2022.

\bibitem{woo2022explore}
Sangmin Woo, Jinyoung Park, Inyong Koo, Sumin Lee, Minki Jeong, and Changick Kim.
\newblock {Explore and Match: End-to-End Video Grounding With Transformer}.
\newblock {\em arXiv preprint arXiv:2201.10168}, 2022.

\bibitem{wu2019sequence}
Haiping Wu, Yuntao Chen, Naiyan Wang, and Zhaoxiang Zhang.
\newblock {Sequence Level Semantics Aggregation for Video Object Detection}.
\newblock In {\em ICCV}, pages 9217--9225, 2019.

\bibitem{xie2018rethinking}
Saining Xie, Chen Sun, Jonathan Huang, Zhuowen Tu, and Kevin Murphy.
\newblock {Rethinking Spatiotemporal Feature Learning: Speed-Accuracy Trade-Offs in Video Classification}.
\newblock In {\em ECCV}, pages 305--321, 2018.

\bibitem{2022_Sketch_survey}
Peng Xu, Timothy~M Hospedales, Qiyue Yin, Yi-Zhe Song, Tao Xiang, and Liang Wang.
\newblock {Deep Learning for Free-Hand Sketch: A Survey}.
\newblock {\em TPAMI}, 2022.

\bibitem{xu2020fine}
Peng Xu, Kun Liu, Tao Xiang, Timothy~M Hospedales, Zhanyu Ma, Jun Guo, and Yi-Zhe Song.
\newblock {Fine-Grained Instance-Level Sketch-Based Video Retrieval}.
\newblock {\em TCSVT}, pages 1995--2007, 2020.

\bibitem{yang2020deep}
Shuai Yang, Zhangyang Wang, Jiaying Liu, and Zongming Guo.
\newblock {Deep Plastic Surgery: Robust and Controllable Image Editing With Human-Drawn Sketches}.
\newblock In {\em ECCV}, pages 601--617. Springer, 2020.

\bibitem{yang2022lavt}
Zhao Yang, Jiaqi Wang, Yansong Tang, Kai Chen, Hengshuang Zhao, and Philip~HS Torr.
\newblock Lavt: Language-aware vision transformer for referring image segmentation.
\newblock In {\em CVPR}, pages 18155--18165, 2022.

\bibitem{yu2016sketch}
Qian Yu, Feng Liu, Yi-Zhe Song, Tao Xiang, Timothy~M Hospedales, and Chen-Change Loy.
\newblock {Sketch Me That Shoe}.
\newblock In {\em CVPR}, pages 799--807, 2016.

\bibitem{yu2021fine}
Qian Yu, Jifei Song, Yi-Zhe Song, Tao Xiang, and Timothy~M Hospedales.
\newblock {Fine-Grained Instance-Level Sketch-Based Image Retrieval}.
\newblock {\em IJCV}, 129(2):484--500, 2021.

\bibitem{zhang2021video}
Hao Zhang, Aixin Sun, Wei Jing, Guoshun Nan, Liangli Zhen, Joey~Tianyi Zhou, and Rick Siow~Mong Goh.
\newblock {Video Corpus Moment Retrieval With Contrastive Learning}.
\newblock In {\em SIGIR}, pages 685--695, 2021.

\bibitem{zhang2020does}
Zhu Zhang, Zhou Zhao, Yang Zhao, Qi Wang, Huasheng Liu, and Lianli Gao.
\newblock {Where Does It Exist: Spatio-Temporal Video Grounding for Multi-Form Sentences}.
\newblock In {\em CVPR}, pages 10668--10677, 2020.

\bibitem{zhou2022transvod}
Qianyu Zhou, Xiangtai Li, Lu He, Yibo Yang, Guangliang Cheng, Yunhai Tong, Lizhuang Ma, and Dacheng Tao.
\newblock Transvod: end-to-end video object detection with spatial-temporal transformers.
\newblock {\em TPAMI}, 2022.

\end{thebibliography}
}

\ifarxiv \clearpage \appendix
\label{sec:appendix}

\begin{figure*}[t!]
    \centering
    \setlength\tabcolsep{3mm}
    \resizebox*{\linewidth}{!}{%
    \subfloat[ID-agnostic
    \label{fig:id_agnostic}]{
        \begin{tabular}{c}
        \includegraphics[width=0.48\linewidth]{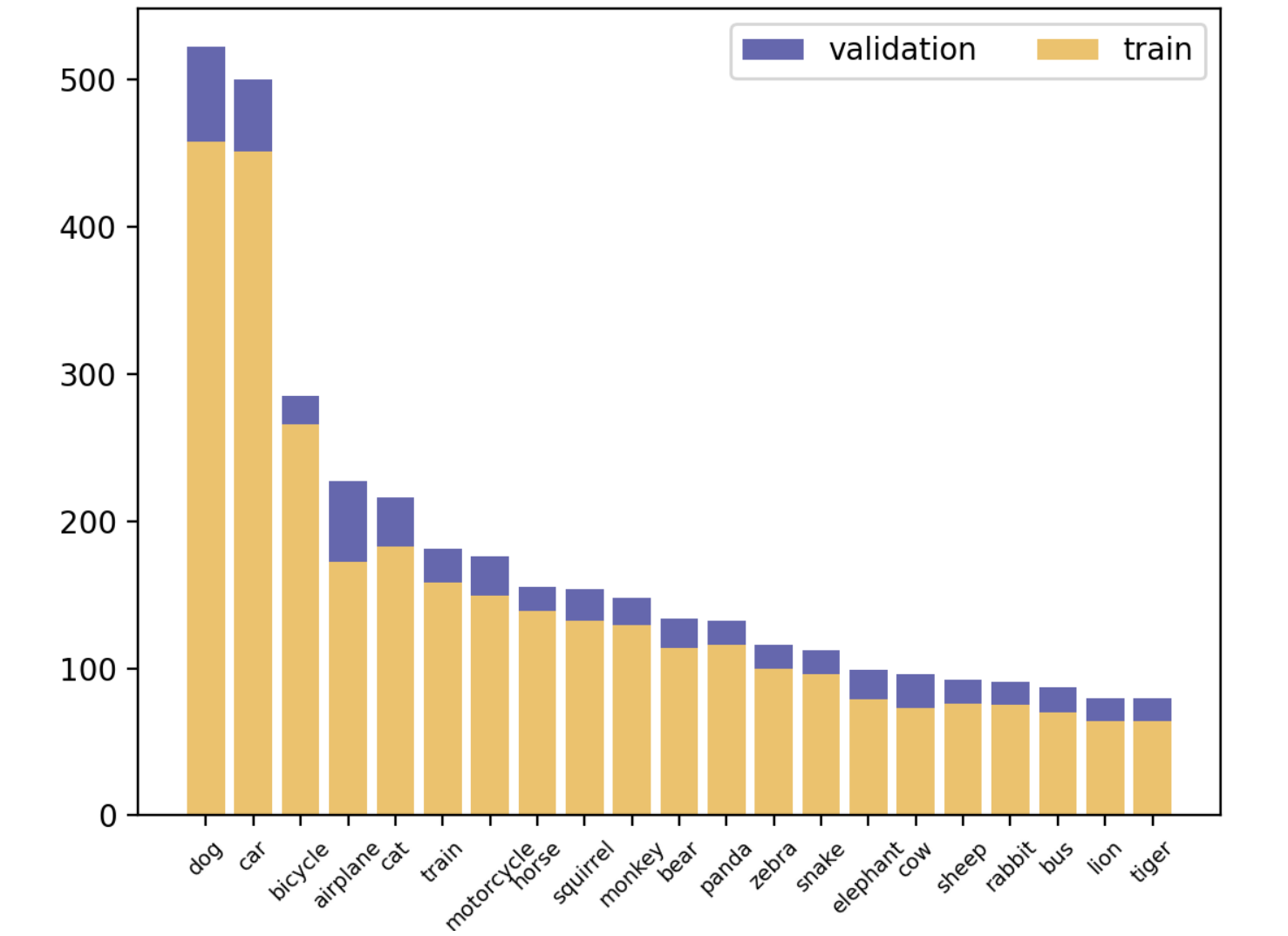}
        \end{tabular}
        }
        \hspace{-1.7ex}
    \subfloat[ID-specific
    \label{fig:id_specific}]{
        \begin{tabular}{c}
        \includegraphics[width=0.48\linewidth]{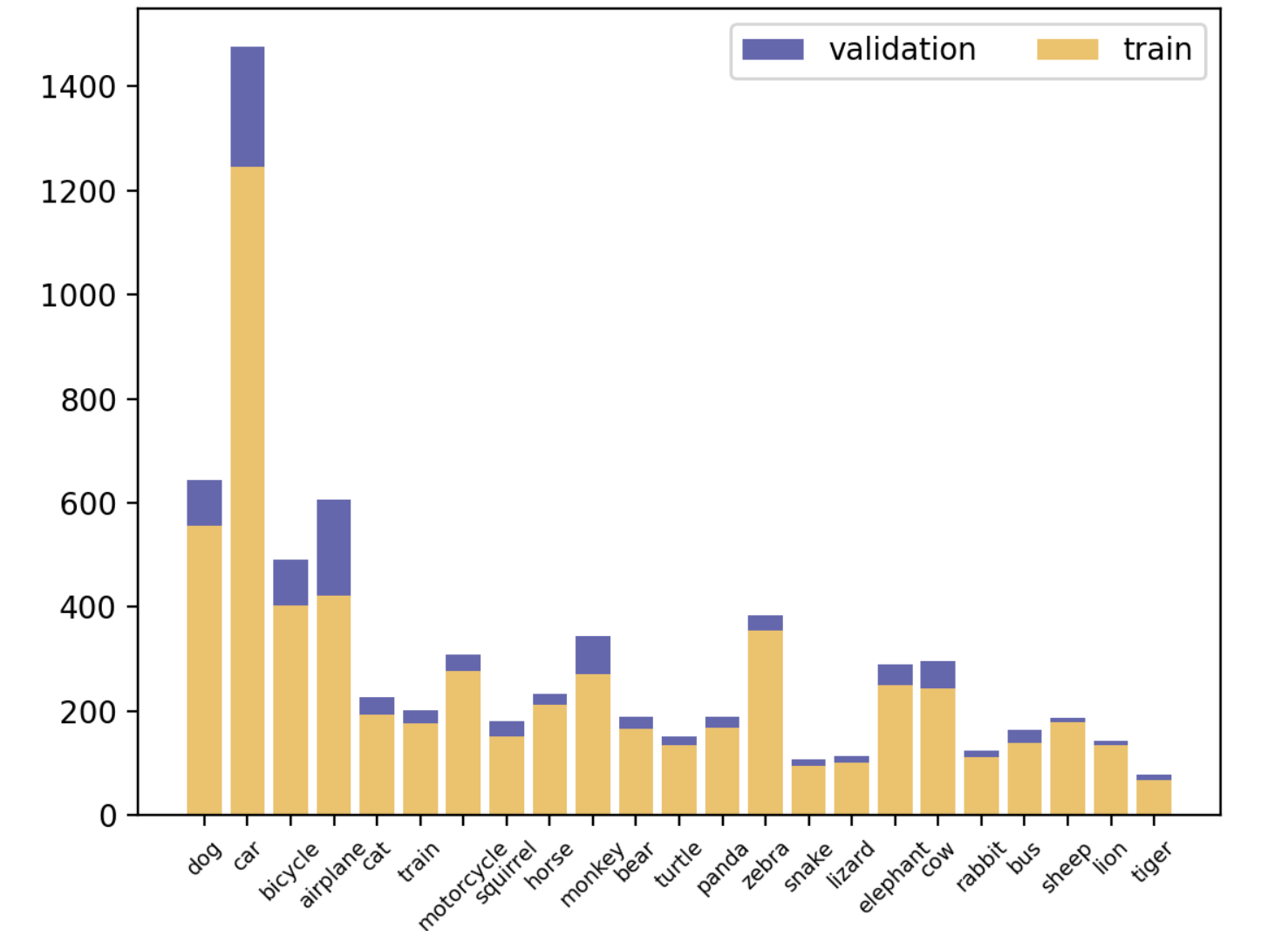}
        \end{tabular}
        }
    }
    \vskip \abovefigcapmargin
    \caption{
        \textbf{Video dataset~\cite{russakovsky2015imagenet} class distribution} when objects of the same category are \protect\subref{fig:id_agnostic}~counted as a whole (\ie, \textit{agnostic} to instance ID) or~\protect\subref{fig:id_specific}~counted individually (\ie, \textit{sensitive} to instance ID).
        The x-axis denotes the category and y-axis denotes the frequency.
    }
    \label{fig:video_dataset}
\end{figure*}

\begin{figure*}[t!]
    \centering
    \setlength\tabcolsep{3mm}
    \resizebox*{\linewidth}{!}{%
    \subfloat[{\sketchy}
    \label{fig:sketchy}]{
        \begin{tabular}{c}
        \includegraphics[width=0.32\linewidth]{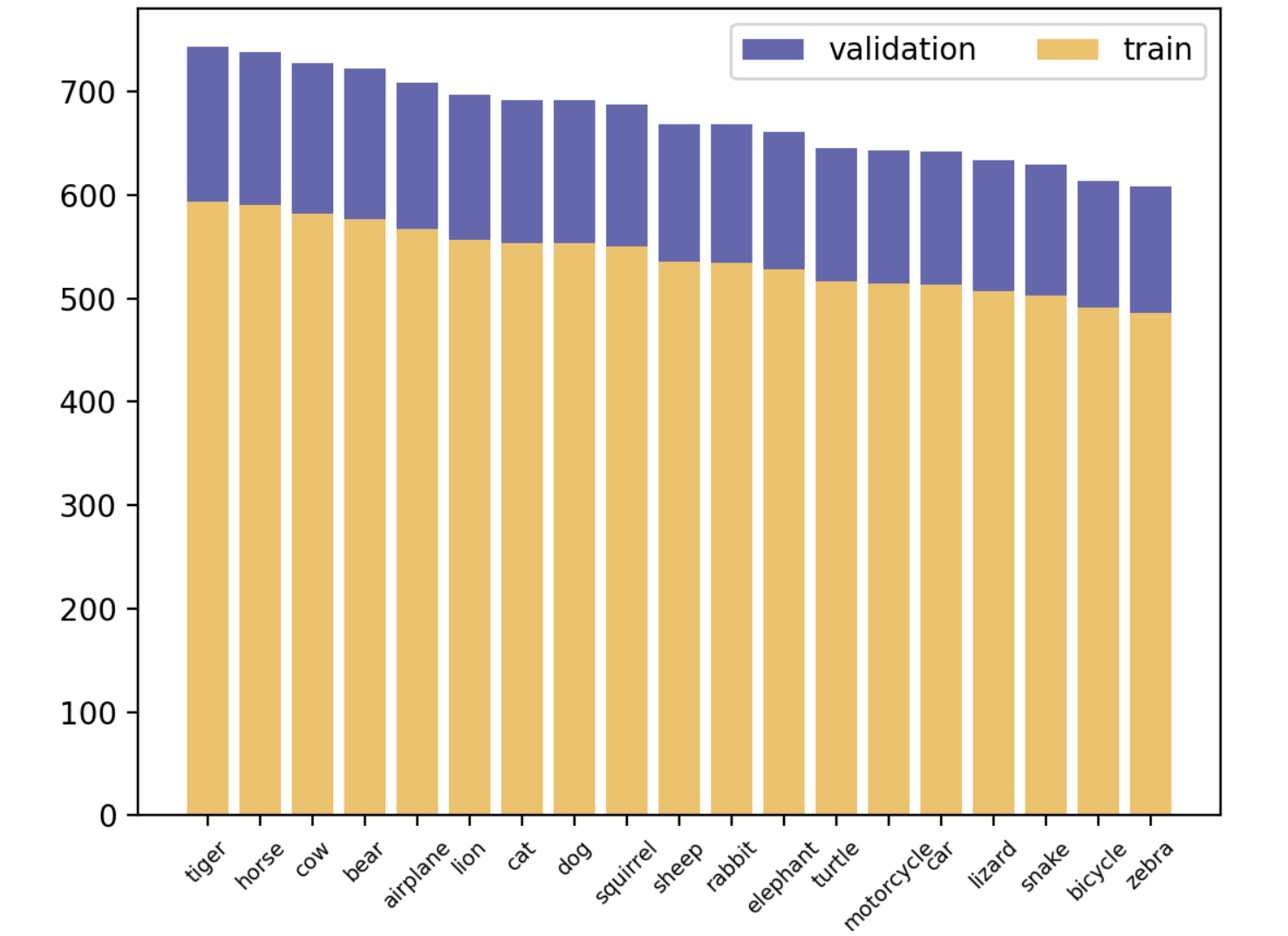}
        \end{tabular}
        }
    \hspace{-1.7ex}
    \subfloat[{\tuberlin}
    \label{fig:tuberlin}]{
        \begin{tabular}{c}
        \includegraphics[width=0.32\linewidth]{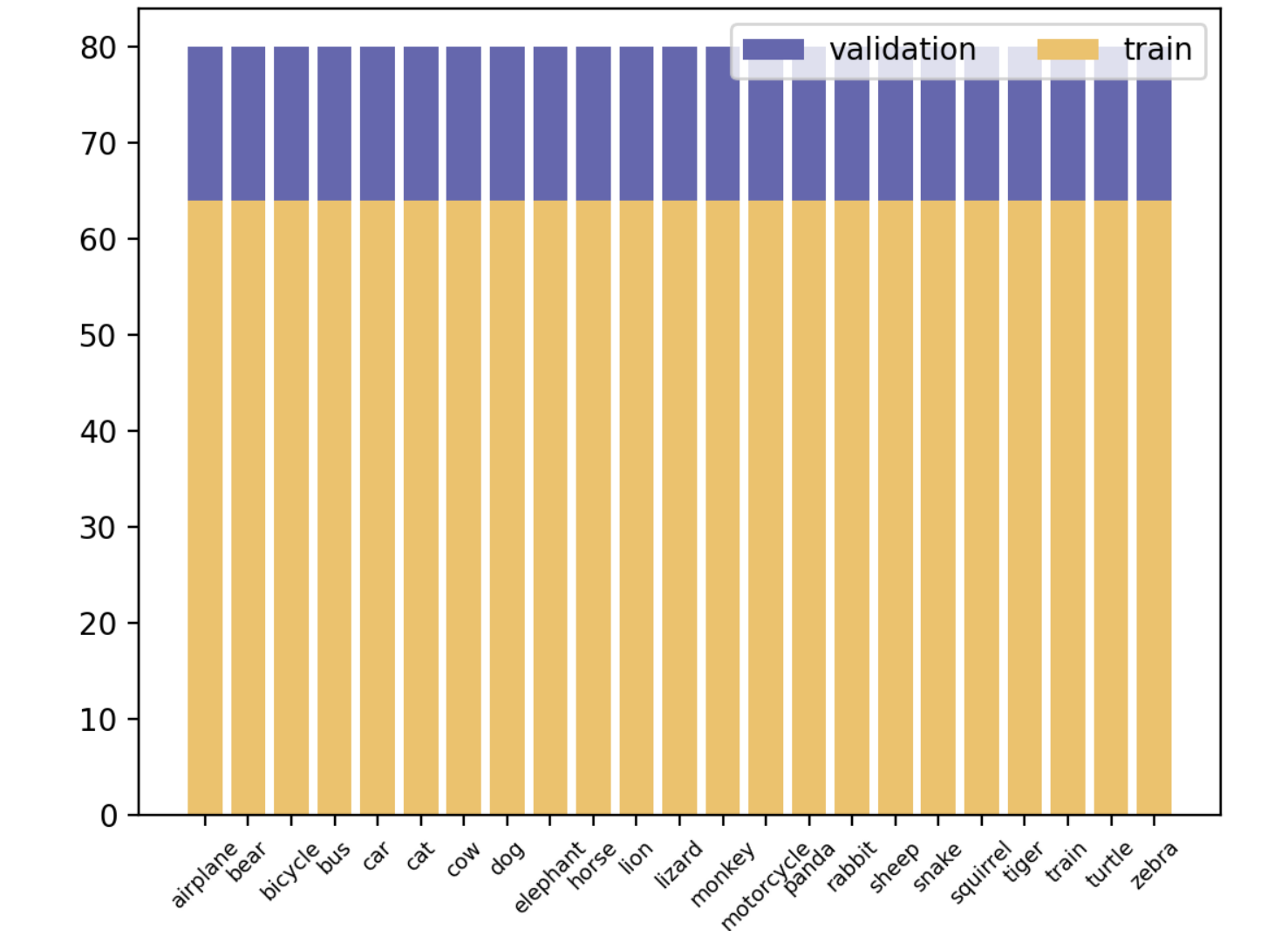}
        \end{tabular}
        }
    \hspace{-1.7ex}
    \subfloat[{\quickdraw}
    \label{fig:quickdraw}]{
        \begin{tabular}{c}
        \includegraphics[width=0.32\linewidth]{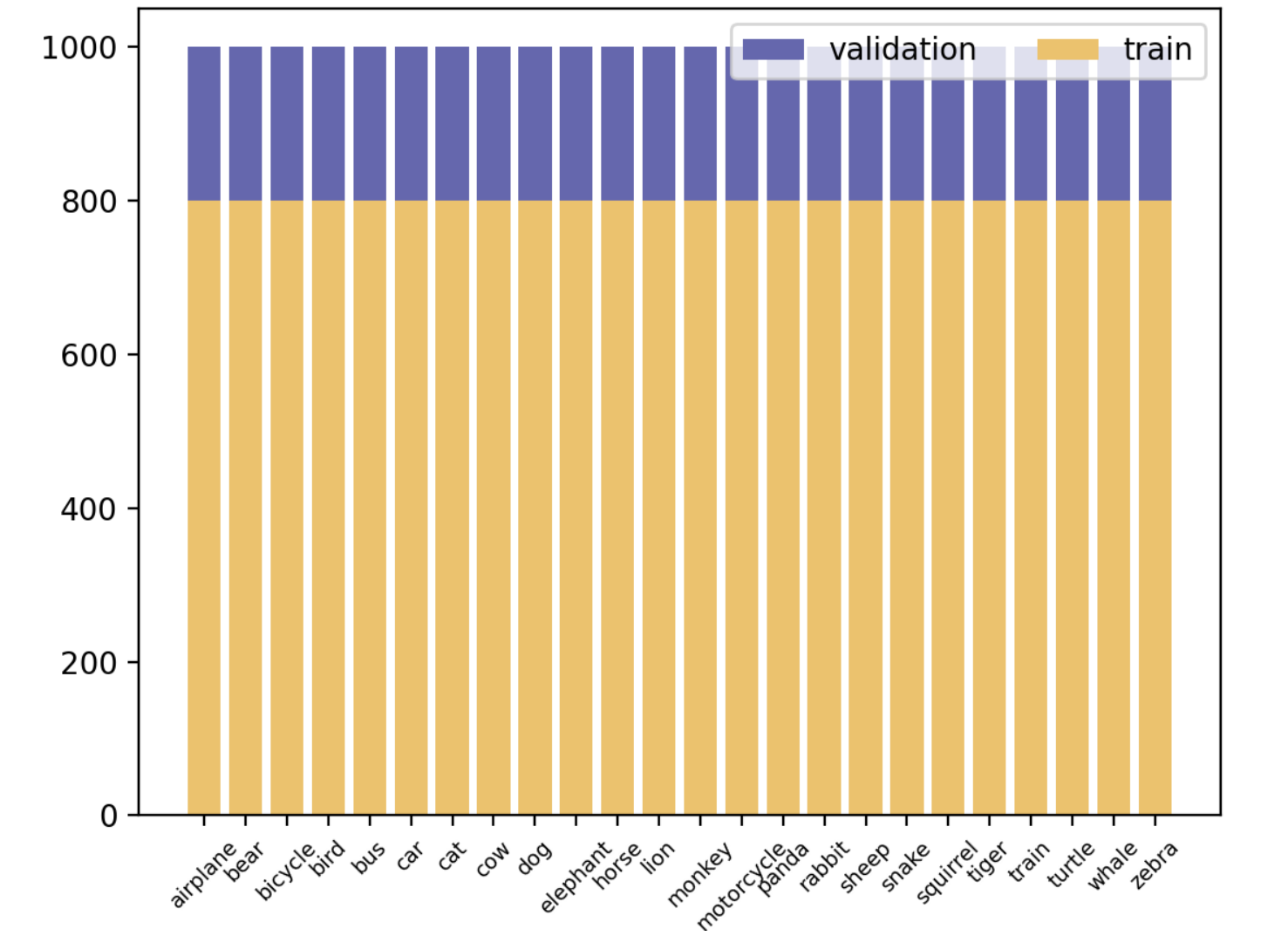}
        \end{tabular}
        }
    }
    \vskip \abovefigcapmargin
    \caption{
        \textbf{Sketch dataset class distribution.}~\protect\subref{fig:sketchy}~{\sketchy}~\cite{sangkloy2016sketchy},~\protect\subref{fig:tuberlin}~{\tuberlin}~\cite{eitz2012humans}, and~\protect\subref{fig:quickdraw}~{\quickdraw}~\cite{jongejan2016quick}.
        The x-axis denotes the class and y-axis denotes the frequency.
    }
    \label{fig:sketch_dataset}
\end{figure*}

\vspace{\abovesecmargin}
\section{SVOL Dataset \& Analysis}
\label{sec:svol_dataset}
\vspace{\belowsecmargin}
The SVOL dataset is built on multiple datasets~\cite{russakovsky2015imagenet,sangkloy2016sketchy,eitz2012humans,jongejan2016quick}.
We consider only the categories that intersects between the video dataset~\cite{russakovsky2015imagenet} and the sketch datasets~\cite{sangkloy2016sketchy,eitz2012humans,jongejan2016quick}.

\vspace{\abovesubsecmargin}
\subsection{SVOL Dataset}
\vspace{\belowsubsecmargin}
\noindent{\bf ImageNet-VID}~\cite{russakovsky2015imagenet} is built for video object detection task.
It contains 5,354 snippets (train/val/test split is 3,852/555/937) that are annotated with 30 object categories, including vehicles (\eg, airplane, bus, \etc.) and animals (\eg, bird, dog, \etc.).
Each object instance is annotated in the form of \{video name, frame number, class label, instance id, bounding box\}.
We use a validation set for evaluation since test annotations are not publicly available.

\vspace{2mm}\noindent{\bf \sketchy}~\cite{sangkloy2016sketchy} is a large-scale collection of sketch-photo pairs that has 75,471 sketches belonging to 125 categories.
Drawers are not allowed to directly trace objects; rather, sketches are drawn after seeing specific photographic objects.
This forces the drawers to sketch from memory in the same way that a user of sketch-based image retrieval systems~\cite{dey2019doodle,sain2021stylemeup,yu2016sketch} would draw from a mental image of the desired object.
23 categories overlap with ImageNet-VID: airplane, bear, bicycle, car, cat, cow, dog, elephant, horse, lion, lizard, motorcycle, rabbit, sheep, snake, squirrel, tiger, turtle, zebra.

\vspace{2mm}\noindent{\bf \tuberlin}~\cite{eitz2012humans} is a crowd-sourced sketch dataset composed of 20,000 unique sketches with 250 categories.
The sketches are uniformly distributed over 250 object categories, which exhaustively cover the vast majority of objects seen in daily life.
The median drawing time for each sketch is 86 seconds.
Due to the low quality of some sketches, humans correctly identify just 73\% of these hand-drawings.
21 categories overlap with ImageNet-VID: airplane, bear, bicycle, bus, car, cat, cow, dog, elephant, horse, lion, monkey, motorcycle, panda, rabbit, sheep, snake, squirrel, tiger, train, zebra.

\vspace{2mm}\noindent{\bf \quickdraw}~\cite{jongejan2016quick} is a huge collection of 50 million sketches organized into 345 categories.
Over 15 million players have contributed millions of sketches playing ``Quick, Draw!" game~\footnote{\url{https://quickdraw.withgoogle.com/}}, where a neural network tries to guess the sketches.
The players are asked to draw a sketch of a given category in 20 seconds while the computer attempts to classify them.
The way the sketches are collected results in a high degree of variety in the dataset, although most sketches are of low quality due to time limit.
24 categories overlap with ImageNet-VID.
\begin{table}[h!]
    \vspace{-7mm}
    \label{tab:categories}
	\begin{center}
	\tablestyle{1pt}{1.05}
	\resizebox{\linewidth}{!}{
	\begin{tabular}{x{100}|x{110}} 
		{Video $\cap$ Sketch}&{Categories}\\\shline
		{ImageNet-VID $\cap$ {\sketchy} \,\,\, (19 classes)}&{airplane, bear, bicycle, car, cat, cow, dog, elephant, horse, lion, lizard, motorcycle, rabbit, sheep, snake, squirrel, tiger, turtle, zebra}\\
		\hline
	    {ImageNet-VID $\cap$ {\tuberlin} (21 classes)}&{airplane, bear, bicycle, bus, car, cat, cow, dog, elephant, horse, lion, monkey, motorcycle, panda, rabbit, sheep, snake, squirrel, tiger, train, zebra}\\
	    \hline
	    {ImageNet-VID $\cap$ {\quickdraw} (24 classes)}&{airplane, bear, bicycle, bird, bus, car, cat, cow, dog, elephant, horse, lion, monkey, motorcycle, panda, rabbit, sheep, snake, squirrel, tiger, train, turtle, whale, zebra}
	\end{tabular}
	}
	\end{center}
    \vspace{-8mm}
\end{table}

\begin{figure*}[t!]
    \centering
    \setlength\tabcolsep{3mm}
    \resizebox*{\linewidth}{!}{%
    \subfloat[Train
    \label{fig:train}]{
        \begin{tabular}{c}
        \includegraphics[width=0.48\linewidth]{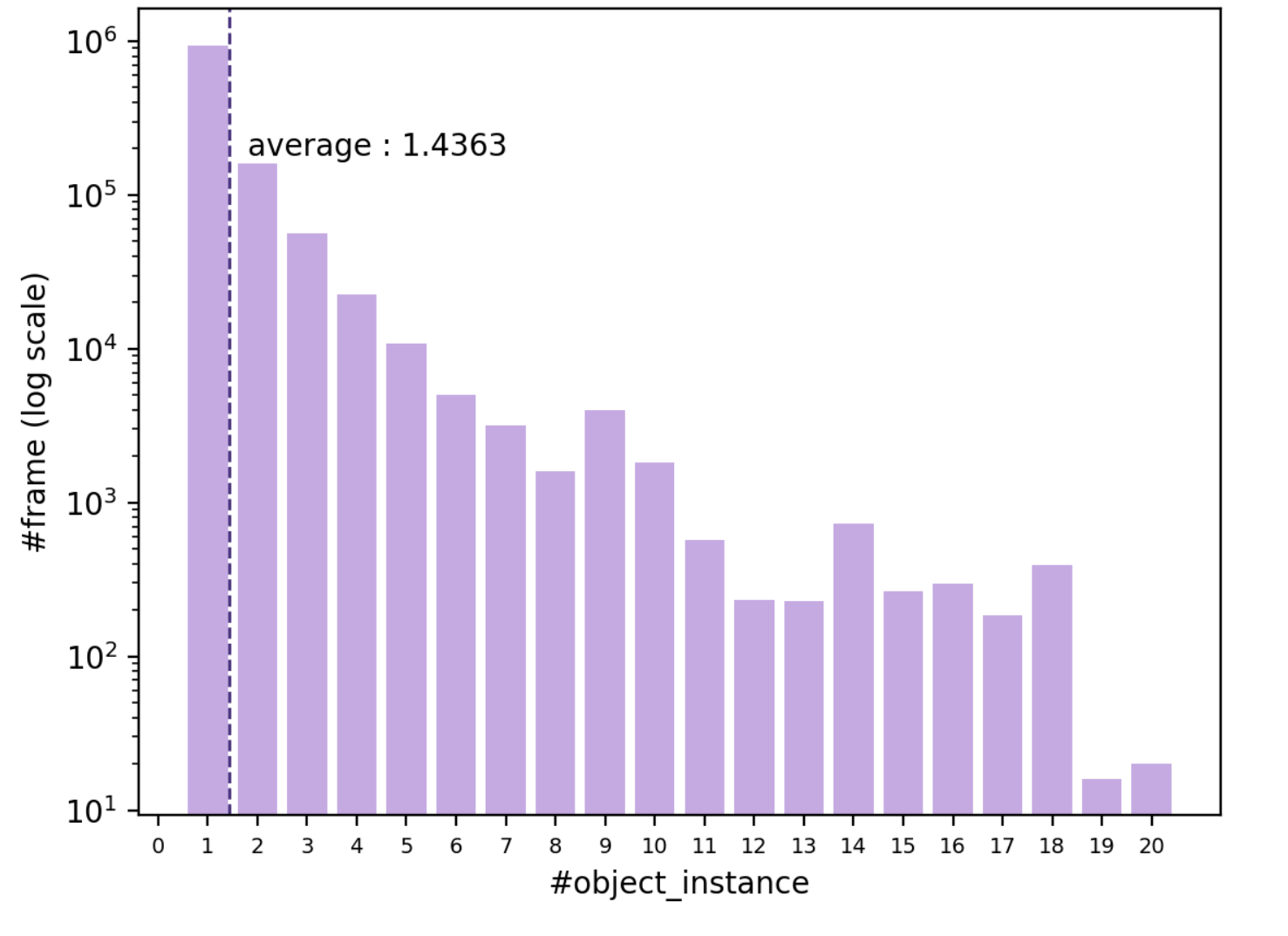}
        \end{tabular}
        }
        \hspace{-1.7ex}
    \subfloat[Val
    \label{fig:val}]{
        \begin{tabular}{c}
        \includegraphics[width=0.48\linewidth]{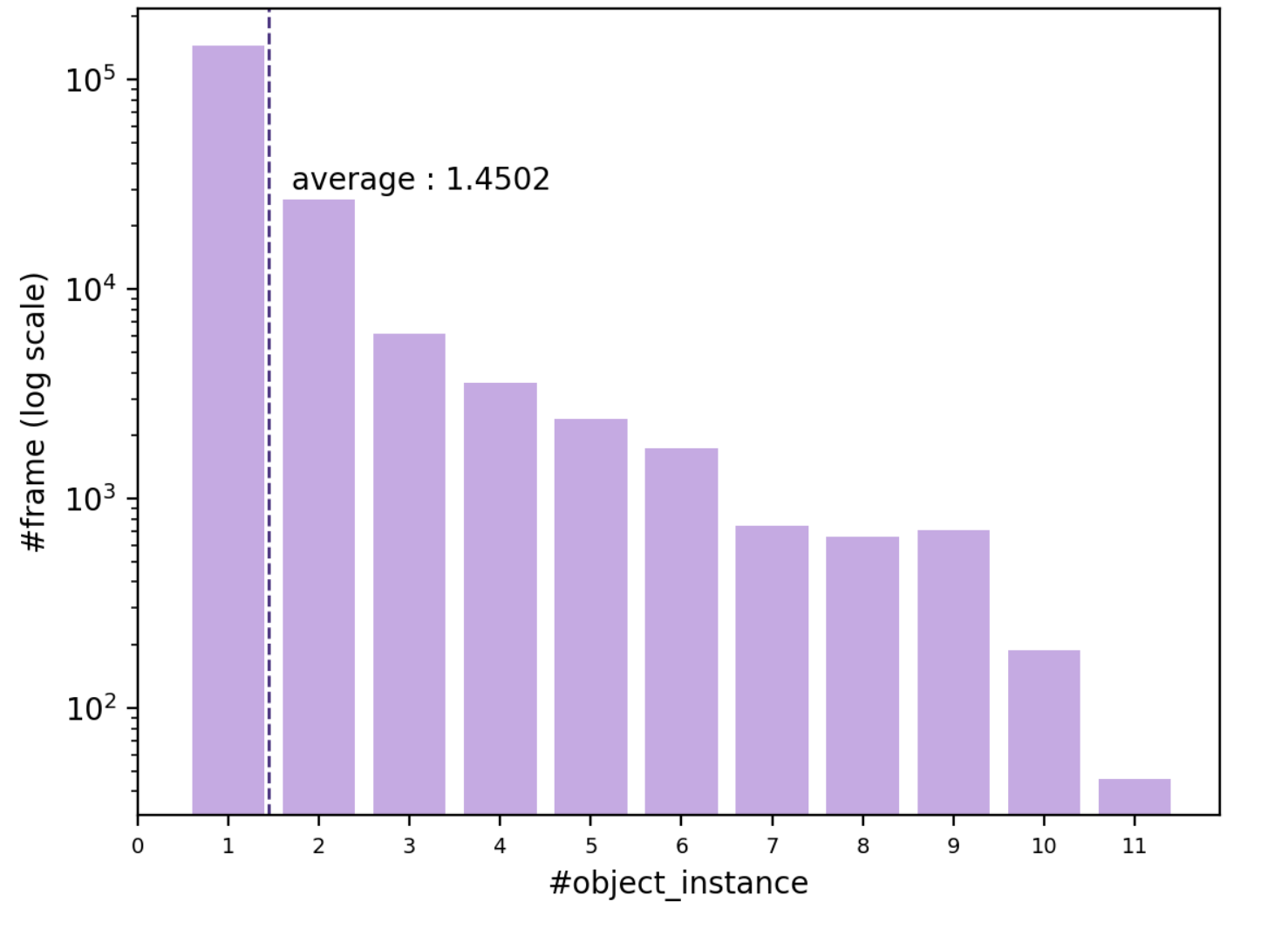}
        \end{tabular}
        }
    }
    \vskip \abovefigcapmargin
    \caption{
        \textbf{Number of object instances per frame} in~ImageNet-VID~\cite{russakovsky2015imagenet}~\protect\subref{fig:train} train and~\protect\subref{fig:val} validation split.
        The x-axis denotes the number of object instances and y-axis denotes the number of frames in a log scale.
    }
    \label{fig:num_instances}
    \vspace{\belowfigcapmargin}
\end{figure*}

\begin{figure*}[t!]
    \centering
    \setlength{\tabcolsep}{0.0em}
    \subfloat[{\bf VID-\sketchy}]{
        \begin{tabular}{c}
        \includegraphics[width=0.32\linewidth]{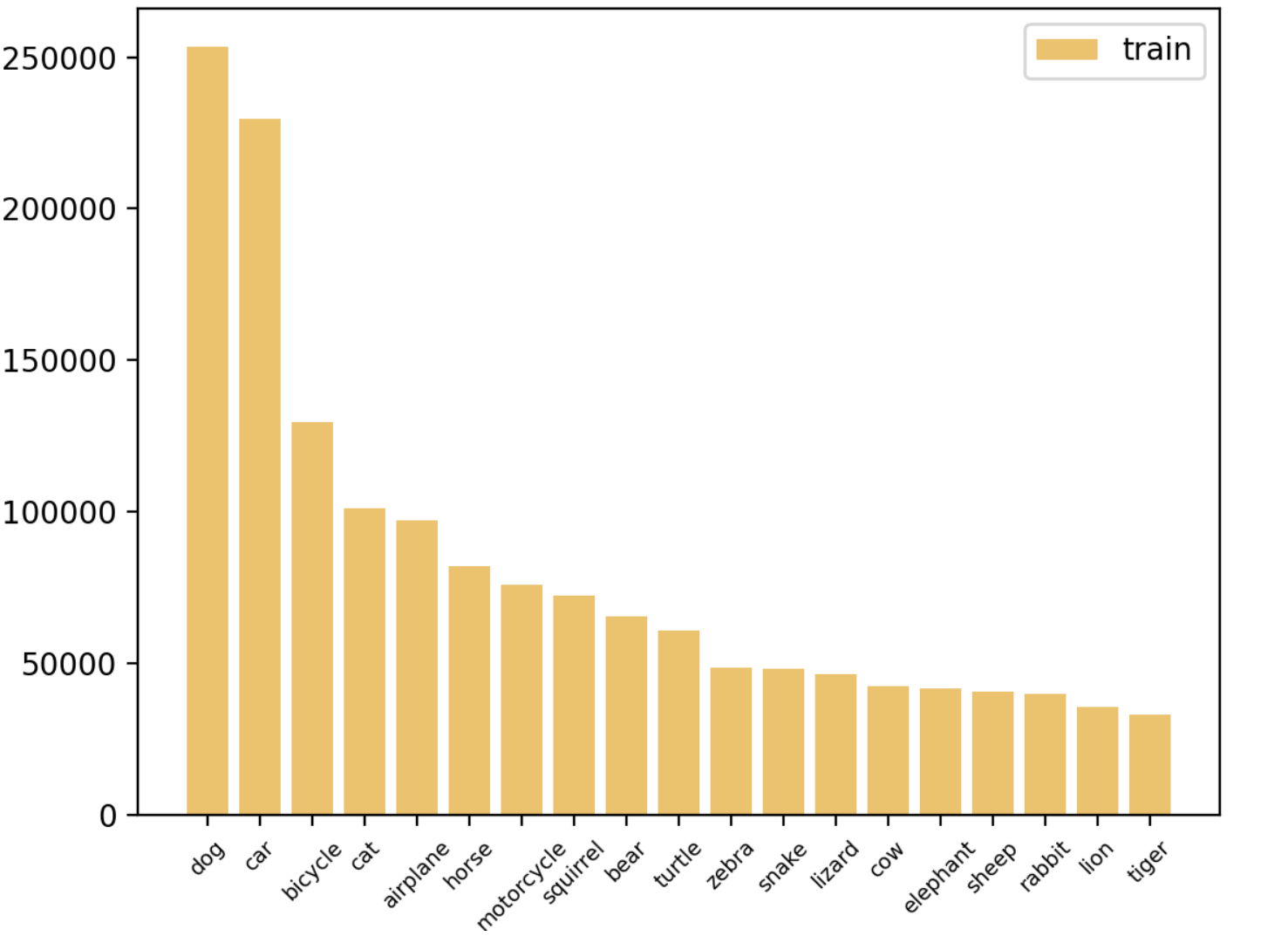} \\
        \includegraphics[width=0.32\linewidth]{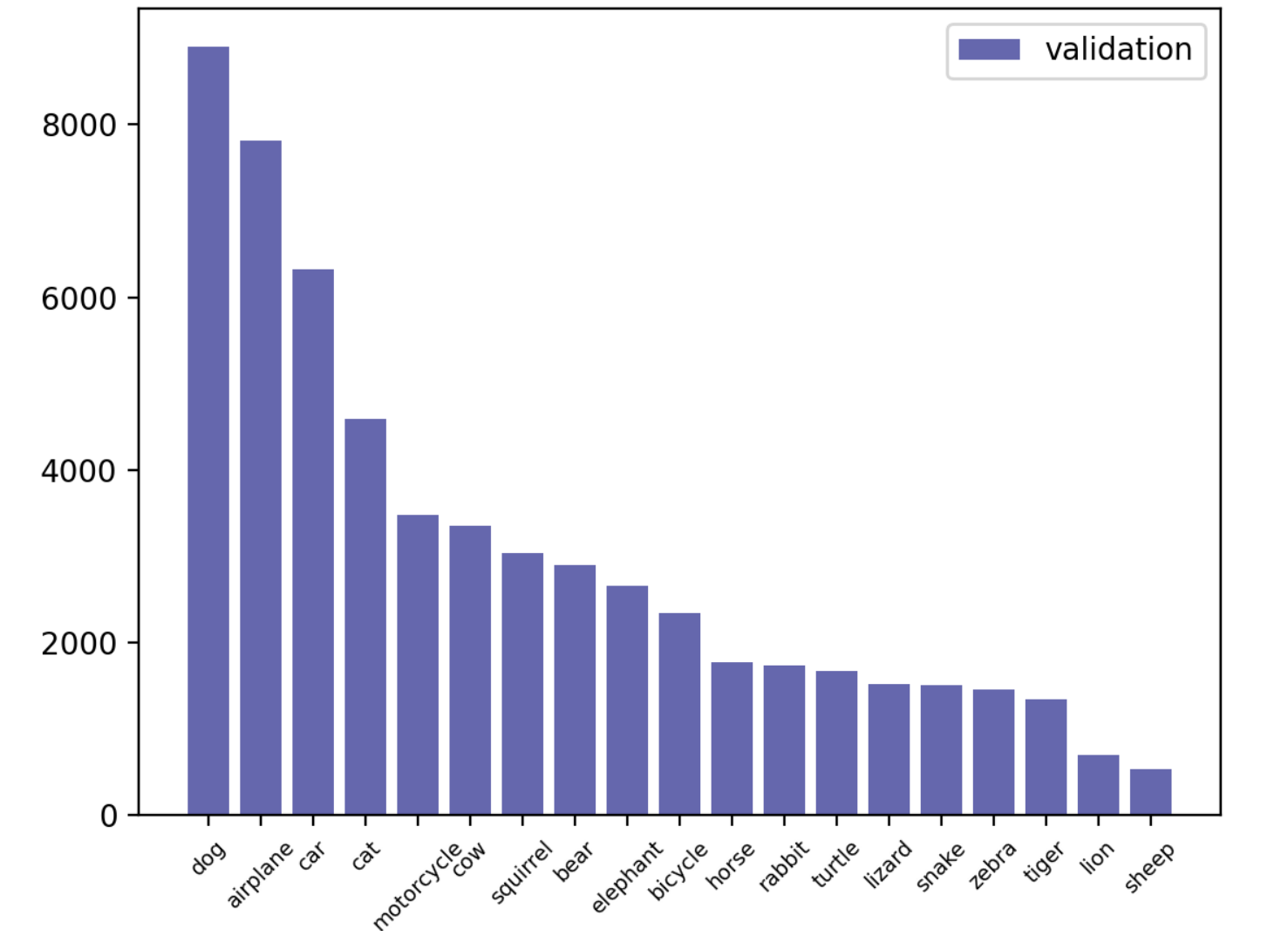}
        \end{tabular}
        \label{fig:vid_sketchy}}
        \hspace{-1.7ex}
    \subfloat[{\bf VID-\tuberlin}]{
        \begin{tabular}{c}
        \includegraphics[width=0.32\linewidth]{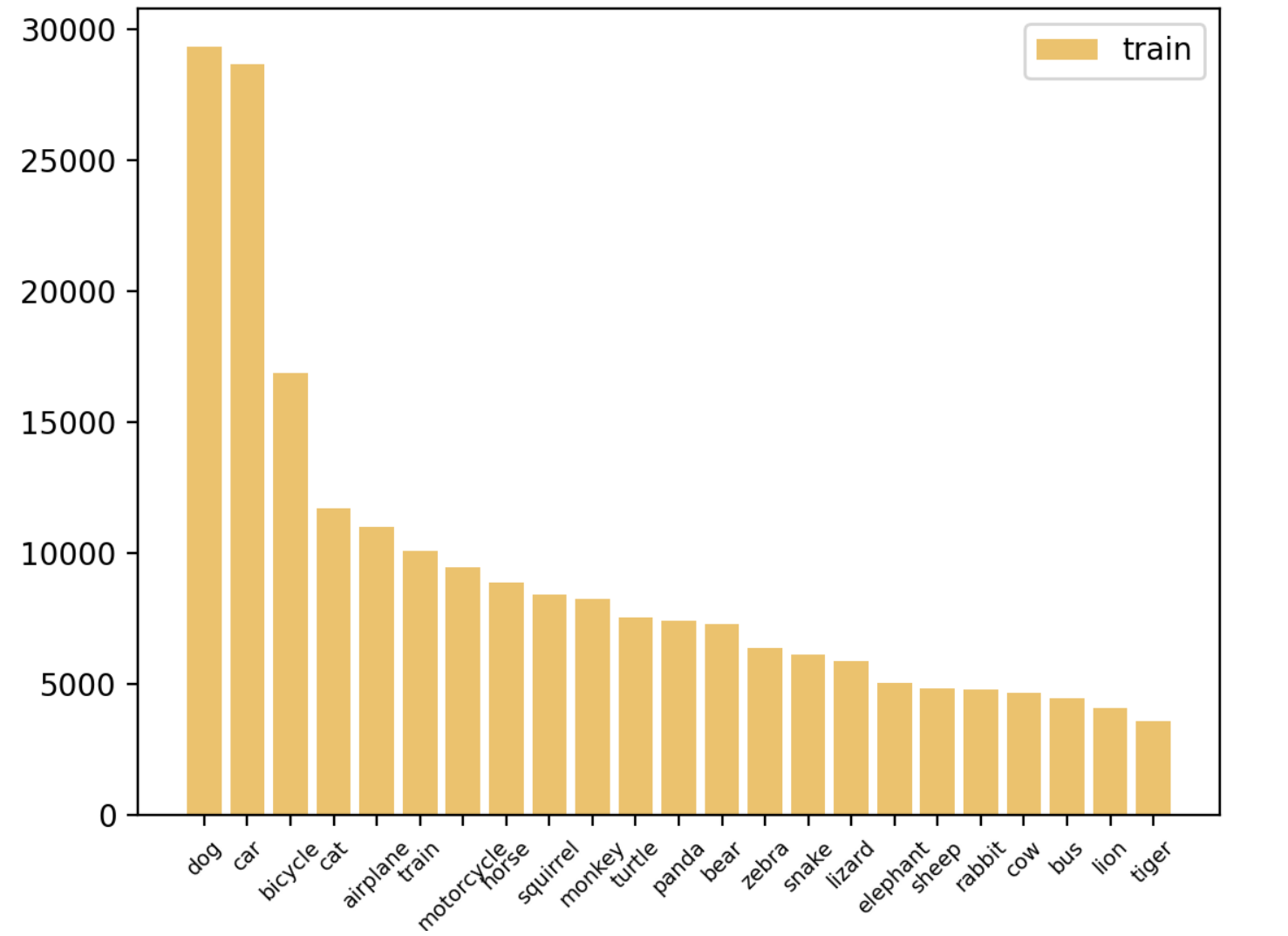} \\
        \includegraphics[width=0.32\linewidth]{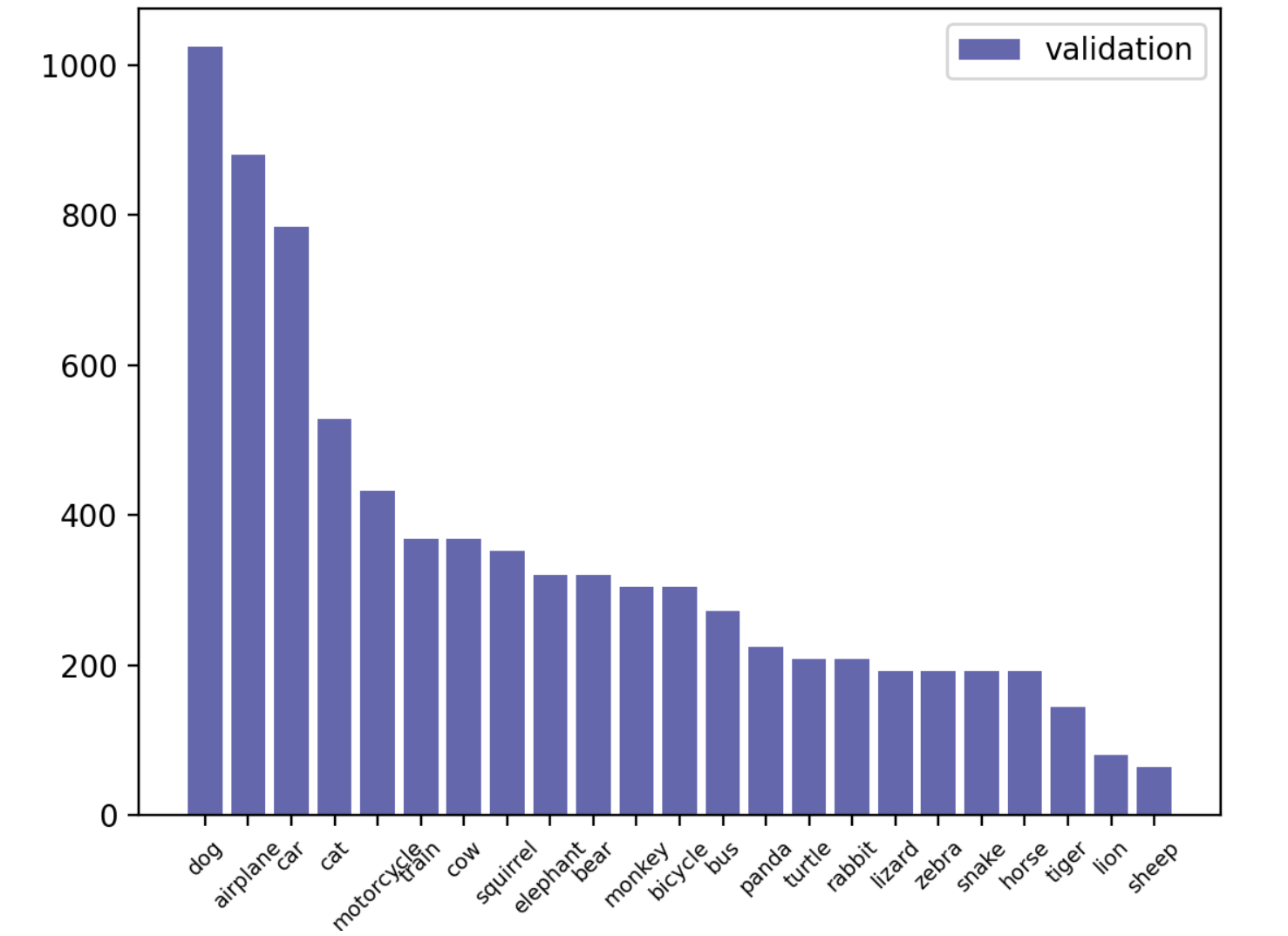}
        \end{tabular}
        \label{fig:vid_tuberlin}}
        \hspace{-1.7ex}
    \subfloat[{\bf VID-\quickdraw}]{
        \begin{tabular}{c}
        \includegraphics[width=0.32\linewidth]{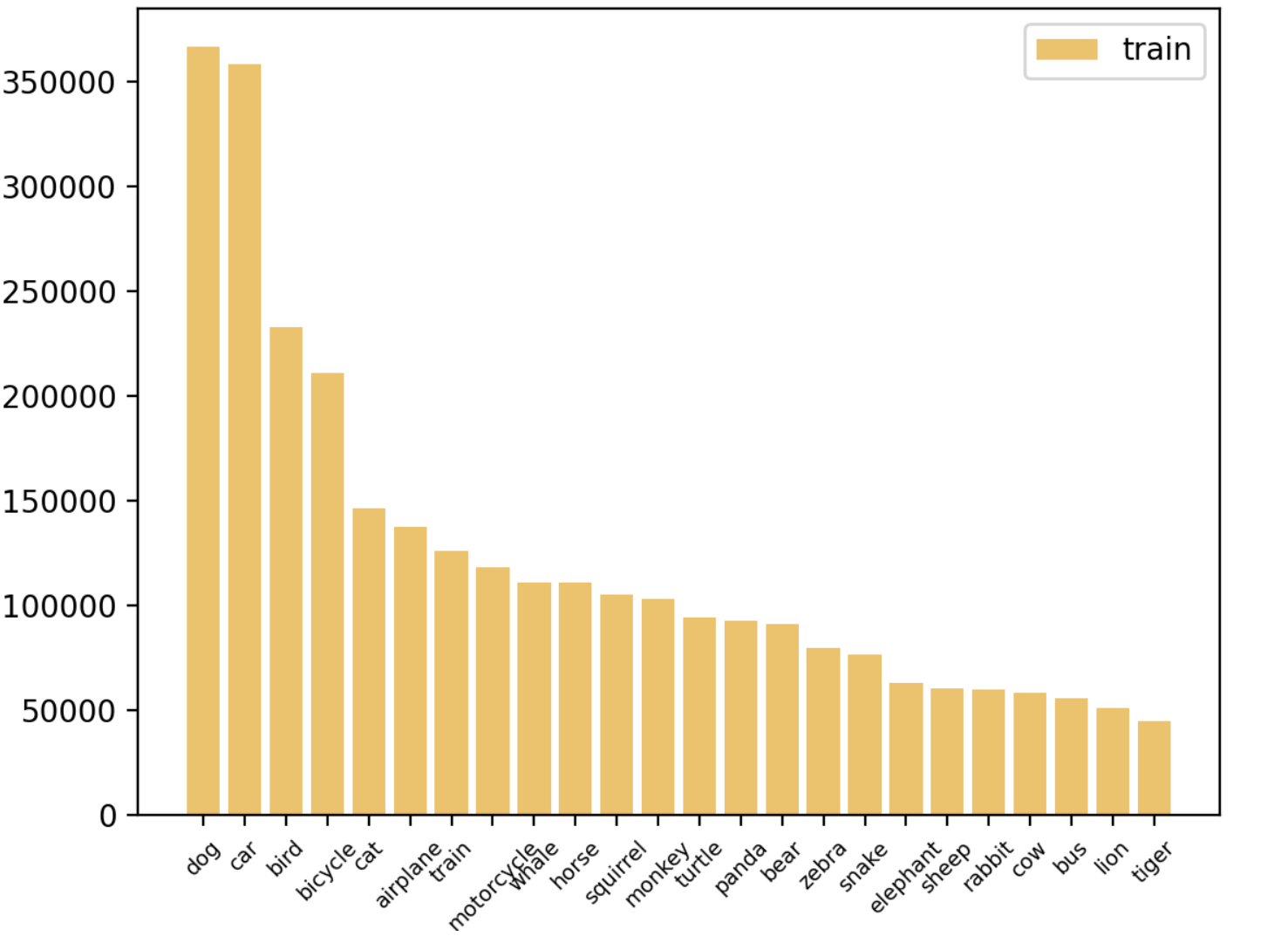} \\
        \includegraphics[width=0.32\linewidth]{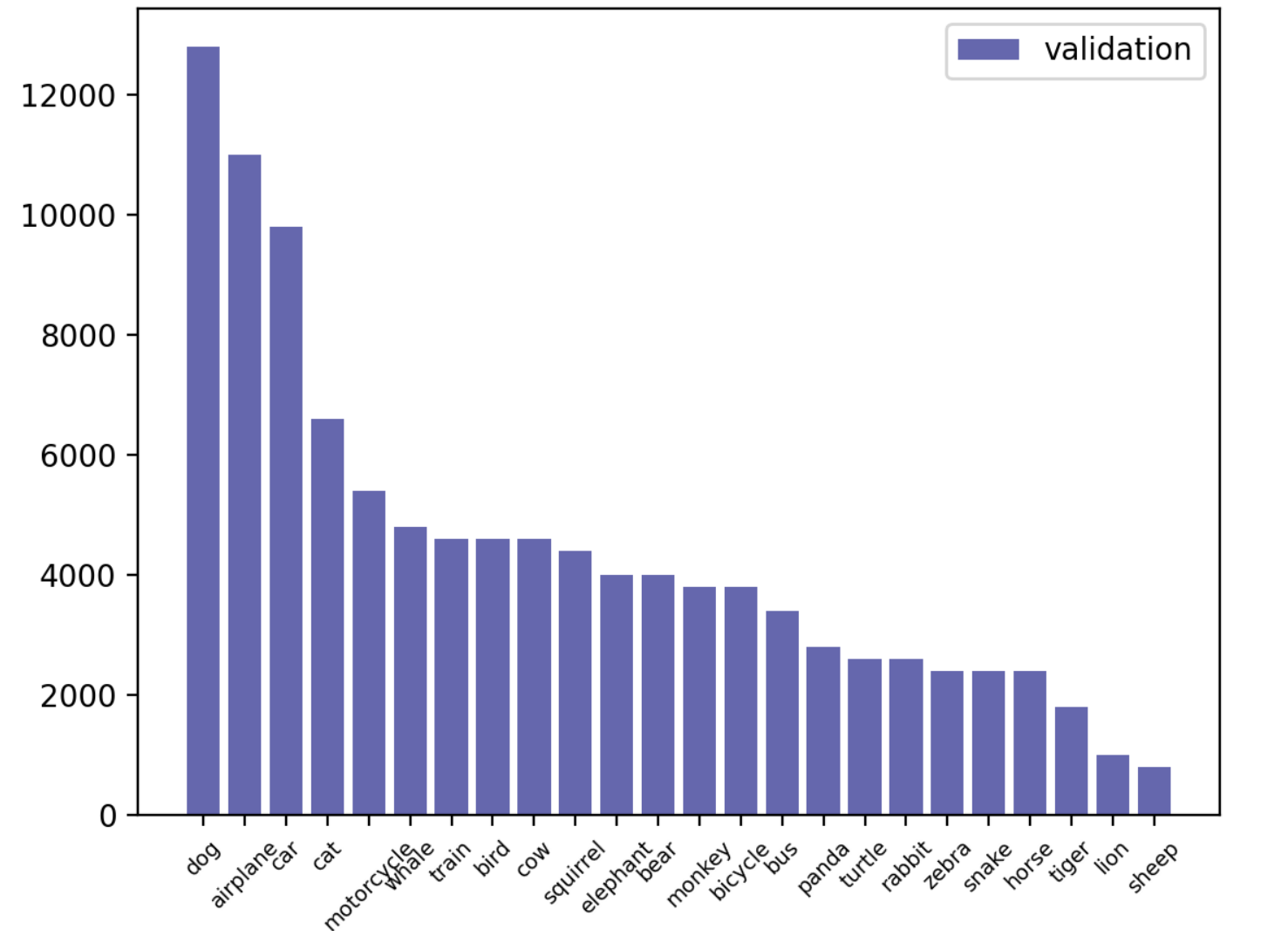}
        \end{tabular}
        \label{fig:vid_quickdraw}}
    \vskip \abovefigcapmargin
    \caption{
        \textbf{Class distribution of video-sketch pair.}
        \protect\subref{fig:vid_sketchy}~{VID-\sketchy},~\protect\subref{fig:vid_tuberlin}~{VID-\tuberlin}, and~\protect\subref{fig:vid_quickdraw}~{VID-\quickdraw}.
        The x-axis denotes the class and y-axis denotes the frequency.
    }
    \label{fig:vid_sketch_pair}
    \vspace{\belowfigcapmargin}
\end{figure*}

\vspace{\abovesubsecmargin}
\subsection{SVOL Data Analysis}

\vspace{\abovesubsubsecmargin}\subsubsection{Frame Length Distribution}\vspace{\belowsubsubsecmargin}
We use the video dataset from the ImageNet-VID dataset~\cite{russakovsky2015imagenet}.
The train split has 3,862 videos that are fully annotated with the 30 object categories, yielding 866,870 bounding boxes for 1,122,397 frames.
In validation split, 555 videos are fully annotated with the 30 object categories, resulting in 135,949 bounding box annotations for 176,126 frames.
We summarize the statistics for the frame length distribution of ImageNet-VID dataset below.
\begin{table}[h!]
    \vspace{-4mm}
    \label{tab:frame_length}
	\begin{center}
	\tablestyle{1pt}{1.05}
	\resizebox{\linewidth}{!}{
	\begin{tabular}{x{50}|x{30}|x{30}x{30}x{30}x{30}}
		{Dataset}&{Split}&{\bf min}&{\bf max}&{\bf mean}&{\bf median}\\\shline
		\multirow{2}{*}{ImageNet-VID}&Train&{6}&{5492}&{290.6}&{180}\\ 
	    {}&Val&{11}&{2898}&{317.3}&{232}
	\end{tabular}
	}
	\end{center}
\end{table}

\vspace{\abovesubsubsecmargin}
\subsubsection{Class Histogram}

    \vspace{\paramargin}\paragraph{Video dataset.}
    We show the class histogram of ImageNet-VID~\cite{russakovsky2015imagenet} dataset in~\cref{fig:video_dataset}.
    Here, \textit{ID-specific} refers to taking into account the identity (ID) of an object instance when counting the number, whereas \textit{ID-agnostic} refers to not taking it into account.
    In \textit{ID-agnostic}, the number is counted only once even if multiple object instances belonging to the same category appear in a video.
    For example, in the case of ``car'', the number is around 500 without considering the track-id, but exceeds 1,400 with considering the track-id. 
    This indicates that there are many scenes in the video in which multiple ``car'' object instances appear concurrently.
    We count only the object categories that are common in both the video and sketch datasets.
    The statistics for the class distribution of the SVOL video dataset are summarized below.
    \begin{table}[h!]
        \vspace{-7mm}
        \label{tab:video_class}
    	\begin{center}
		\tablestyle{1pt}{1.05} 
		\resizebox{\linewidth}{!}{
		\begin{tabular}{x{50}|x{40}|x{30}|x{25}x{25}x{25}x{25}}
			{Dataset}&{Track-ID}&{Split}&{\bf min}&{\bf max}&{\bf mean}&{\bf median}\\\shline
			\multirow{4}{*}{ImageNet-VID}&\multirow{2}{*}{id-specific}&Train&{67}&{1246}&{276.8}&{194}\\ 
		    {}&{}&Val&{9}&{229}&{47.7}&{29}\\\cline{2-7}
		    {}&\multirow{2}{*}{id-agnostic}&Train&{56}&{458}&{151.8}&{118}\\ 
		    {}&{}&Val&{4}&{64}&{21.8}&{19}
    	\end{tabular}
    	}
    	\end{center}
        \vspace{-7mm}
    \end{table}
    
    \vspace{\paramargin}\paragraph{Sketch dataset.}
    We show the class histograms of {\sketchy}~\cite{sangkloy2016sketchy} and {\tuberlin}~\cite{eitz2012humans} datasets in~\cref{fig:sketch_dataset}.
    The {\quickdraw}~\cite{jongejan2016quick} dataset has 1,000 sketch images per class are uniformly distributed and all train/val splits are 800/200.
    The number of sketches per class is relatively evenly distributed in the {\sketchy} dataset, however there is an imbalance between classes in the {\tuberlin} dataset.
    The following table summarizes the statistics for the class distribution of SVOL sketch datasets.
    \begin{table}[h!]
        \vspace{-3mm}
        \label{tab:sketch_class}
    	\begin{center}
		\tablestyle{1pt}{1.05}
		\resizebox{\linewidth}{!}{
		\begin{tabular}{x{40}|x{30}|x{30}x{30}x{30}x{30}}
			{Dataset}&{Split}&{\bf min}&{\bf max}&{\bf mean}&{\bf median}\\\shline
			\multirow{2}{*}{\sketchy}&Train&{486}&{594}&{539.5}&{535}\\ 
		    {}&Val&{122}&{149}&{135.4}&{134}\\\hline
		    \multirow{2}{*}{\tuberlin}&Train&{64}&{458}&{150.7}&{116}\\ 
		    {}&Val&{16}&{64}&{24.7}&{19}\\\hline
		    \multirow{2}{*}{\quickdraw}&Train&{800}&{800}&{800}&{800}\\ 
		    {}&Val&{200}&{200}&{200}&{200}
    	\end{tabular}
    	}
    	\end{center}
        \vspace{-8mm}
    \end{table}

    \vspace{\paramargin}\paragraph{Number of object instances.}
    \Cref{fig:num_instances}~shows the distribution of the number of object instances per frame in the SVOL video dataset.
    The average object instances per video is 1.4363 in the train split and 1.4502 in the validation split.

\vspace{\abovesubsecmargin}
\subsection{Data Curation}
\vspace{\belowsubsecmargin}
The SVOL dataset is made up of a combination of video dataset and sketch datasets.
To ensure that the SVOL evaluation set remain unseen in the training phase, we split videos and sketches into training and evaluation sets, respectively, and then construct a training set of SVOL with a combination of both training sets, and an evaluation set of SVOL with a combination of both evaluation sets.
This guarantees the models to be evaluated on video-sketch pairs that are totally unseen throughout the training phase.
While this setting is most closest to the actual environment in which the SVOL system operates, it requires the model to be generalized to both videos and sketches.

Formally, let $\{\mathcal{V}_{tr}, \mathcal{V}_{ev}\}$ train/eval video datasets, $\{\mathcal{S}_{tr}, \mathcal{S}_{ev}\}$ train/eval sketch datasets, and $\{\mathbb{C}_{\mathcal V}, \mathbb{C}_{\mathcal S}\}$ video/sketch category sets.
For all categories that are common for video and sketch datasets, \ie, $\forall c \in \mathbb{C}_{\mathcal V} \cap \mathbb{C}_{\mathcal S}$, we construct SVOL train set by pairing $\mathcal{V}_{tr}$ and $\mathcal{S}_{tr}$, and SVOL eval set with $\mathcal{V}_{ev}$ and $\mathcal{S}_{ev}$.
Only video and sketch are paired when they have the same class label.
The number of pairs generated for each video-sketch datasets are summarized in the table below (\Cref{fig:vid_sketch_pair} shows the per-category distribution).
\begin{table}[h!]
    \vspace{-6mm}
    \label{tab:vid_sketch_pairs}
    \begin{center}
    \resizebox{\linewidth}{!}{
    \tablestyle{1pt}{1.05}
    \begin{tabular}{x{30}|x{65}x{65}x{65}}
        {Split}&\textbf{VID-{\sketchy}}&\textbf{VID-{\tuberlin}}&\textbf{VID-{\quickdraw}}\\\shline
        {Train}&{1,545,801}&{215,040}&{2,958,400}\\ 
        {Eval}&{57,660}&{7,952}&{10,6400} 
    \end{tabular}
    }
    \end{center}
    \vspace{-7mm}
\end{table}

In practice, we only use videos that contain at least one query sketch object within 32 frames uniformly sampled from the video, and bounding box annotations that correspond to the sketch object are regarded as the ground truths for that pairing.
A video can be paired more than once since it can contain multiple objects.
We note that the class label is only a means for pairing and is not considered in actual training.
\begin{figure}[t!]
    \centering
    \setlength\tabcolsep{0.5mm}
    \resizebox*{\linewidth}{!}{%
    \begin{tabular}{cccc}
        \addlinespace[0.2em]
        \rot{~~~{\bf \ours (Ours)}}&\includegraphics[width=0.33\linewidth, height=0.33\linewidth]{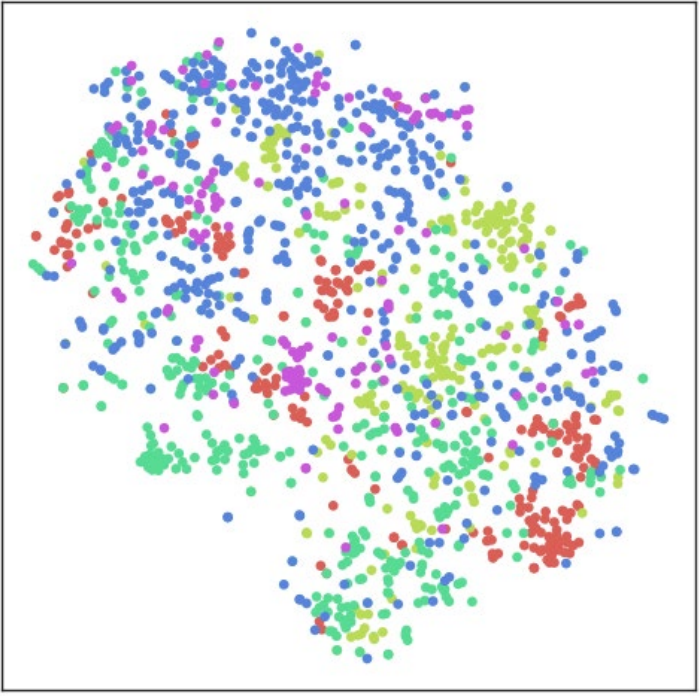}&\includegraphics[width=0.33\linewidth, height=0.33\linewidth]{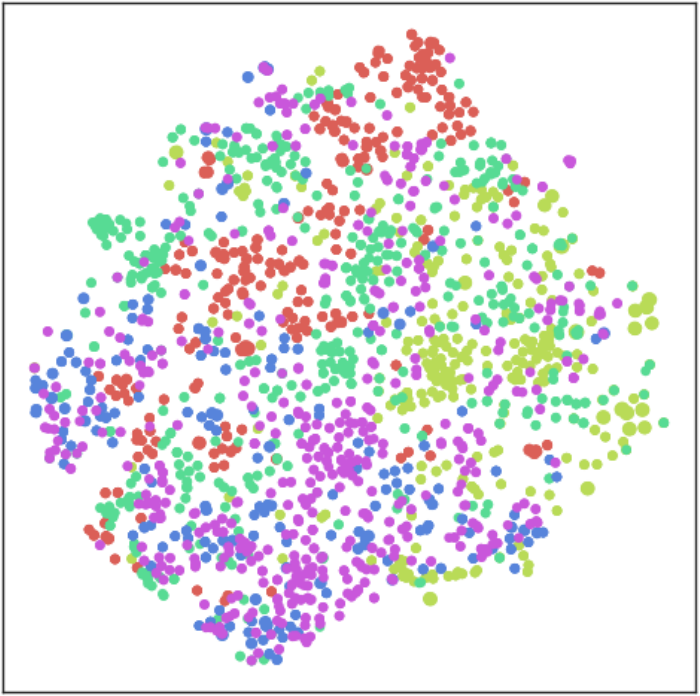}&\includegraphics[width=0.33\linewidth, height=0.33\linewidth]{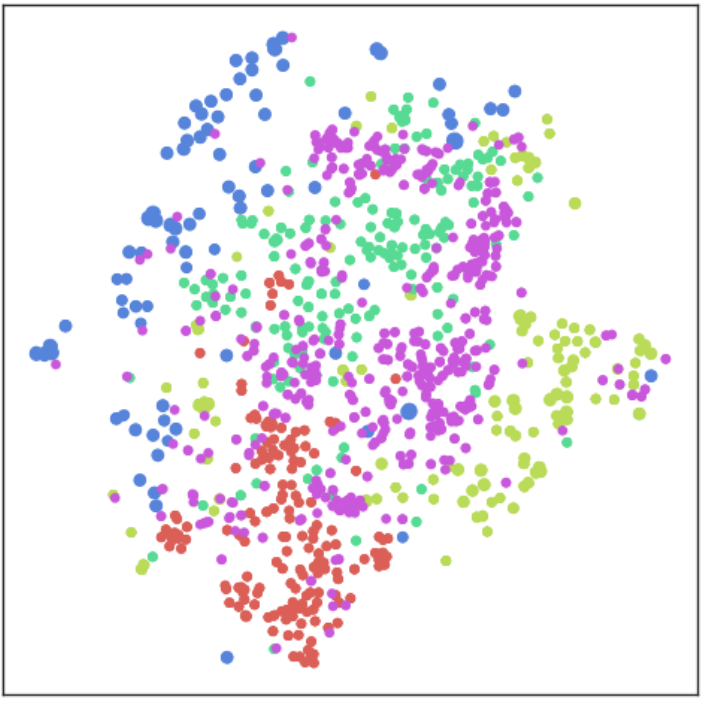}\\
        \rot{~~~~{\bf Sketch-DETR}}&\includegraphics[width=0.33\linewidth, height=0.33\linewidth]{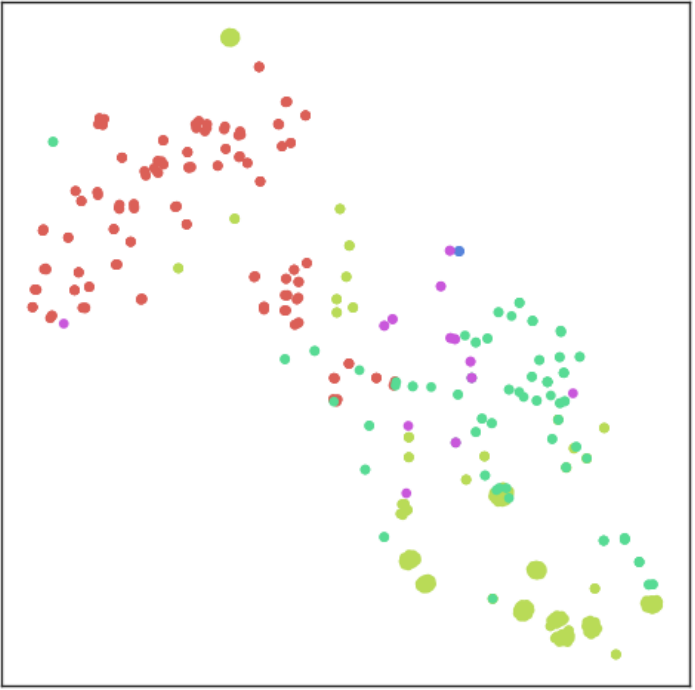}&\includegraphics[width=0.33\linewidth, height=0.33\linewidth]{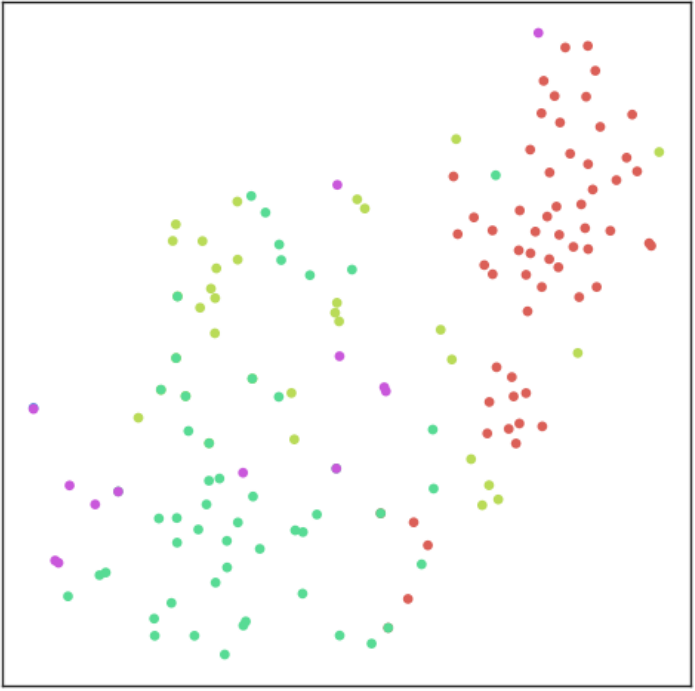}&\includegraphics[width=0.33\linewidth, height=0.33\linewidth]{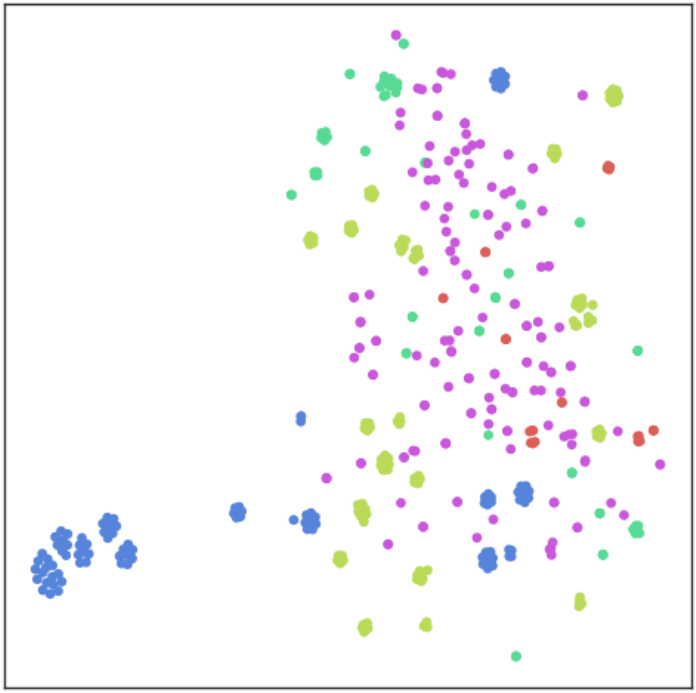}\\
        &(a)~{\color{sketchy} S}$\rightarrow${\color{tuberlin} T}&(b)~{\color{sketchy} S}$\rightarrow${\color{quickdraw} Q}&(c)~SC$\rightarrow$USC (w/ {\color{sketchy} S})
    \end{tabular}%
    }
    \vskip \abovefigcapmargin
    \caption{
        \textbf{Feature distribution of \ours \vs Sketch-DETR~\cite{riba2021localizing}} when transfer is performed for the cases of (a)~{\sketchy}$\rightarrow${\tuberlin}, (b)~{\sketchy}$\rightarrow${\quickdraw}, and (c)~Seen$\rightarrow$Unseen Categories with the {\sketchy} dataset.
        Each data point represents the last hidden state of the CMT, and the color indicates the category it belongs to.
        We plot samples of 5 random categories with a confidence score higher than 0.9.
    }
    \label{fig:tsne}
    \vspace{\belowfigcapmargin}
\end{figure}

\begin{figure*}[t!]
    \centering
    \includegraphics[width=\linewidth]{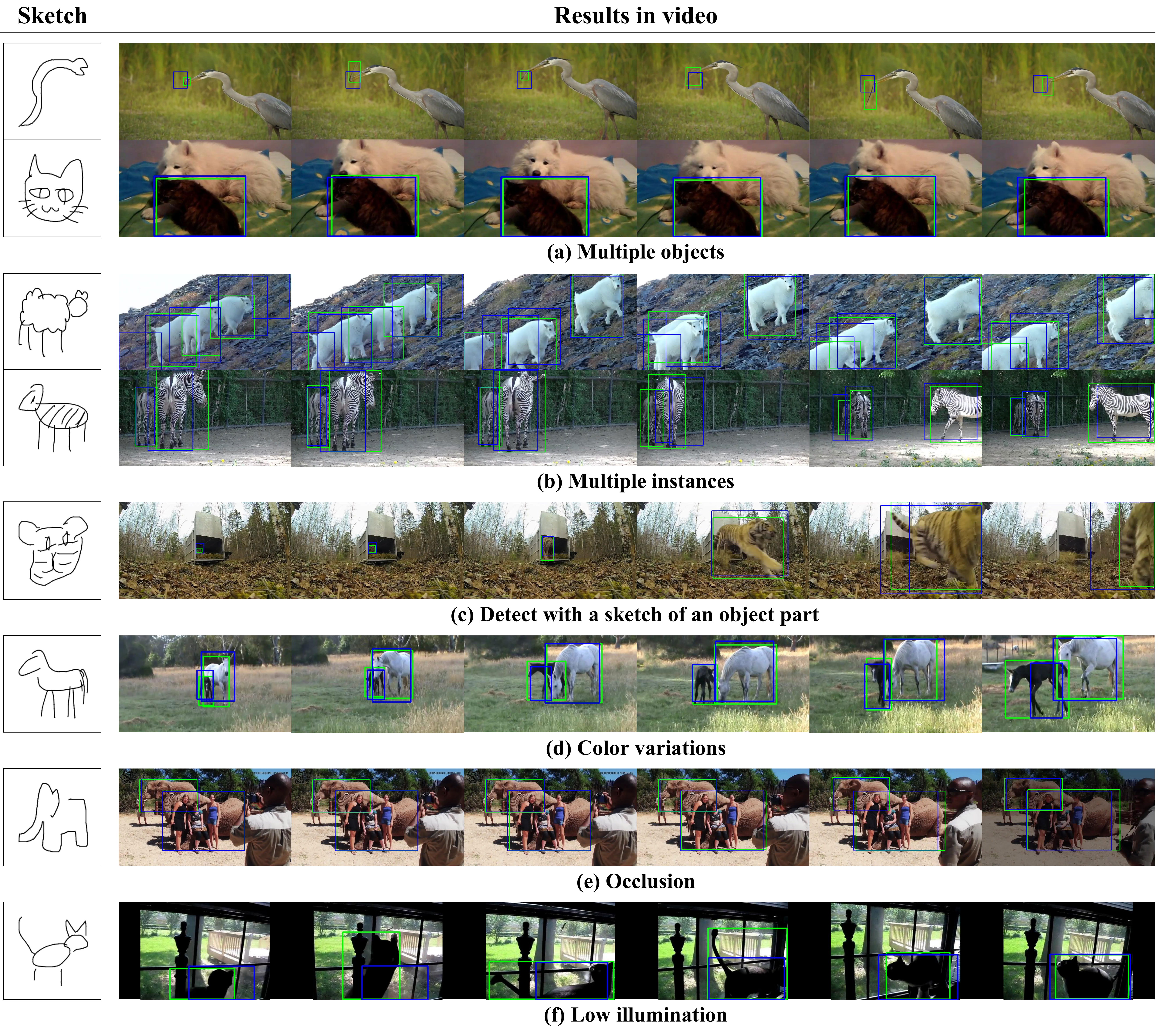}
    \vskip \abovefigcapmargin
    \caption{
        \textbf{\color{blue} Success cases} of \ours on {\quickdraw} dataset.
        {\color{green} Green} and {\color{blue} blue} boxes represent {\color{green} ground truths} and {\color{blue} predictions}, respectively.
    }
    \label{fig:qualitative_success}
    \vspace{\belowfigcapmargin}
\end{figure*}

\begin{figure*}[t!]
    \centering
    \includegraphics[width=\linewidth]{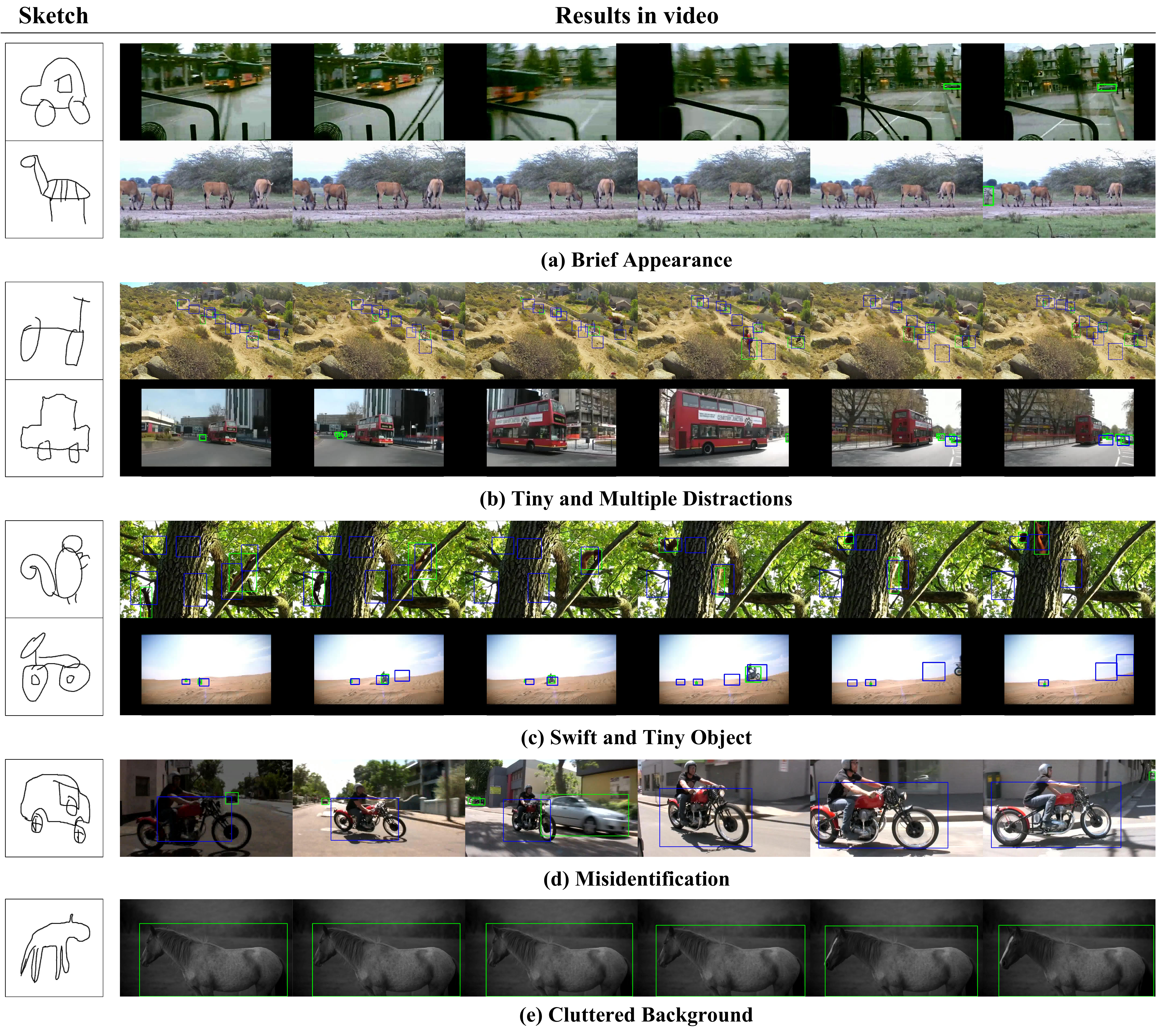}
    \vskip \abovefigcapmargin
    \caption{
        \textbf{\color{red} Failure cases} of \ours on {\quickdraw} dataset.
        {\color{green}Green} and {\color{blue}blue} boxes represent {\color{green}ground truths} and {\color{blue}predictions}, respectively.
    }
    \label{fig:qualitative_fail}
    \vspace{\belowfigcapmargin}
\end{figure*}

\vspace{\abovesecmargin}
\section{Additional Qualitative Results}
\label{sec:additional_qualitative}

    \vspace{\paramargin}\paragraph{Feature distribution in transfer evaluation.}
    In~\cref{fig:tsne}, we perform transfer evaluation and visualize the feature distribution of \ours and that of Sketch-DETR using t-SNE~\cite{van2008visualizing}.
    Here, (a) and (b) depict the results in dataset-level transfer, whereas (c) represents the result of category-level transfer.
    In other words, for (a) and (b), models trained on the {\sketchy} dataset ({\tuberlin} dataset for (b)) were employed to map the feature distributions of samples from the {\tuberlin} dataset ({\quickdraw} dataset for (b)).
    On the other hand, for (c), models trained on certain categories were utilized to visualize the feature distributions of samples from previously unseen categories.
    Compared to Sketch-DETR, \ours appears to be nicely clustered when transferred to unseen datasets, which implies that \ours effectively captures class-discriminative representations.
    When transferred to unseen categories, \ours embeds the same category into a similar subspace, demonstrating that learned sketch-video mapping can generalize well.
    Moreover, \ours shows denser distribution than Sketch-DETR, \ie, only a few data points reach the threshold confidence 0.9 in Sketch-DETR, indicating that our \ours produces more reliable predictions.
    %
    %

\vspace{\paramargin}\paragraph{SVOL results: {\color{blue} success cases}.}
\Cref{fig:qualitative_success} shows success cases of \ours.
Our system successfully recognizes the objects that correspond to the query sketch and accurately localizes their bounding boxes in a variety of challenging conditions:
(a) various objects appears in a video;
(b) multiple object instances with different pose and shape appear in a video;
(c) only sketch of a part (face) of object is given;
(d) the target objects have different colors;
(e) the target objects are occluded by other objects;
(f) bad illumination condition.

\vspace{\paramargin}\paragraph{SVOL results: {\color{red} failure cases}.}
\Cref{fig:qualitative_fail} shows failure cases of \ours.
\ours suffers particularly when the target object:
(a) appears for a very short time (almost 1 or 2 frames out of 32 frames);
(b) is too small, and there are numerous distracting factors;
(c) is small and moves quickly;
(d) is non-salient (here, the target object is a car, but a motorcycle, is detected);
(e) is similar to the background.
\vspace{\abovesecmargin}
\section{Discussion}
\label{sec:discussion}

\vspace{\abovesubsecmargin}\subsection{Why Sketch Query?}\vspace{\belowsubsecmargin}
Sketch query can be more flexible and efficient than language or image query as it allows for more natural and intuitive user input.
With sketch query, users can quickly and easily provide a rough sketch of the object they are looking for, rather than having to use specific keywords or search through a pre-existing database of images.
This can make it easier for users to find the specific object they are looking for, especially if the image is not easily describable with keywords or if the image does not exist in a pre-existing database.
Moreover, sketch query can transcend the language barrier, and can be less prone to ambiguity and errors as the user is able to provide a visual representation of the desired object.
On the other hand, language query requires additional translation when the user's language changes (\eg, English $\rightarrow$ French).
Additionally, since sketch queries are basically embodiments of real-world objects, the model inherently learns over the visual similarity between the query sketch and the video objects.
Therefore, the model can leverage such inductive bias of appearance matching for unseen categories.
Sketch query offers a great degree of freedom and can overcome several limitations that other queries (\eg, language or image) may include.
There are several more advantages in using sketch query:
\begin{enumerate}
\item As the use of touch screen devices (\eg, smartphones, tablets) has recently increased, acquisition of sketch data has become easier.
\item Sketch is not bound by the user's age, race, and nation, and even those with language difficulties can communicate their thoughts;
\item Our model is effective even with a low-quality sketch (\eg, {\quickdraw}), therefore users are not required to draw well;
\end{enumerate}

\vspace{\abovesubsecmargin}\subsection{Limitations}\vspace{\belowsubsecmargin}
While we believe that focusing on category-level localization in the SVOL problem can take advantage of the abstract nature of sketches, it is also important to consider the limitations of this setting.
In this setting, the system may lose nuanced understanding of sketches that could be useful for precisely identifying objects of the same category with different details.
For example, if we want to localize a car of a specific make and model, the system may not be able to do so accurately as it is not explicitly taught to differentiate objects within the same category during training.
In order to improve the versatility of the SVOL system, future research may investigate on incorporating fine-grained data sources to differentiate objects within the same category.

We also recognize that transfer performance for unseen categories is still far from enough, yet this shows that SVOL is a challenging problem and suggests that better solutions should be found.
We hope our findings and analysis will encourage further research in this direction.

\vspace{\abovesubsecmargin}\subsection{Future Work}\vspace{\belowsubsecmargin}
We hope future work will develop approaches for the following.

\vspace{\paramargin}\paragraph{SVOL in large-scale video collection.}
On an online video platform, users often need to quickly and efficiently find the location of a specific object of interest amid large-scale video collections.
In order to be practical in such situations, the SVOL system should be able to retrieve relevant videos, and accurately localize the target objects within the set of retrieved videos.
This is similar to the setting for video corpus moment retrieval~\cite{zhang2021video}.
Such a system could greatly enhance the user experience by allowing them to quickly locate the desired object within the video corpus, making the process of finding relevant information faster and more efficient.
Although it is beyond the scope of our current work, we believe it to be promising area for future research.

\vspace{\paramargin}\paragraph{Domain adaptative methods.}
The significant difference in the appearance and structure between sketches and natural videos poses a challenge for the SVOL system to accurately match them.
To alleviate this issue, various domain adaptation techniques~\cite{ganin2015unsupervised,ganin2016domain,peng2019moment} can be employed.
These techniques aim to align the feature representations of the sketches and natural videos, thus reducing the domain gap.
By utilizing these techniques, we anticipate further improvements in the performance of the SVOL system.

\vspace{\paramargin}\paragraph{Fine-grained SVOL.}
In this work, we define the SVOL task to be agnostic to shape and pose within the same class, allowing us to localize objects in a video by sketching only key features that are unique and distinctive to that object, such as the ears, eyes, and tails of a cat.
This setting has the advantage of being able to identify and locate the object in the video, even if the object's shape and pose change over time.
However, this setting also has its limitations, as it may miss important details that could be useful for differentiating objects within the same class.
Therefore, it may be worth exploring a more fine-grained approach to the SVOL problem, by focusing on detailed instance-level information such as shape, pattern, and pose, which can be used to distinguish objects of the same category~\cite{yu2021fine}.
This shape- and pose-specific approach, however, also comes with its own challenges.
For instance, it may be difficult to match a still sketch to a moving object in a video, as the shape or pose of objects continues to change over time.
Thus, to make this approach work effectively, it is essential to have a suitable data pairing that takes into account the dynamic nature of the video.
Additionally, further research could explore ways to effectively balance between leveraging the abstract nature of sketches and preserving enough fine-grained details for precise object identification.

\vspace{\paramargin}\paragraph{On-the-fly SVOL.}
As opposed to our SVOL setting, which requires a \textit{complete} sketch to be drawn before localization can begin, the ``on-the-fly'' setting allows for localization to start as soon as the user begins drawing~\cite{bhunia2020sketch}.
This approach utilizes each stroke that is drawn in real-time to match it to objects in the video.
This allows for sketch-object matching with an \textit{incomplete} sketch (\ie, just a few strokes), which can greatly reduce the time and effort required to draw an accurate sketch.
Furthermore, the system can provide immediate feedback based on the ongoing localization results as the user continues to draw, allowing for a more efficient and user-friendly experience.
It can help the user to understand how well their sketch is matching with the objects in the video and make adjustments accordingly.
This can also make the task of drawing accurate sketches more manageable for users with less experience or skill.

\vspace{\abovesubsecmargin}\subsection{Broader Impacts}\vspace{\belowsubsecmargin}
Our \ours makes predictions based on learned statistics of the collected dataset, which may reflect biases present the data, including ones with negative societal impacts.
The predictions may not be accurate, thus users exercise caution and should not rely solely on them in real-world applications and it is recommended to use it in conjunction with other forms of analysis and decision-making.
Further consideration is warranted regarding this issue.

 \fi

\end{document}